\documentclass[Afour,sageh,times]{sagej}
\usepackage{moreverb,url}

\usepackage{xcolor}
\definecolor{darkblue}{rgb}{0.0, 0.0, 0.5}
\usepackage[colorlinks,bookmarksopen,bookmarksnumbered,citecolor=darkblue,urlcolor=darkblue,linkcolor=darkblue]{hyperref}

\newcommand\BibTeX{{\rmfamily B\kern-.05em \textsc{i\kern-.025em b}\kern-.08em
T\kern-.1667em\lower.7ex\hbox{E}\kern-.125emX}}

\usepackage{amsmath,amsfonts}
\usepackage{algorithmic}
\usepackage{algorithm}
\usepackage{array}
\usepackage[caption=false,font=normalsize,labelfont=sf,textfont=sf]{subfig}
\usepackage{textcomp}
\usepackage{stfloats}
\usepackage{float}
\usepackage{url}
\usepackage{verbatim}
\usepackage{graphicx}
\usepackage{svg}
\usepackage{epstopdf}
\usepackage{adjustbox}

\usepackage{xstring}

\newcommand\mycite[1]{\defcitealias{#1}{\StrBefore{#1}{_}[\Name]\StrLeft{\Name}{6}-\StrBetween[1,2]{#1}{_}{_}[\Pname]\StrLeft{\Pname}{3}}[\citetalias{#1}]}

\setcitestyle{round, aysep={},yysep={,}, citesep={;}} 

\usepackage{tabularx}
\newcolumntype{Y}{>{\centering\arraybackslash}X}
\usepackage{tikz}
\usepackage{soul}

\usepackage{pgfplots}
\usepackage{pgfplotstable}
\pgfplotsset{
  compat=newest,
  xlabel near ticks,
  ylabel near ticks
}

\begin{document}

\runninghead{Singamaneni et al.}

\title{A Survey on Socially Aware Robot Navigation: Taxonomy and Future Challenges
}

\author{Phani Teja Singamaneni\affilnum{1},
Pilar Bachiller-Burgos\affilnum{2}, Luis J. Manso\affilnum{3}, 
Anaís Garrell\affilnum{4},
Alberto Sanfeliu\affilnum{4},
Anne Spalanzani\affilnum{5},
and Rachid Alami\affilnum{1}
}

\affiliation{\affilnum{1}{LAAS-CNRS, Universite de Toulouse, Toulouse, France}\\
\affilnum{2}{RoboLab, Universidad de Extremadura, C\'aceres, Spain}\\
\affilnum{3}{Aston University, Birmingham, UK}\\
\affilnum{4}{Universitat Politècnica de Catalunya, Barcelona, Spain}\\
\affilnum{5}{Universite Grenoble Alpes, Inria, 38000 Grenoble, France}
}

\corrauth{Phani Teja Singamaneni, LAAS-CNRS, Toulouse, France.}

\email{ptsingaman@laas.fr}

\begin{abstract}

Socially aware robot navigation is gaining popularity with the increase in delivery and assistive robots. The research is further fueled by a need for socially aware navigation skills in autonomous vehicles to move safely and appropriately in spaces shared with humans. 
Although most of these are ground robots, drones are also entering the field.
In this paper, we present a literature survey of the works on socially aware robot navigation in the past 10 years. We propose four different faceted taxonomies to navigate the literature and examine the field from four different perspectives. Through the taxonomic review, we discuss the current research directions and the extending scope of applications in various domains. Further, we put forward a list of current research opportunities and present a discussion on possible future challenges that are likely to emerge in the field.

\end{abstract}


\maketitle
\footnotetext{\textit{This is the accepted version of manuscript at International Journal of Robotics Research.}}

\section{Introduction}
\begin{figure}
    \centering
    \includegraphics[width=0.9\columnwidth]{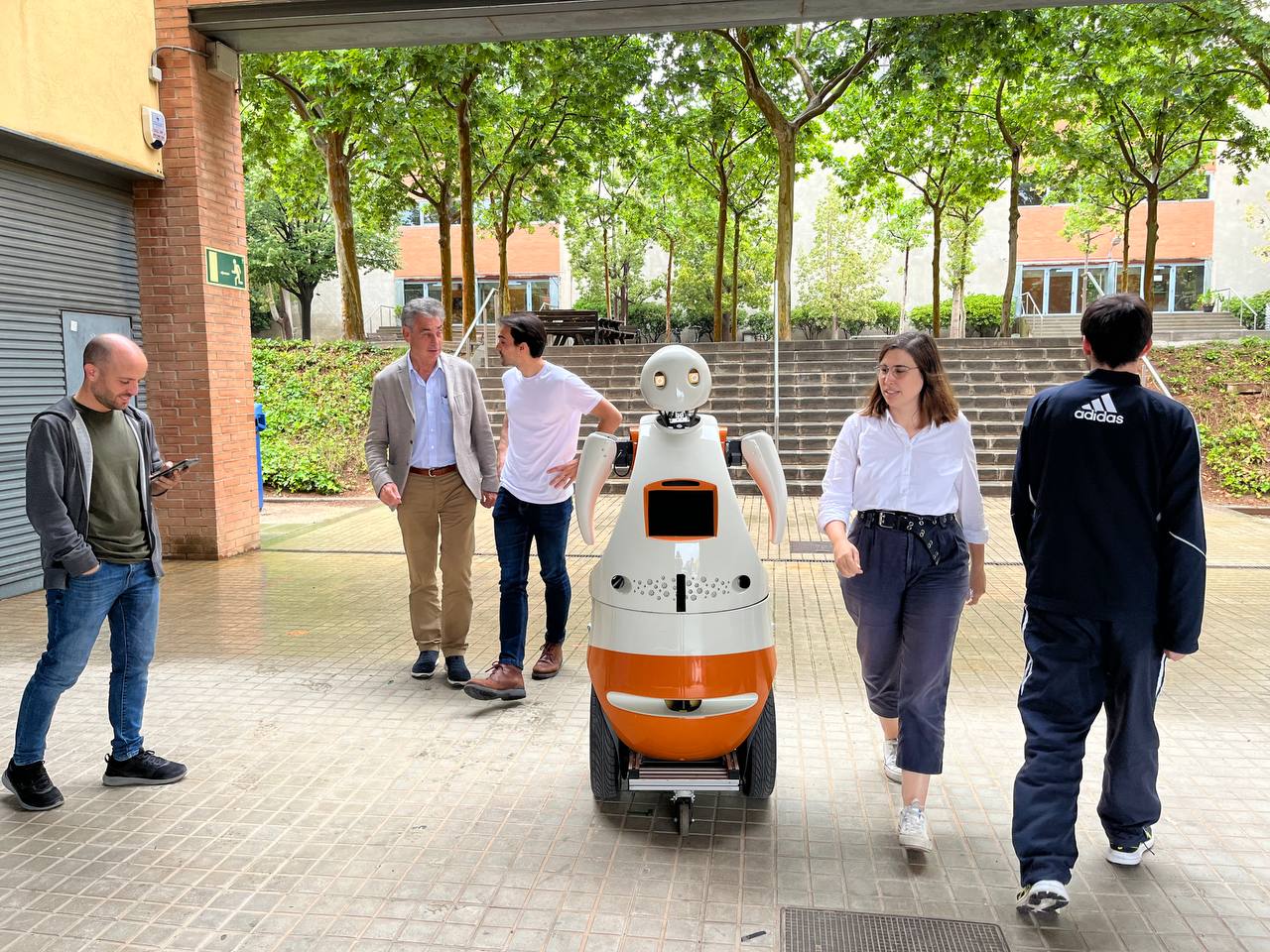}
    \caption{An autonomous ground robot figuring its way among humans.}
    \label{main-fig}
\end{figure}

Socially aware robot navigation has steadily gained interest in recent years, becoming a research field of its own.
With the increase in the number of service robots and autonomous vehicles, it is crucial for them to be able to carry out their tasks around humans efficiently and seamlessly.
This applies not only to their ultimate goals but also to all skills that these build on top of, including navigation.
Socially appropriate behavior is key for a robot to gain acceptance from humans and prevent causing any discomfort.
In addition to mobile robots, drones and autonomous vehicles have also entered this field in recent years,  taking inspiration from existing research on socially aware mobile robot navigation.
In this paper, we conduct a survey and analyze the literature based on different aspects of socially aware robot navigation, with special emphasis on how the navigation is currently implemented in different types of robots and how these differences affect human perception of comfort and safety (Fig. \ref{main-fig}).

While existing surveys have provided valuable insights, there is a need for a new comprehensive survey that addresses a broader range of robots and contributes with a more inclusive definition of socially aware robot navigation.
Furthermore, our survey analyzes various research areas required to support and advance the field.
These include studies, tools, methods of evaluation, and human trajectory and intention prediction techniques.
Most importantly, we present multi-faceted taxonomy-based classifications for socially aware robot navigation and examine the problem from different angles. Our aim in proposing taxonomies is to help the reader navigate the vast number of contributions and select those that are most relevant to their discipline and objectives. We hope that this classification will not only be pertinent to robot developers but also to application designers and evaluators and act as a basis for interdisciplinary cooperation on the topic.

\par
The remaining of this section introduces essential definitions and presents a preliminary analysis of the literature, including previous surveys in the field.
In section~\ref{sec:class_and_def}, we propose a taxonomy that is used to refine the analysis in section~\ref{sec:robot} (Types of robots), section~\ref{sec:plan} (Planning and Decision-Making), section~\ref{sec:situation} (Situation Awareness and Assessment), and section~\ref{sec:evaluation} (Evaluation methods and tools).
Section~\ref{sec:proposals} puts forward a set of recommendations to enhance socially aware robot navigation. Finally, section~\ref{sec:future} covers prospective challenges in the field.



\subsection{Socially Aware Robot Navigation}
Navigation is the activity whereby an embodied agent (a robot or a person) changes its position in an environment to reach a goal. While navigating, the agent may encounter other agents who are sharing the same environment. This subject has received different names in the robotics community, being the frequently used ones `\textit{human-aware navigation}', `\textit{social navigation}', and `\textit{socially aware navigation}'. `\textit{Human-aware}' is used in the sense that the design and the algorithm need to take into account specifically the presence of the human in the proximity of the robot, their activity, and preferences. No assumption is made in the wording on the fact that the robot acts naturally or socially. `\textit{Social navigation}' means navigation that integrates social rules, protocols, and roles that are generally used by humans when they act or interact with other humans. The risk here is the inability to distinguish, at least in the wording, between human social behavior and robot social behavior and capabilities. `\textit{Socially aware navigation}', while insisting on the social aspect of navigation in the proximity of humans, does not, in the wording, enforce the need for the robot social navigation to be identical to human social navigation. Therefore, we choose to use the term \textit{\textbf{socially aware robot navigation}} in the rest of the paper and call the agents involved in such navigation as \textit{\textbf{social (navigation) agents}}.

\par
To move toward specific definitions, we propose a set of properties for social (robotic) agents. Thus, a robot can be called \textbf{socially aware} if: 
\begin{enumerate}
    \item It detects human agents and treats them as special entities, with their safety as the utmost priority. 
   \item Its behavior is designed to minimize disturbance and discomfort to human agents and cause little to no confusion to them.
    \item It exhibits its navigation intentions, explicitly or implicitly.
    \item In case of a conflict, it assesses the situation and takes the action that it expects will resolve the conflict in a social manner, potentially compromising its own task.\label{it:conflict}
\end{enumerate}
\par

In the above, the term \textbf{human agent} is used to refer not only to individual humans but also to vehicles and robots controlled by them. To satisfy the last property (\ref{it:conflict}, above), the robot requires an understanding of human intentions and negotiation capabilities. However, human intentions are hard to predict as they are often context dependent. Therefore, most of the existing research in socially aware navigation systems focuses on the first three aspects. Although, to the best of our knowledge, there is no explicit consensus, in our opinion, for a robot navigation algorithm to be called socially aware it should at least satisfy the first two properties, which can be seen as the minimal requirements. 

\subsection{Article Collection and Preliminary Analysis}
\begin{figure*}[!th]
    \begin{center}
    \vspace{-1cm}
    \hspace{-1cm}
        \resizebox{1.6\columnwidth}{!}{%
        \input{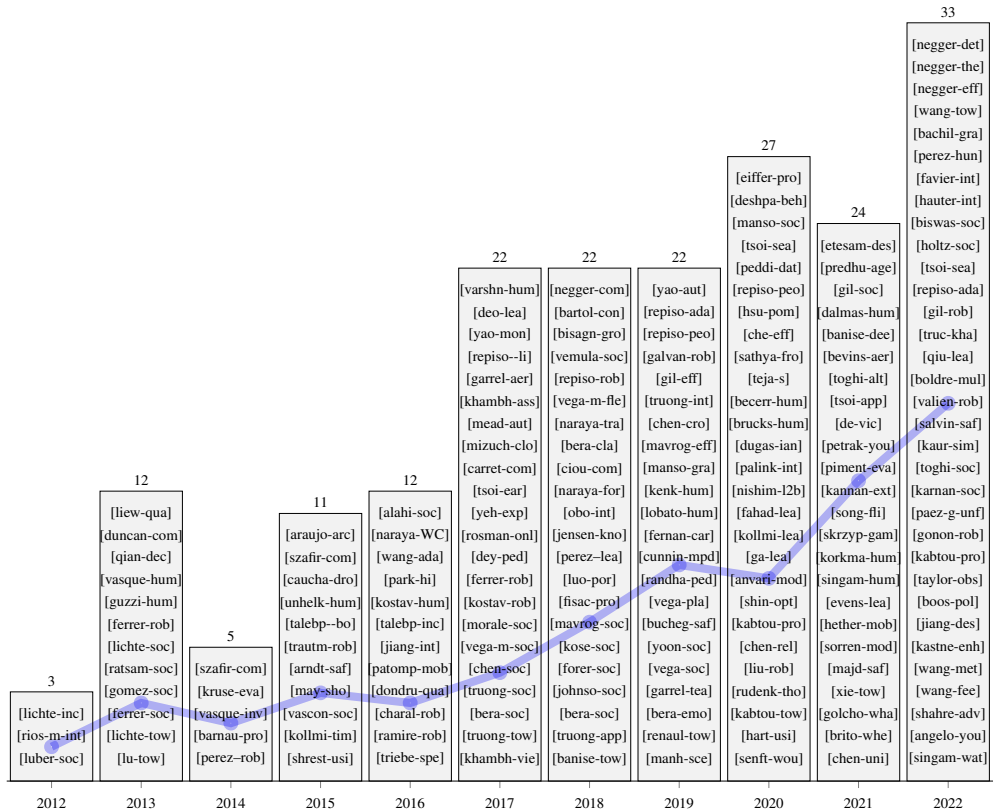}}
        \caption{Distribution of papers in this survey by year. 
        {Yearly trend of publications in IEEE XPlore is shown in blue.}}
        \label{fig:paper_list} 
    \end{center}
    \vspace{-0.5cm}
\end{figure*}

Although socially aware robot navigation has been a topic of research for over 20 years \citep{tadokoro1995motion, wilkes1998toward}, there has been an increasing number of publications over the past 6 years. 
This can be observed in Fig. \ref{fig:paper_list}, where the yearly trend of the number of papers on socially aware navigation in IEEE Xplore is shown as a blue line.
This trend is approximately exponential, showing a fast-growing interest in the field.
The slight decrease in 2020 could be attributed to COVID-19.
We have collected articles from different sources like IEEE Xplore, ACM digital library, and Google Scholar that match the search query, (`social' OR `human-aware') AND `navigation' AND (`robot' OR `autonomous vehicle' OR `drone').
More than 200 articles have been used to write this survey, which are either directly associated with socially aware robot navigation or the supportive literature that is required by the field.
To keep the length of this survey within reasonable limits, a comprehensive review of the collected papers was conducted to select distinctive proposals that would form a representative subset of what has been done in socially aware robot navigation.
For this selection, we have restricted ourselves to the past 10 years since there are previously existing surveys like \cite{kruse_human-aware_2013} that cover most of the papers until 2012.
The papers in the survey at hand and their distribution by year are presented in Fig.~\ref{fig:paper_list}.
Although there are papers addressing socially aware navigation in autonomous vehicles and drones, a large portion of papers correspond to mobile robots, as it has generally been the core area of focus. Since our main goal is taxonomic analysis, we do not think that it suffers from this limitation.

\subsection{Previous Surveys}
The rise of research in socially aware robot navigation has led to an increase in surveys in the field.
In one of the early surveys, \cite{kruse_human-aware_2013} presented diverse approaches used to tackle socially aware robot navigation and the accompanying challenges.

It also briefly covered evaluation methodologies. The survey in \citep{rios-martinez_proxemics_2015} focused on how proxemics has been adapted to perform socially aware robot navigation around individuals and groups of people while taking affordance spaces into account. The review presented in \citep{pol_review_2015} covered different strategies employed for planning, \cite{chik_review_2016} reviewed literature based on different navigation frameworks and their components. A literature survey on the required level of robot perception, mapping, and awareness to properly navigate human environments was provided in \citep{charalampous_recent_2017}. It further provided an integrated framework for analyzing pedestrian behavior in shared spaces and described the limitations of the approaches at the time. Recent works like \cite{ridel_literature_2018} and \cite{rudenko_human_2020} present detailed literature reviews on pedestrian behavior and human motion prediction methodologies. \cite{honig_toward_2018} presented a comprehensive review of person-following robots and the different elements involved in their design and evaluation. In recent years, research on the social aspects of autonomous vehicles (AVs) has started to exploit the vast literature available on vehicle-pedestrian interactions. For example, the survey in \citep{rasouli_autonomous_2020} addresses pedestrian behaviors, communication modalities, and strategies for AVs.
More recent works like \cite{m_predhumeau_a_spalanzani_j_dugdale_pedestrian_2021} study human-robot-vehicle interactions in shared spaces and propose an integrated framework to systematically analyze studies in the field. 
\par

The survey in \citep{mavrogiannis_core_2023} divides the problem into three types of challenges (planning, behavioral, and evaluation) and explains how they are approached in the literature, along with open questions. The work in \citep{moller_survey_2021} views the problem from the perspective of visual understanding and planning to provide a deeper understanding of different aspects of socially aware robot navigation and the available datasets. A list of works in the field is provided in \citep{ngo_recent_2021}.
There are also more focussed surveys like \cite{gao_evaluation_2022} that present different types of evaluation strategies employed in the field. The work in \citep{mirsky_prevention_2021} defines conflicts in socially aware navigation settings and proposes a taxonomy around them to organize the literature. It presents the possible future extensions of the taxonomy and provides a checklist to verify while contributing to the field. 

\par
In this context, we extend the boundaries of existing surveys by providing a comprehensive multi-faceted classification system. 
Although \cite{mirsky_prevention_2021} also provide a taxonomy in their recent survey, our classification covers multiple dimensions of socially aware robot navigation, taking into account the type of robot, planning and decision-making aspects, situation awareness and assessment, as well as evaluation methods and tools.
By adopting this multi-dimensional approach, we aim to offer a holistic view of the field and address the complex interplay of the different factors taking place in socially aware robot navigation.
For example, while surveys that focus on particular types of robots exist (\textit{e.g.}, autonomous vehicles in~\citep{rasouli_autonomous_2020}), ours is the first explicitly considering the different types of robots and how their specific features influence navigation.
\par
Furthermore, we incorporate topics that have not received much attention in previous surveys.
This is the case of contextual information that can be obtained from the environment, the task, and pedestrian intention detection.
Environmental and task contexts are included as part of the planning open problems in~\citep{mavrogiannis_core_2023}, however, we provide a deeper analysis of the related works.
In our review, we extend the literature analysis of these topics through specific branches of the proposed taxonomy.
While \cite{moller_survey_2021} address different contexts and tasks, they emphasize human-robot interaction rather than socially aware navigation. 
Our primary focus is socially aware navigation. 
Additionally, although intentions are analyzed in~\citep{mavrogiannis_core_2023}, the discussion pertains to how robots should communicate their intentions. 
In our survey, we include the detection of human intentions as a specific topic in our situation awareness and assessment taxonomy. 
The communication of robot intentions is also analyzed in the proposed planning and decision-making taxonomy.
\par
Regarding evaluation, most recent surveys analyze in depth the tools used for evaluating socially aware robot navigation proposals~\citep{gao_evaluation_2022, mavrogiannis_core_2023, moller_survey_2021}.
These tools encompass studies, datasets, simulators, and metrics.
We extend the analysis of socially aware navigation evaluation provided in other reviews by classifying the evaluation methods into qualitative and quantitative.
This classification aims to offer an additional perspective on the existing evaluation methodologies, providing deeper insight into the limitations and challenges in evaluating socially aware navigation.
\par
In summary, our survey stands out by offering a multi-dimensional perspective on socially aware navigation, covering aspects often discussed in less depth, and providing an inclusive classification system.
This inclusive approach fosters a more holistic understanding of the field, offering a comprehensive overview of the state of the art in socially aware robot navigation. 
We believe these unique features distinguish our work and establish its significance in complementing existing surveys in the domain.


\section{Proposed Taxonomies}
\label{sec:class_and_def}

We propose multi-faceted taxonomies for classifying and arranging the literature into four distinct aspects related to socially aware navigation. For the classification, we conducted a deductive thematic analysis~\citep{clarke2015thematic}, where a predefined set of themes (taxonomy trees) are defined from the start and the articles fit into them. This section presents the review process and the final taxonomies.

\subsection{Review Process}
The review procedure started with multiple discussions on arranging the articles into a unified classification that can reflect different aspects of socially aware robot navigation. Four different multi-faceted taxonomies were selected, as combining multiple non-overlapping perspectives into a single taxonomy is not ideal.
\par
Once the taxonomies were decided and the nodes were defined, we used a \textit{tag} to represent each node. The \textit{tags} can be seen as the \textit{codes}, and the taxonomies can be seen as the \textit{themes} in thematic analysis. Using these node definitions and \textit{tags}, we reviewed every article collected and made a summary for each. Some articles were discarded during this process, and the remaining ones were assigned multiple \textit{tags} corresponding to one or more of the four taxonomies. 
The exclusion criteria used were: a) the paper refers to socially aware navigation but it is not the core topic, and b) the paper provides only a small incremental contribution over those already covered in the survey (frequently by the same authors).
We did multiple passes on these reviews, which resulted in improved taxonomies and definitions. There were major revisions in some of these taxonomies, which required revisiting the literature and reassigning \textit{tags} according to the new classification. 

At the end of this process, we were left with 193 articles with multiple \textit{tags} that were utilized to classify them across different perspectives. For each article, along with a summary, we included descriptions of the aspects of the work associated with each assigned \textit{tag} to facilitate analysis.

\subsection{Taxonomies Description}
This section provides a detailed description of the proposed taxonomies and their associated taxa: robot type, planning and decision-making, situation awareness and assessment, and evaluation and tools.

\subsubsection{Taxonomy for Robot type:}
With the advancements in delivery, logistics, automation, and service sectors, various types of robots are being deployed in human environments. 
\begin{figure}[h]
\vspace{-0.1cm}
    \centering
    \includegraphics[width=0.98\columnwidth]{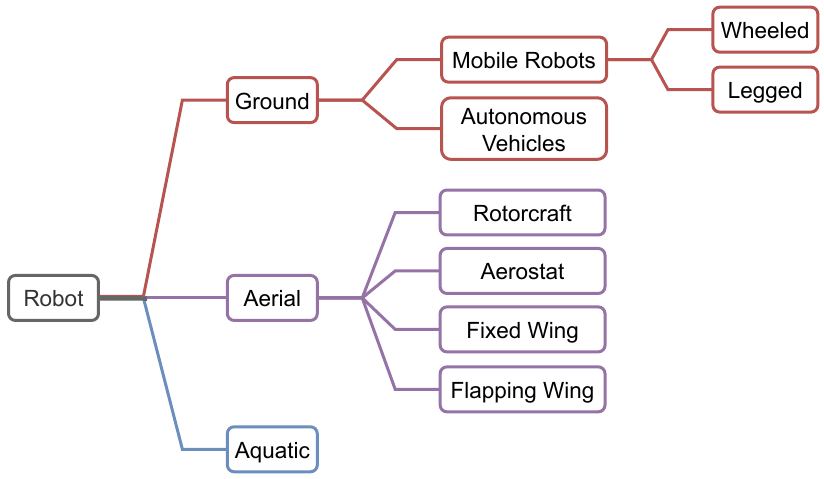}
    \caption{Taxonomy based on robot type.}
    \label{fig:robot_tree}
\end{figure}
Although there are common norms that apply to each type of robot, there are also specific norms that differentiate them. These specificities affect the design of their socially aware navigation strategies. Therefore, we propose our first classification of socially aware navigation papers based on the robot type as shown in Fig.~\ref{fig:robot_tree}. 
In this taxonomy, the facets in the leaf nodes are mutually exclusive for most works.

\subsubsection*{Definitions}
    \begin{enumerate}
        \item \textit{\textbf{Ground}}: This taxon contains all the articles with robots that maintain contact with ground while they move.
        \begin{enumerate}
            \item \textit{\textbf{Mobile Robots}}: A robot that has the capabilities to sense and move in an environment autonomously. They do not carry any human passengers while they move.
            \begin{enumerate}
                \item \textit{\textbf{Wheeled Robots}}: Any kind of mobile robot with wheels (differential, omni, Ackermann, etc.). 
                \item \textit{\textbf{Legged Robots}}: Any kind of legged robot (bi-ped, tri-ped, quadra-ped, etc.).
            \end{enumerate} 
            \item \textit{\textbf{Autonomous Vehicles}}: Autonomous systems that can sense and move among human environments while carrying or transporting human passengers. It includes personal mobility vehicles (PMVs) as well.
        \end{enumerate}
        \item \textit{\textbf{Aerial}}: This taxon consists of the articles with robots that can move or fly in the air without any physical support from the ground. This classification is based on the type of mechanism used to generate the flight \citep{hassanalian2017classifications}. 
        \begin{enumerate}
            \item \textit{\textbf{Rotorcraft}}: The rotors are used to generate the thrust in this of robots. It consists of single rotor systems like helicopters and multi-rotor drones. 
            \item \textit{\textbf{Aerostat}}: The lighter-than-air flying robots that float in the air and use small propelling systems to move around. It contains systems like blimps and hot-air balloons.
            \item \textit{\textbf{Fixed Wing}}: All the fixed wing drones like airplanes and gliders. 
            \item \textit{\textbf{Flapping Wing}}: All the `ornithopter' drones that use bird or insect type wing flapping mechanisms.
        \end{enumerate}
        \item \textit{\textbf{Aquatic}}: This taxon contains all the articles that deal with socially aware navigation in the robots that move on or under the surface of the water. For now, we have not included any further classification as this taxon of robots does not have any works on socially aware navigation yet.
    \end{enumerate}

\subsubsection{Taxonomy for Planning and Decision-Making:}
\begin{figure}[!ht]
     \centering
   \includegraphics[width=0.98\columnwidth]{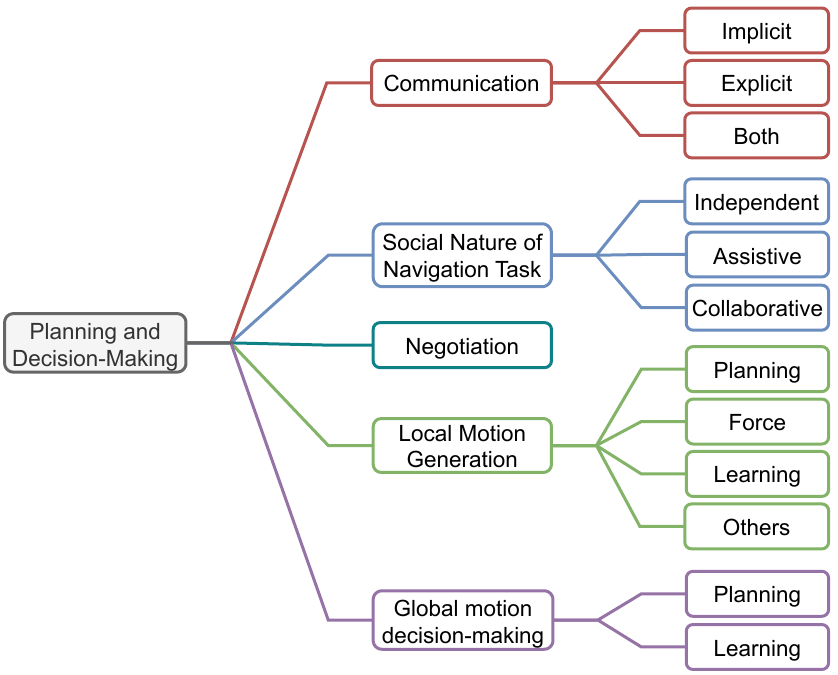}
    \caption{Taxonomy based on planning and decision making.}
    \label{fig:planning_decision-making_tree}
    \vspace{-0.4cm}
\end{figure}

This classification includes planning and motion decision-making, which are core topics in robot navigation. Other characteristics related to different types of decision-making have also been included in this classification, namely types of tasks, communication, and negotiation strategies. Fig.~\ref{fig:planning_decision-making_tree} shows this classification and its sub-divisions. Here, the facets are not mutually exclusive.

\subsubsection*{Definitions}
    \begin{enumerate}
        \item \textit{\textbf{Communication}}: This taxon considers articles that use some form of intentional communication where the robot communicates or responds to humans' signals.
        \begin{enumerate}
            \item \textit{\textbf{Implicit}}: The form of communication where the recipient is expected to infer the message from implicit signals like body motion or posture, force, or gaze.
            \item \textit{\textbf{Explicit}}: The form of communication is through speech, video, or gestures, where agents explicitly convey their intentions. 
            \item \textit{\textbf{Both}}: Strategies that use a mixture of implicit and explicit communication forms.
        \end{enumerate}
        \item \textit{\textbf{Types of Navigation Task}}: This taxon classifies the works based on robots' and humans' roles in the navigation task.
        \begin{enumerate}
            \item \textit{\textbf{Independent}}: 
            Tasks where the robot performs socially aware navigation and is not tightly bound to any human (e.g., crowd navigation, delivery). The pedestrians are treated as social dynamic obstacles, but no interaction occurs.
            
            \item \textit{\textbf{Assistive}}: The navigation task where a robot or a vehicle
            provides assistance or support to one or more people. Assistance can be provided in several ways, like following or accompanying a person, or taking the shape of transportation services (e.g., pushing a wheelchair or running a shuttle).
            
            \item \textit{\textbf{Collaborative}}: A robot and human agent working together to coordinate and successfully navigate through complex environments, such as narrow hallways or doorways, where cooperative effort and coordination are needed to reach the desired destination.
            
        \end{enumerate}
        \item \textit{\textbf{Negotiation}}: This taxon considers articles that adapt the robot's navigation based on some form of dynamic information exchange (e.g., asking for permission to pass, different forms of inducement).
        \item \textit{\textbf{Local Motion Generation}}: This taxon includes articles that present methodologies or improvements for lower-level motion generation like trajectory or velocity commands.
        \begin{enumerate}
            \item \textit{\textbf{Planning}}: Methodologies that rely on trajectory generation or forward simulations for getting the robot's command velocity (e.g., DWA, MPC, Elastic Bands).  
            \item \textit{\textbf{Force}}: Methodologies that rely on potential fields and object forces to generate velocity command for the robot (e.g., Social Force Model, Artificial Potential Fields).
            \item \textit{\textbf{Learning}}: Methodologies that use data and/or learn models to generate the velocity command directly from the observations (or input).
            \item \textit{\textbf{Others}}: Any other methodology that cannot be fit into the above strategies.
        \end{enumerate}
        \item \textit{\textbf{Global Motion Decision-Making}}: This taxon includes the articles that use a global representation to generate a decision and/or path to assist motion generation.
        \begin{enumerate}
            \item \textit{\textbf{Planning}}: Methodologies that use geometric or formal planning approaches.  
            \item \textit{\textbf{Learning}}: Methodologies that are data-driven and/or use learned models.  
        \end{enumerate}
    \end{enumerate}

\subsubsection{Taxonomy for Situation Awareness and Assessment:}
\begin{figure}[!h]
\vspace{-0.5cm}
     \centering
    \includegraphics[width=0.89\columnwidth]{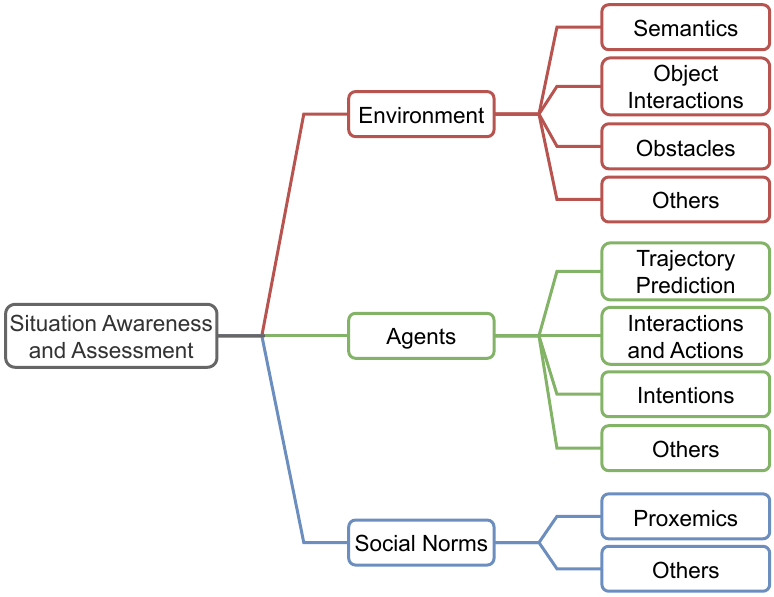}
  \caption{Taxonomy based on situation awareness and assessment.}
    \label{fig:situation_awareness_tree}
    \vspace{-1mm}
\end{figure}

This classification covers situation awareness following the definition by~\cite{ensley1995toward}: ``\textit{the perception of the elements in the environment within a volume of time and space, the comprehension of their meaning, and the projection of their status in the near future}''. 
Because the meaning of the term `{comprehension}' is debatable when it comes to robots, this classification will mainly focus on the representation and prediction of the state of \textit{agents} and other items that are modeled for the purpose of socially aware navigation. This includes other aspects related to physical elements of the \textit{environment} and non-tangible elements involved in socially aware navigation that may affect decision-making, like the \textit{social norms}. Fig.~\ref{fig:situation_awareness_tree} shows the main taxa and the different branches that arise from them. 
\subsubsection*{Definitions}
    \begin{enumerate}
        \item \textit{\textbf{Environment}}: This taxon considers aspects related to the physical space in which the robot navigates. Collective issues such as the density of humans are also considered within this.
        \begin{enumerate}
            \item \textit{\textbf{Semantics}}: Approaches that consider information related to the type or purpose of the area where socially aware navigation takes place.
            \item \textit{\textbf{Object Interactions}}: Approaches that consider human-object or robot-object relations.
            \item \textit{\textbf{Obstacles}}: The approaches that represent the area of the space that is not available for navigation, regardless of whether the representation is purely metric (\textit{e.g.}, occupancy grids), symbolic, or hybrid.
            \item \textit{\textbf{Others}}: Any other aspect of the environment apart from those mentioned above. 
        \end{enumerate}
        \item \textit{\textbf{Agents}}: This taxon describes how are agents represented in the articles, if any. Although the definition does not explicitly restrict the concept of agents to humans, in practice, they are the only external agents found in the literature, except for the case of autonomous vehicles.
        \begin{enumerate}
            \item \textit{\textbf{Trajectory Prediction}}: Approaches using future human pose estimations.
            \item \textit{Interactions and Actions}: Approaches considering representation and usage of actions, as well as human-human and human-robot interactions.
            \item \textit{\textbf{Intentions}}: Approaches that use or detect agent's intention for socially aware navigation.
            \item \textit{\textbf{Others}}: For other aspects of the agents that may be exploited for the navigation.
        \end{enumerate}
        \item \textit{\textbf{Social Norms}}: This taxon includes the articles that discuss the aspects related to the comfort, safety and humans' preferences. 
        \begin{enumerate}
            \item \textit{\textbf{Proxemics}}: Approaches considering social distances rather than just collision avoidance.
            \item \textit{\textbf{Others}}: Approaches including other frequently used conventions during human navigation (\textit{e.g.}, walking on the right or left-hand side).
        \end{enumerate}
    \end{enumerate}

\subsubsection{Taxonomy for Evaluation and Tools:}
\begin{figure}[H]
\vspace{-0.1cm}
    \centering
    \includegraphics[width=0.85\columnwidth]{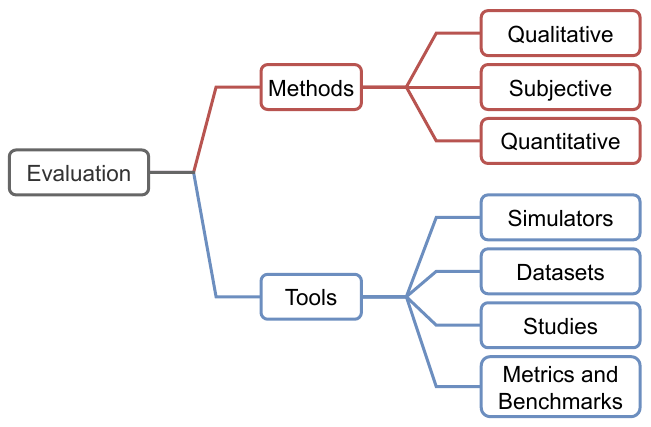}
    \caption{Taxonomy for tools and evaluation methods.}
    \label{fig:evaluation_tree}
    \vspace{-0.5cm}
\end{figure}

This classification considers the strategies employed for the evaluation of the socially aware navigation schemes. Various types of evaluation methodologies that are employed to assess the robot's behavior and the tools that support or are required for the evaluation are included in this classification. The taxa of this classification are shown in Fig.~\ref{fig:evaluation_tree}. 

\subsubsection*{Definitions}
    \begin{enumerate}
        \item \textit{\textbf{Methods}}: This taxon contains the articles that have some form of evaluation of socially aware navigation.
        \begin{enumerate}
            \item \textit{\textbf{Qualitative}}: Methods that use numerical/non-numerical data and use subjective or comparative analysis for evaluation.
            \item \textit{\textbf{Quantitative}}: Methods that use numerical data and objective analysis (based on metrics or benchmarks) for evaluation.
        \end{enumerate}
        \item \textit{\textbf{Tools}}: This taxon contains the articles that provide or propose tools for advancement and evaluation of socially aware navigation.
        \begin{enumerate}
            \item \textit{\textbf{Simulators}}: Articles that propose new simulators or strategies to improve the human-robot navigational interaction in simulation.
            \item \textit{\textbf{Datasets}}: Articles that propose new datasets that can advance socially aware navigation. This could be in the form of human-robot navigational data or rich human-human interaction data.
            \item \textit{\textbf{Studies}}: Articles containing user studies in wild or controlled spaces that analyze human-robot interaction which could be employed to improve socially aware navigation.
            \item \textit{\textbf{Metrics and Benchmarks}}: Articles that propose new metrics or benchmarks.
        \end{enumerate}
    \end{enumerate}

\section{Types of robots}
\label{sec:robot}
\begin{figure}[!h]
    \centering
   \includegraphics[width=0.98\columnwidth]{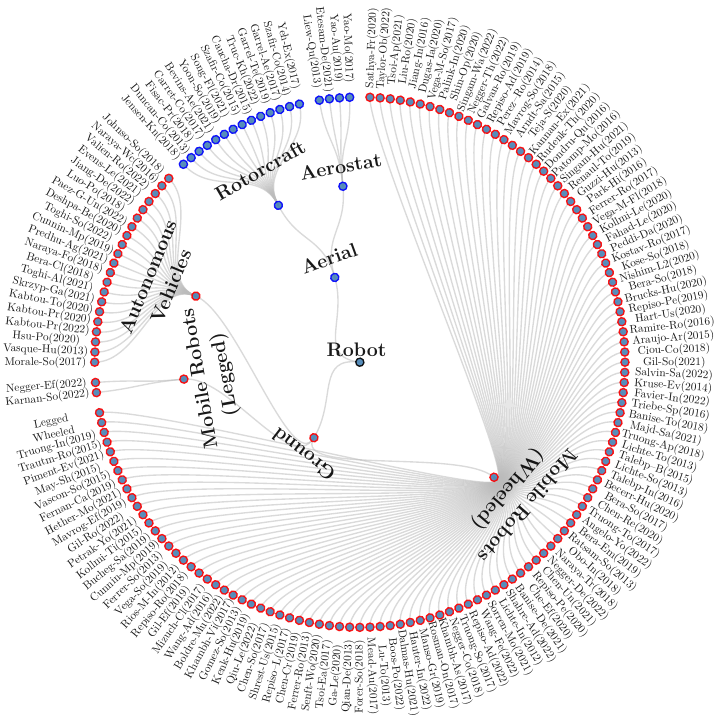}
    \caption{Distribution of papers by Robot type. The figure is best viewed zoomed in using a digital version.}
    \label{fig:robot_list}
\end{figure}

More than 80\% of the articles in this survey use some kind of robot either physically or in simulation to implement or test a socially aware navigation scheme, study the interactions or collect data.
Based on the proposed classification, 156 papers are distributed among various types of robots. As it can be seen from Fig.~\ref{fig:robot_list}, a large portion of papers (116) fall under the \textit{mobile robots} taxon, and the rest are distributed between \textit{autonomous vehicles} (22) and \textit{aerial robots} (18). Only one paper by \cite{cunningham_mpdm_2019} applies their navigation scheme to a mobile robot and an autonomous car.
Aerial robots and autonomous vehicles recently started exploring the idea of socially aware navigation and there are preliminary works that use the term \textbf{‘social’} or \textbf{‘human-aware’} or `\textbf{socially aware}'. 
The navigation of autonomous wheelchairs has been a research topic for quite some time, but the field has not been as active in recent years \citep{sivakanthan2022mini}.
We have not found in the literature any paper dealing with socially aware navigation for sea or underwater robots.
Thus, this section focuses only on ground and aerial robots.

\subsection{Ground Robots}
Ground robots taxon encompasses a wide variety of platforms like mobile humanoid robots, simple mobile bases, wheelchairs, autonomous cars, delivery pods, legged robots, etc., that majorly operate on the ground. Although most of the works presented in this survey are based on wheeled robots (or vehicles), this taxon does not discard the possibility of having socially aware navigation using legged robots. For instance, \cite{karnan_socially_2022} uses Boston Dynamics’ Spot to build the dataset for socially-aware navigation. \cite{neggers_effect_2022} uses bi-pedal NAO robot for a user study. Robots like Spot and Cheetah already have good controllers for mobility \citep{di2018dynamic, zimmermann2021go} and we expect to see more works dealing with socially aware navigation using these robots or other kinds like bi-peds.

socially aware robot navigation originated as a part of human-robot interaction (HRI) research \citep{singamaneni2022combining}.
Architectures were developed to deploy an interactive robot among humans and navigation remained a challenging task from the very beginning.
The initial works on socially aware navigation always consisted of a higher-order task to accomplish like guidance or assistance.
Due to this, many of the works on socially aware navigation use \textbf{mobile humanoid robots} \citep{singamaneni_human-aware_2021, teja_s_hateb-2_2020, hauterville_interactive_2022, ferrer_robot_2013} that have the appearance of a humanoid but use wheels instead of legs to move.
As time progressed, socially aware navigation became a field on its own rather than a mere requirement for other tasks.

\begin{figure}[h]
    \centering
    \includegraphics[width=0.8\columnwidth]{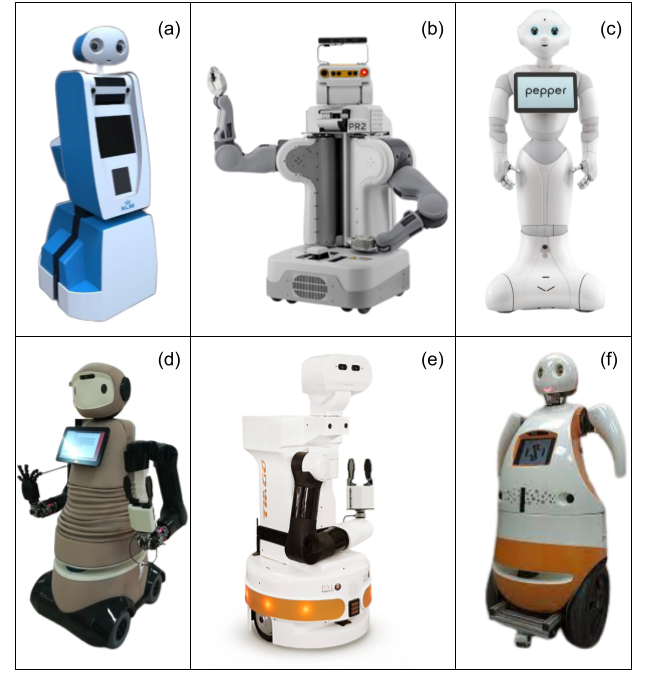}
    \caption{Mobile Robots: (a) Spencer \citep{triebel_spencer_2016}, (b) PR2 \citep{singamaneni_human-aware_2021}, (c) Pepper \citep{angelopoulos_you_2022}, (d) Ivo \citep{gil_social_2021}, (e) Tiago \citep{hauterville_interactive_2022} and (f) Tibi \citep{ferrer_robot_2013}.}
    \label{mobile_robots}
\end{figure}

Some of the mobile humanoid robots that are used by the research community are shown in Fig.~\ref{mobile_robots}.
Although the PR2 robot is relatively old, it is still being used by many researchers \citep{khambhaita_assessing_2017, mead_autonomous_2017, kruse_evaluating_2014, teja_s_hateb-2_2020, singamaneni_human-aware_2021, ramirez_robots_2016, lu_towards_2013, khambhaita_viewing_2017, forer_socially-aware_2018, singamaneni_watch_2022} because of its robust hardware, multiple high fidelity sensors for perception and open source support of the platform. The other frequently used humanoid robot for socially aware navigation is Pepper, which is commercially available and closer to human in appearance~\citep{teja_s_hateb-2_2020, singamaneni_human-aware_2021, dugas_ian_2020, bera_emotionally_2019, randhavane_pedestrian_2019, bera_socially_2018, angelopoulos_you_2022}.
Pepper has tactile sensors and multiple language support for human-robot interaction and is often used to investigate short-range navigation tasks near humans. Tiago is a more recent commercial humanoid robot that is being used by the robot navigation community \citep{macenski2020marathon2}. This robot is built on an open-source platform which allows the user to make modifications to the packages as required. \cite{hauterville_interactive_2022} uses this robot to test a socially aware navigation stack in Gazebo. 
From time to time, specialized robots like Tibi \citep{ferrer_robot_2013}, IVO \citep{gil_social_2021}, Robovie \citep{anvari_modelling_2020, senft_would_2020} or SPENCER \citep{triebel_spencer_2016} are built to have more customizability and to address specific needs of the researchers. These robots allow the user to modify the components or the software stacks easier in comparison to commercial robots.
\par
\begin{figure}[h]
    \centering
    \includegraphics[width=0.9\columnwidth]{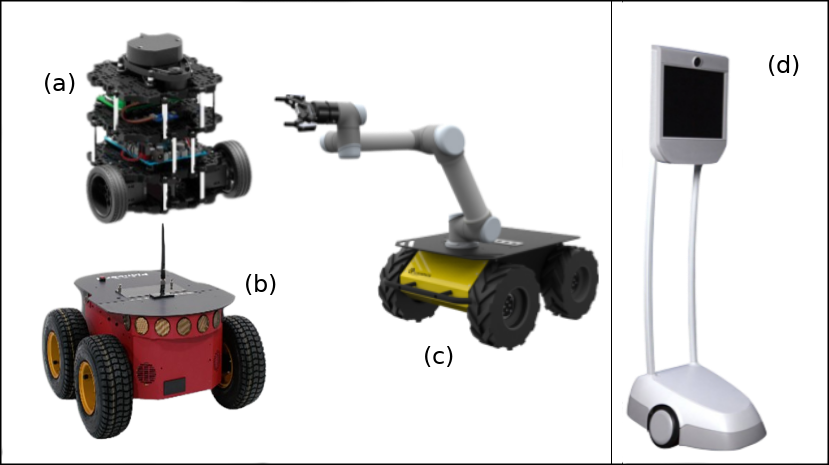}
    \caption{Mobile Bases: (a) Turtlebot \citep{kollmitz_time_2015}, (b) Pioneer 3-DX \citep{chen_relational_2020} , (c) Husky \citep{hart_using_2020}, and (d) Beam Pro Robot \citep{mavrogiannis_effects_2019}.}
    \label{mobile_bases}
\end{figure}

\par
Given that arms are not strictly necessary to perform socially aware navigation, most works just use \textbf{mobile bases} \citep{boldrer_multi-agent_2022, vasconcelos_socially_2015, truong_socially_2017, liu_robot_2020, chen_socially_2017, charalampous_robot_2016, ga_learning_2020, chen_relational_2020, banisetty_deep_2021, sathyamoorthy_frozone_2020, jiang_interactive_2016, chen_unified_2021,wang_feedback_2022}. They are typically fitted with proximity sensors and/or LiDARs to detect and avoid obstacles, which are used by socially aware navigation researchers to detect humans using leg detection. Some of the commonly used mobile bases are shown in Fig.~\ref{mobile_bases}. Among the articles collected in this survey, we found that the Pioneer 3-DX is used by many works \citep{chen_relational_2020, buchegger_safe_2019, trautman_robot_2015, banisetty_deep_2021, ciou_composite_2018, fahad_learning_2020, rosmann_online_2017, shahrezaie_advancing_2022} followed by different versions of Turtlebot \citep{che_efficient_2020, sathyamoorthy_frozone_2020, jiang_interactive_2016, qiu_learning_2022, kostavelis_robots_2017, kollmitz_time_2015}.
There are also other commercial platforms like ClearPath Jackal and MiR100 that are used in crowd navigation \citep{liu_robot_2020, chen_socially_2017, hart_using_2020, karnan_socially_2022} and warehouse navigation \citep{ga_learning_2020} respectively. Custom-built robot bases always offer more personalization compared to commercial robots and some works in this survey like \cite{lichtenthaler_towards_2013, charalampous_robot_2016, arndt_safe_2015, truong_approach_2018} and \cite{shrestha_using_2015} use these personalized robots to test their frameworks.

\begin{figure}[h]
    \centering
    \includegraphics[width=0.9\columnwidth]{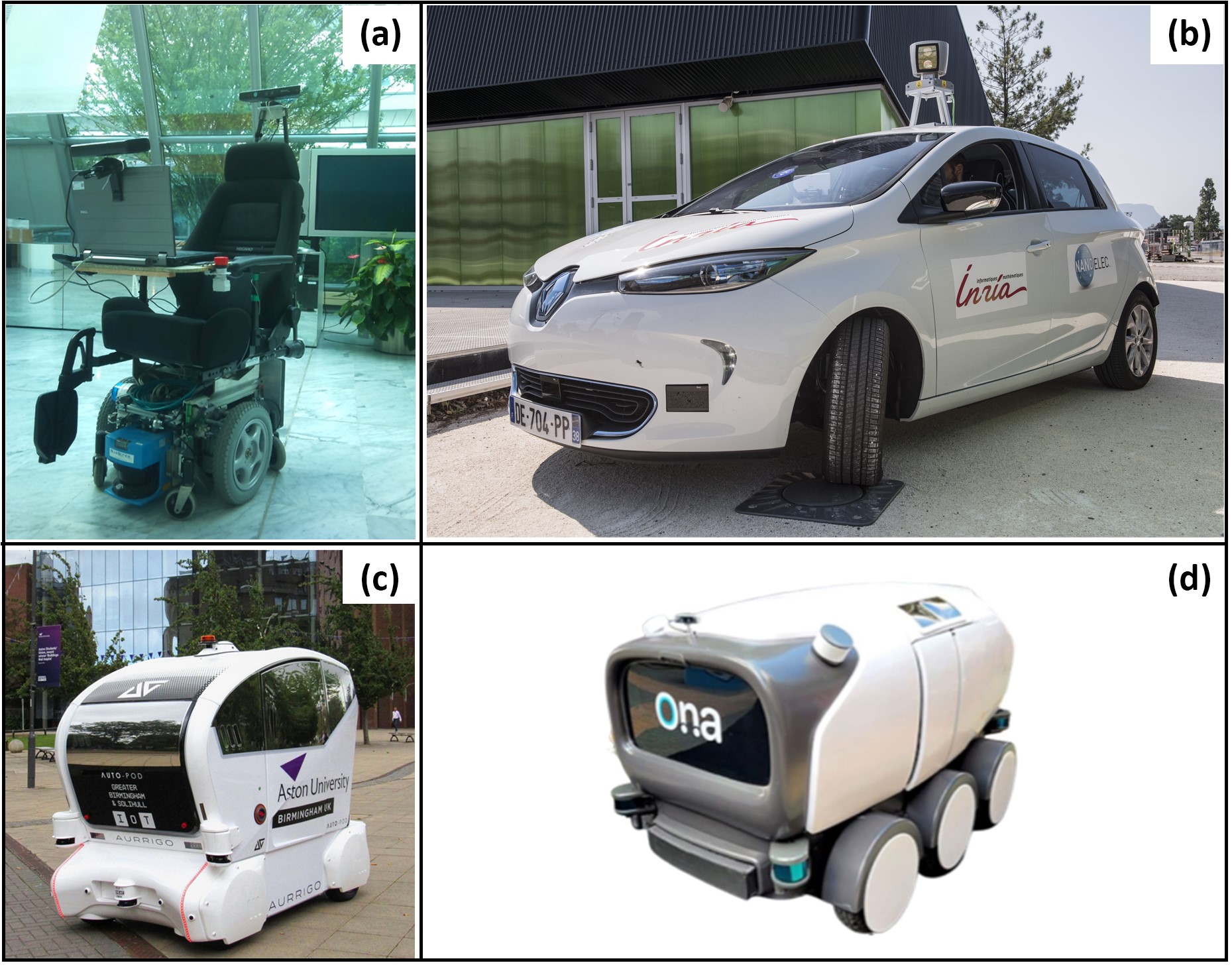}
    \caption{Autonomous vehicles and Wheelchairs: (a) Wheelchair from \citep{rios-martinez_intention_2012} (b) Autonomous Car at INRIA (c) Aurrigo Auto-Pod{\protect\footnotemark} (d) CARNET Ona Robot {\citep{puig-pey2023}}.}
    \label{avs_cars}
\end{figure}

Sometimes the mobile bases are accompanied by screens to display signs or emulate expressions \citep{mavrogiannis_effects_2019, sorrentino_modeling_2021, hart_using_2020}. These robots are either custom-built \citep{hart_using_2020, araujo_architecture_2015, pimentel_evaluation_2021, truong_approach_2018} or chosen from the available ones in the market \citep{qian_decision-theoretical_2013, mavrogiannis_effects_2019}. The idea behind the extra screen is to make the robot more human-friendly by displaying faces and communicating intentions. For instance, \cite{hart_using_2020} use the additional screen to display a face to study the communication strategies using gaze, whereas \cite{mavrogiannis_effects_2019} uses it to just display a smiley face while studying different navigation algorithms. In general, simple mobile bases are used in works that do not include any kind of explicit communication strategies and rely only on implicit cues \citep{che_efficient_2020, kannan_external_2021, hetherington_mobile_2021, palinko_intention_2020} whereas screens or signal lights (or LEDs) are attached for explicit conveyance in some cases \citep{may_show_2015, mavrogiannis_effects_2019, hart_using_2020, palinko_intention_2020}.
When mobile robots are equipped with moving heads, it is also possible to use head movements to communicate intention and attention \citep{khambhaita2016head}.
Wearable communication devices are a new addition and \cite{che_efficient_2020} use a haptic device along with vision and audio for explicit communication. 
Urban \textbf{delivery robots} are the best examples where socially aware navigation and good communication strategies are essential. These are discussed in the works by \cite{kannan_external_2021, boos_polite_2022} and one such robot is shown in Fig.~\ref{avs_cars} (d).

\footnotetext{\url{https://aurrigo.com/autopod/}}

socially aware \textbf{autonomous vehicle} navigation (Fig. \ref{avs_cars} (b)) is a relatively new topic. When autonomous vehicles such as \textbf{cars and shuttles} share their environment with pedestrians, the interaction between them needs to be understood and modeled. This requires additional hardware, new designs, and protocols that are different from those of mobile robots.
Recently, there has been a growing interest in pedestrian trajectory or crowd behavior modeling when they are close to an autonomous vehicle \citep{predhumeau_agent-based_2021, m_predhumeau_a_spalanzani_j_dugdale_pedestrian_2021, song_human_2018, kabtoul_towards_2020, hsu_pomdp_2020, deo_learning_2017}.
Some works have explored this in pedestrian-aware navigation \citep{luo_porca_2018, cunningham_mpdm_2019, randhavane_pedestrian_2019, kabtoul_proactive_2020, hsu_pomdp_2020, kabtoul_proactivesmooth_2022}.
Autonomous cars can also cooperate with other human drivers in traffic, and this is yet another research area that came into existence recently, and can be considered as a part of socially aware navigation.
The works by \cite{evens_learning_2021, toghi_altruistic_2021, valiente_robustness_2022} focus on this issue particularly. \textbf{Personal mobility vehicles (PMVs)} is an umbrella term for a wide range of devices including cars, shuttles, wheelchairs, Segways, scooters, etc., that can carry one or more persons and usually move at lower speeds among shared human spaces. Fig.~\ref{avs_cars} (a, c) shows some pictures of PMVs used by the researchers. Although autonomous \textbf{wheelchair} navigation research has slowed down gradually \citep{sivakanthan2022mini}, there are relevant papers that study autonomous or semi-autonomous navigation in wheelchairs taking social norms into account \citep{rios-martinez_intention_2012, vasquez_human_2013, narayanan_WC_2016, morales_social_2017, johnson_socially-aware_2018, skrzypczyk_game_2021}. Other kinds of PMVs like \textbf{Segways and scooters} were also used for developing socially aware navigation strategies to move among the pedestrians \citep{luo_porca_2018, chen_crowd-robot_2019, paez-granados_unfreezing_2022}. 

\subsection{Aerial Robots}
This taxon considers various kinds of \textbf{unmanned aerial robots or vehicles} that can be used for deliveries, construction, signaling, etc. A majority of the works in this survey fall under the \textit{rotorcraft} taxon, specifically, \textbf{multi-rotor} drones. The vast availability, the ease of use and the precise control of these systems might be a reason for this. Although the drones in the other taxa like \textit{aerostat} were explored in socially aware navigation from time-to-time, \textit{winged drones} (fixed and flapping) were not explored much.


\textbf{Drones} are a recent addition to the field of socially aware navigation and pose a contrasting set of challenges compared to mobile robots. This sparked a several studies using drones to investigate proxemics \citep{duncan_comfortable_2013, yeh_exploring_2017} and communication strategies \citep{bevins_aerial_2021, szafir_communicating_2015, szafir_communication_2014, cauchard_drone_2015, jensen_knowing_2018, yao_autonomous_2019} to enable their deployment in the real world.
Regarding communication, some of these works include additional hardware like LEDs \citep{szafir_communicating_2015} to mimic traffic signals while some others study gestures and flight paths. The noise and wind generated by multi-rotor drones \citep{cauchard_drone_2015} coupled with the lack of familiarity often affect the humans' perception of safety, comfort, and reliability - compelling new designs, studies, and ways to integrate drones in a better way around humans. The study by \cite{liew_quadrotor_2013} suggested that a blimp might be better for socially aware navigation compared to a multi-rotor drone as they are quieter. Recently, \cite{etesami_design_2021} investigated the design of a social blimp and found that people felt safer and comfortable around the blimp.

The current socially aware navigation planning for drones tries to transfer knowledge from the mobile robot and make suitable adjustments. For instance, \cite{truc_khaos_2022} utilizes the cost functions from the social mobile robot navigation for planning socially compliant trajectories for flying robots. A series of works on the aerial social force model by \cite{garrell_aerial_2017, carretero_comfort-oriented_2017, garrell_teaching_2019} modified the classical social force model to define a 3D social force and applied it to quad-rotors in simulation and the physical world. Unlike these, the approach by \cite{yoon_socially_2019} proposes a hidden Markov model-based social trajectory generation using a learned model of the human.
The works by \cite{yao_monocular_2017, yao_autonomous_2019} used a blimp drone to follow humans indoors while reacting to human gestures.
We expect to see more works on socially aware navigation in drones and other kinds of aerial robots in the near future that will populate this taxon.

\section{Planning and Decision-Making}
\label{sec:plan}


In this section, we have included not only classical planning and decision-making techniques, local motion generation, and global motion techniques, but also other aspects that affect planning and decision-making: communication, negotiation, and the type of navigation task (collaborative, assistive, or independent). Local motion generation uses sensing and perception to create trajectory or velocity commands that guide the robot. The global motion process requires employing an extensive spatial representation to provide a command (or decision) that directs the robot's motion. 

While a robot performs socially aware navigation, it may move through crowded pedestrian regions autonomously or take on the role of guiding or escorting one or more people. All of these situations demand planning and decision-making techniques that appropriately take into account the existence of bystanders and/or those in need of aid. In the field of HRI, the development of two-way communication between humans and robots is essential for both cooperative and autonomous robot navigation in a pedestrian environment with varying motion patterns. Additionally, bidirectional negotiation is a crucial part of the built-in planning and decision-making processes. It is worth mentioning that differentiating papers based on the type of task provides a useful framework for understanding the current state and the challenges that remain in socially aware navigation research. 

\begin{figure*}[!hb]
    \centering
    \subfloat{
          \includegraphics[width=0.95\columnwidth]{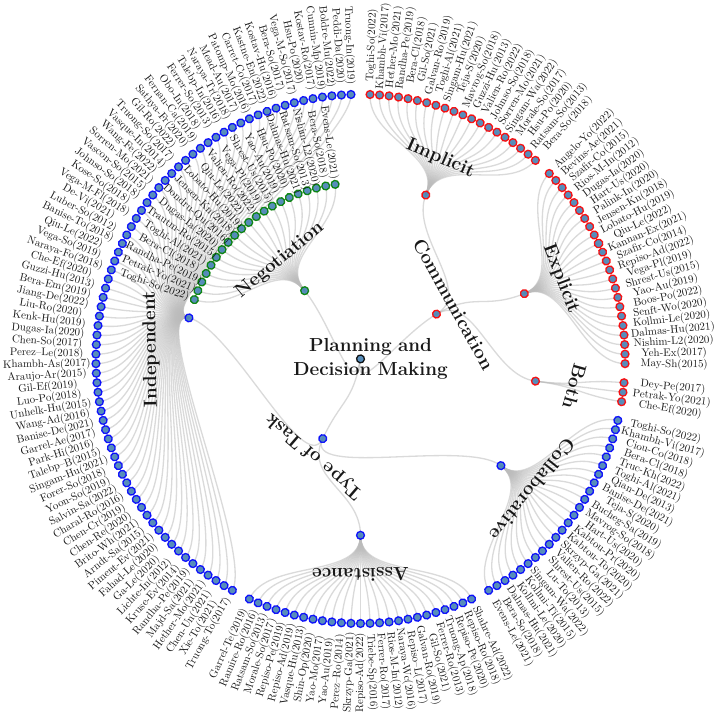}
    } 
    ~
    \subfloat{
          \includegraphics[width=1.05\columnwidth]{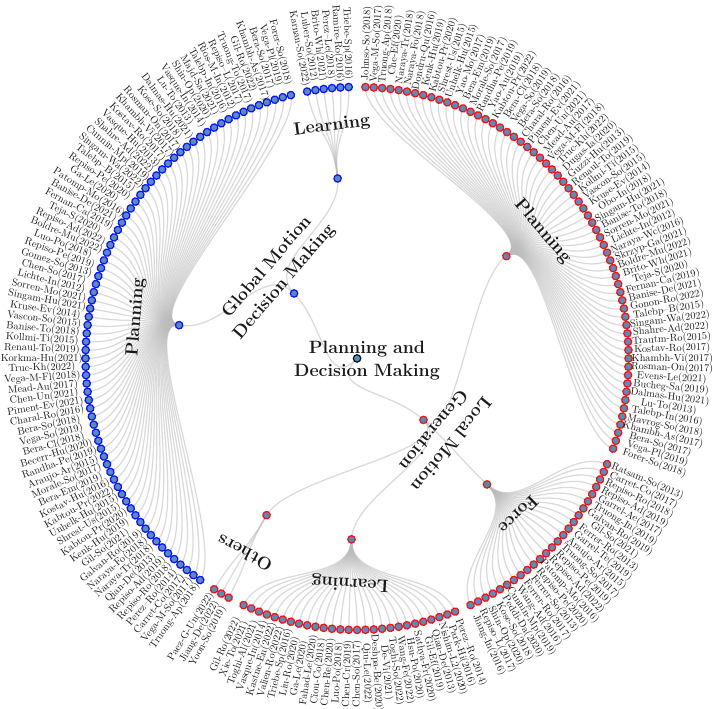}
    }
    
    \caption{Distribution of papers by planning and decision making. The figures are best viewed zoomed in using a digital version.}
    \label{fig:planning_list}
\end{figure*}

Based on this classification, all the methodological papers specific to socially aware navigation (149) matched some of the planning and decision-making criteria. The distribution is illustrated in Figure \ref{fig:planning_list}.

\subsection{Communication}
The study of effective human-robot \textbf{communication} is a fast-expanding topic in HRI in general, as well as in socially aware navigation specifically.
For their seamless integration, robots or humans -- for example, pedestrians or accompanied people -- must understand and respond to the communication signals of the other agent. In the case of robots, they have to grasp pedestrians' intentions and let others know about their own. Humans also have to understand the robot’s intentions and tell about their intentions. It will be always a bidirectional communication that will help to improve safety, efficiency, and comfort.
Although research in this area is especially important to navigate in complex and crowded environments, communication can enhance the overall user experience even in less challenging environments \citep{senft_would_2020,hetherington_mobile_2021}.

\par
On the one hand, \textbf{explicit communication} between robots and humans is critical to facilitate successful decision-making in socially aware navigation. Robots must be able to comprehend and react to the explicit communication of humans, such as spoken language and written instructions.
Simultaneously, robots should be able to communicate their own goals and choices to humans in a way that is simple to understand.
Explicit robot communication strategies employed by roboticists include \textit{verbal} (speech) and \textit{visual} (display or video, gestures) communications. For instance, \cite{yeh_exploring_2017} present a visual approach using a custom-designed social drone  with a social shape, face, and voice for human interaction. The work by \cite{kannan_external_2021} studies visual robot communication using words, symbols, and lights while \cite{palinko_intention_2020} use lights along with gestures to convey the robot's intention.
The work in \citep{rios-martinez_intention_2012} studies gestures in the context of a robotic wheelchair, and integrates a technique to interpret user intentions using head movements into a socially aware motion policy.
Further, in \citep{jensen_knowing_2018}, three studies on drones' gestures to acknowledge human presence and clarify suitable acknowledging distances are presented.
\cite{yao_autonomous_2019} use the human gestures to understand their intentions and provide feedback through LED display.
In the case of mobile robots as well, gestures are explored to show the robot's intentions while crossing corridors \citep{hart_using_2020, senft_would_2020, angelopoulos_you_2022} or taking turns \citep{may_show_2015, palinko_intention_2020}. Lastly, verbal communication is used by \cite{dugas_ian_2020} and \cite{boos_polite_2022} to study the effectiveness of robot's speech in clearing its way while \cite{repiso_adaptive_2022} use it for interaction. 

Explicit communication is especially important in decision-making situations, where clear and accurate communication is necessary for both parties to make informed choices \citep{nishimura_l2b_2020,vega_planning_2019,may_show_2015, kollmitz_learning_2020}. Research in this area has the potential to significantly advance the capabilities of robots in a variety of settings, including aerial and ground robots.
Enhancing robots' comprehension and use of explicit communication to increase their effectiveness when engaging with people has also been studied. For instance, External Human-Machine Interfaces (eHMI) \citep{kannan_external_2021} were enhanced to convey intents to humans; \cite{szafir_communicating_2015} explored the design space regarding explicit robot communication of ﬂight intentions to nearby viewers; and \cite{dalmasso_human-robot_2021} created a new interface where robots and humans can communicate to perform collaborative tasks.
\par

On the other hand, \textbf{implicit communication} is frequently considered a more natural  exchange of information.
Implementing implicit communication in                          HRI, however, can also present some difficulties because it might be more challenging for robots to correctly decipher and react to complex cues and context.
This may result in misinterpretations or communication mistakes that may reduce the effectiveness of the encounter. \cite{teja_s_hateb-2_2020}  proposed a new framework combining decision-making and planning in the human-robot co-navigation scenario to address such misinterpretations and exhibit pro-actively a proposed solution for a navigation conflict. 
This is done by introducing different modalities of planning and shifting between them based on the situation at hand.
However, the study of implicit communication in this field is still an underdeveloped area of research, with relatively few papers addressing the subject like \citep{singamaneni_human-aware_2021,mavrogiannis_social_2018, hsu_pomdp_2020, khambhaita_viewing_2017} and \citep{ hetherington_mobile_2021}. 
Further, it can be difficult to use implicit communication for complex human-robot interactions involving decision-making \citep{repiso_peoples_2020}. 

Despite these challenges, the study of implicit communication between robots and humans in the context of decision-making in socially aware navigation is a valuable and important area of research. \cite{repiso_adaptive_2019,repiso_robot_2018} presented some work where robots became more ubiquitous in society, and they were increasingly being used in HRI scenarios, concretely, in socially aware navigation. The works by \cite{khambhaita_viewing_2017} and \cite{singamaneni_human-aware_2021} use early intention-show of robot as implicit communication in corridor crossing. \cite{hetherington_mobile_2021} study different implicit communication strategies to convey robot's yielding intentions in a door-crossing setting.

\par
 While both implicit and explicit communication between robots and humans play important roles in facilitating successful decision-making in socially aware navigation, the study of both forms of communication in this context is limited, for instance \citep{dey_pedestrian_2017, petrak_you_2021} and \citep{che_efficient_2020}.
Instead of looking at how implicit and explicit communication interact, many academics have concentrated on one of the two. Robots might also not be able to express nonverbal signs in the same manner that people can, such as showing empathy or worry through body language and facial expressions. Establishing trust and rapport between the robot and the human might be challenging as a result, adding to the complexity of decision-making. This underlines how crucial it is to create efficient communication plans that consider the special capabilities and constraints of robots to promote successful interactions.

\subsection{Types of  Navigation Task}

socially aware navigation tasks can be broadly classified into three types: \textit{independent}, \textit{assistive}, and \textit{collaborative}, based on the interaction between people and the robot while the navigation task is being accomplished. It has to be noted that human environments are highly dynamic. Depending on the context and the evolution of a situation, navigation tasks may change.
For instance, in narrow passages or corridors, it may happen that an independent navigation task may call for a collaborative solution. 
\par

\textbf{Collaborative socially aware robot navigation} is especially difficult since the robot and humans must share the same goal and they aim to navigate along.
This means that robots must be able to model other agents in the environment \citep{kollmitz_time_2015}, to include their intentions \citep{kabtoul_towards_2020}, and to adapt their behavior \citep{shrestha_using_2015, ciou_composite_2018}.
In such tasks, robots need to communicate and cooperate with humans to accomplish a specific navigation task \citep{luber_socially-aware_2012}.
For example, the works by \citep{khambhaita_viewing_2017, truong_approach_2018,bera_classifying_2018, singamaneni_human-aware_2021, kollmitz_learning_2020} address the challenge of humans and robots navigating together sharing the same plan.

\par

\textbf{Collaborative robot navigation} can be effectively categorized into two major approaches: \textit{rule-based methods} \citep{parhi2010heuristic, bayazit2004better} and \textit{learning methods }\citep{wang2018facilitating, marge2017exploring}. Rule-based methods involve the formulation of a predefined set of guidelines and regulations that the robot rigorously follows to execute its intended navigation behavior. These rules are often crafted based on a priori knowledge and established norms, providing a structured framework for the robot's interactions with its environment and other agents. For instance, in scenarios where robots are deployed in manufacturing facilities, rule-based navigation can be tailored to ensure safe and efficient movement, adhering to spatial boundaries and predefined routes  \citep{kabtoul_towards_2020}. While rule-based methods offer a high degree of predictability and safety, they may lack adaptability in dynamic and unstructured environments, where situations can change rapidly \citep{teja_s_hateb-2_2020, galvan_robot_2019}.

On the other hand, learning methods in collaborative robot navigation rely on machine learning algorithms to enable robots to acquire knowledge and adapt their navigation strategies over time  \citep{gil_social_2021, toghi_altruistic_2021}. These approaches leverage data-driven techniques to make informed decisions based on the robot's interactions with its surroundings and other agents \citep{garrell_aerial_2017, garrell_teaching_2019}. For example, reinforcement learning algorithms can enable robots to learn optimal paths and navigation behaviors through trial and error, while deep learning models can be used to recognize and respond to various environmental cues \citep{yen2004reinforcement}. Learning methods excel in scenarios where the environment is constantly changing, and adaptability is essential. However, they may require significant amounts of training data and computational resources to develop robust navigation capabilities, which can be a challenge in some applications \citep{toghi_altruistic_2021, kabtoul_proactive_2022, kabtoul_proactivesmooth_2022, toghi_social_2022}. The choice between rule-based and learning methods often depends on the specific requirements of the collaborative robot navigation task and the level of adaptability and autonomy desired.

\textbf{Assistive socially aware navigation} is an important domain in the field of robotics that seeks to create robotic systems that can help people with navigation tasks without requiring either explicit or active cooperation \citep{rios-martinez_intention_2012, morales_social_2017, shin_optimization-based_2020, yao_autonomous_2019, repiso_peoples_2019, repiso_adaptive_2019}. This discipline has been through a huge transition in recent years, mostly due to rapid advancements in artificial intelligence and machine learning. These developments in technology have made it possible to create complex algorithms  with context-aware and sophisticated assistance while navigating \citep{garrell_aerial_2017, vasquez_human_2013, triebel_spencer_2016, yao_monocular_2017, shin_optimization-based_2020}.

One significant avenue of development in assistive socially aware navigation is the integration of deep learning-based techniques \citep{garrell_teaching_2019}. These techniques have demonstrated great potential, especially in urban environments where robots can produce natural language instructions specific to the individual's position and destination \citep{chen2017deep, ye2020context}. Robots can interpret user intent, examine the environment, and communicate instructions in a manner that is human-friendly by employing deep learning. This method has the potential to completely transform urban navigation by increasing its efficiency and accessibility, particularly in complex and crowded urban settings \citep{zhu2021deep, liang2018cirl}.

Reinforcement learning-based methods are another important area in this field, as they allow to factor in variables that are frequently difficult to handcraft. 
These methods allow robots to observe humans and modify their behavior and actions accordingly \citep{hua2021learning}. They have also shown to be quite beneficial for assisting with transportation-related activities \citep{toghi_social_2022}. These methods allow robots to observe humans and modify their behavior and actions accordingly. The robot can adapt and adjust its actions to the dynamic and frequently unexpected traits of real-world surroundings, such as small-scale public spaces thanks to reinforcement learning. This flexibility guarantees that the robot will be able to efficiently address the numerous and changing demands of those who need help in a navigation task \citep{li2019deep}.

In summary, machine learning and artificial intelligence are enabling assistive socially aware navigation, which is rapidly developing intelligent, adaptive, and supportive robotic systems that can help people navigate challenging and constantly changing environments. These developments have the potential to significantly improve people's mobility and quality of life in a variety of settings.

\textbf{Independent} socially aware navigation is an important area where robots autonomously move without direct human interaction \citep{peddi_data-driven_2020, narayanan_transient-goal_2018, park_hi_2016}. This paradigm is especially applicable to situations where robots are working autonomously and need to navigate through crowded environments \citep{vasquez_inverse_2014, chen_relational_2020}. The use of these techniques is necessary when navigating through crowded areas in order to guarantee safe and efficient motion.

Crowd navigation employs a diverse set of strategies. For instance, \cite{narayanan_formalizing_2018} and \cite{brito_where_2021} predict sub-goals to move towards the goal while \citep{nishimura_l2b_2020} learns how to efficiently move through the crowd without freezing or timing out. A similar approach focused on avoiding robot freezing among human groups was proposed by \cite{sathyamoorthy_frozone_2020}. \cite{dugas_ian_2020} use different communication modalities to assist in passing through dense spaces. A number of advanced obstacle avoidance policies are also proposed to pass through crowds like \citep{chen_socially_2017, chen_crowd-robot_2019,  bera_emotionally_2019, cunningham_mpdm_2019, qiu_learning_2022, gonon_robots_2022, kastner_enhancing_2022, wang_feedback_2022}. 

Furthermore, robots moving in warehouses \citep{fernandez_carmona_making_2019, kenk_human-aware_2019, ga_learning_2020} and approaching people for interaction \citep{truong_approach_2018, ramirez_robots_2016} could be included into both independent and assistive socially aware navigation category.

Despite significant progress in independent socially aware navigation, several challenges remain. For instance, the development of algorithms that can handle a wide range of social environments and cultural contexts, the integration of multiple modalities for perception and sensing, and the improvement of safety and privacy in socially aware navigation \citep{qiu_learning_2022,salvini_safety_2022,narayanan_formalizing_2018,shrestha_using_2015,bera_emotionally_2019}.

\subsection{Negotiation}

Negotiation, in the context of robotics, and particularly in the domain of robot navigation, refers to the dynamic interaction and communication between robots and other entities, including humans and other robots, in order to accomplish efficient and successful movements. It involves a process where robots actively exchange information to coordinate their actions and resolve potential conflicts. Thus, negotiation strategies, in robot navigation, encompass a wide spectrum, from simple actions like requesting passage through a congested area, to more complex decision-making processes that balance different objectives,  eventually helping in the beneficial interaction of humans and robots in shared areas.

Therefore, this section explores research that includes explicit negotiation, which comprises elements such as requesting permission to move forward or clearly expressing intentions, and implicit negotiations like dynamic behavior adaptation by detecting intentions. Agents (human or robotic) can come to agreements or solve issues through the process of negotiation.

Negotiation in robot navigation requires the robot to be able to interpret and respond to the needs and motivations of pedestrians with whom it is negotiating, as well as to effectively communicate its own goals and constraints. 
In \citep{dalmasso_human-robot_2021}, the robot computes a multi-agent plan for both itself and the human which is then communicated to the human for review; this planner is based on a decentralized variant of Monte Carlo Tree Search (MCTS) with one robot and one human as agents. In \citep{hsu_pomdp_2020} researchers contribute with a minimal model to manage ambiguity and produce actions that are expressive and encode aspects of humans' intents. Furthermore, \cite{lobato_human-robot_2019} present a socially aware navigation system that allows to establish a negotiation framework to improve the socially aware navigation system. 

The ability to effectively negotiate in robot navigation can be a key factor in enabling robots to interact and cooperate with humans and their environment in a natural and intuitive manner \citep{jensen_knowing_2018}. \cite{vega_planning_2019} focus on planning algorithms that facilitate negotiation between robots and humans in dynamic environments. \cite{kabtoul_proactive_2020} propose a proactive negotiation approach to enhance human-robot collaboration. \cite{dondrup_qualitative_2016} explore qualitative spatial reasoning techniques for negotiating spatial relations in human-robot interaction scenarios. Furthermore, \citep{chen_relational_2020} investigate graph strategies that enable robots to establish relations among agents and maintain advanced predictions of humans to negotiate their plan better during navigation tasks. \cite{nishimura_l2b_2020} explore the robot beeping mechanism to negotiate with the humans to clear the way. These works contribute valuable insights into the field of negotiation in socially aware robot navigation, paving the way for the development of more efficient and interactive robotic systems.

Another important aspect of negotiation is the ability to adapt to changing circumstances. The negotiation process may involve a number of back-and-forth exchanges as the agents work to reach an agreement, and the robot must be able to adjust its negotiation strategy as needed to reach a mutually acceptable solution. \cite{trautman_robot_2015} explore the use of adaptive negotiation strategies in the context of human-robot collaboration, emphasizing the importance of dynamically adjusting negotiation behaviors based on situational cues. 
\cite{bera_socially_2018} propose a socially adaptive negotiation framework that enables robots to learn and modify their negotiation strategies based on user preferences and interaction history. \cite{shrestha_using_2015} investigate the use of contact-based inducement to negotiate in a congested scenario. Finally, \cite{ratsamee_social_2013} focus on the role of adaptability in negotiation, demonstrating the need for robots to continuously learn and adapt their negotiation behaviors to foster successful human-robot interactions. All these works highlight the significance of adaptive negotiation strategies in enabling robots to effectively navigate and interact with humans in dynamic environments.

In addition to these aspects, negotiation in socially aware robot navigation may also involve resolving conflicts or obstacles that arise during task execution \citep{toghi_altruistic_2021}, as well as coordinating actions and sharing resources with other robots or autonomous systems. By effectively negotiating with these parties, the robot can facilitate cooperation and coordination, enabling it to achieve its goals and to complete tasks more effectively \citep{evens_learning_2021}.

\subsection{Local Motion Generation}
Local motion generation, which often involves local sensing and perception, is the creation of a trajectory or velocity commands for guiding the robot's motion at a lower level \citep{boldrer_multi-agent_2022}.  Local motion generation can be divided into several categories, including \textit{planning-based approaches} \citep{unhelkar_human-robot_2015, bera_sociosense_2017, singamaneni_human-aware_2021, banisetty_deep_2021, chen_unified_2021, gonon_robots_2022}, \textit{force-based approaches} (like potential fields and social forces) \citep{patompak_mobile_2016,cunningham_mpdm_2019, repiso_adaptive_2019, kose_socially_2018, jiang_interactive_2016}, \textit{learning-based approaches} \citep{liu_robot_2020,chen_socially_2017, sathyamoorthy_frozone_2020}, and \textit{others}, which are not included in the previous categories \citep{paez-granados_unfreezing_2022}.

\par
\textbf{Planning-based approaches} involve generating a trajectory for a robot to follow and, then, converting it to velocity commands. These approaches are often used in complex environments where obstacles and other dynamic elements need to be taken into account. Different planning methodologies can be used to generate trajectories, such as Model Predictive Control (MPC), dynamic windows \citep{truong_approach_2018, kabtoul_proactivesmooth_2022}, elastic bands \citep{rosmann_online_2017, vega_planning_2019, singamaneni_human-aware_2021, khambhaita_viewing_2017, singamaneni_watch_2022}, and obstacle avoidance techniques \citep{jiang_design_2022, gonon_robots_2022, bera_sociosense_2017}.

\par
MPC-based approaches use a predictive model of the robot's motion to generate an optimal trajectory over a finite time horizon. The trajectory is generated by solving an optimization problem that takes into account the robot's kinematics, dynamics, and environmental constraints. For instance, \cite{che_efficient_2020} propose a planning framework, based on MPC, that generates explicit communication (finite number of discrete signals) and robot motions. In \citep{brito_where_2021}, the robot combines a sub-goal prediction mechanism with an MPC controller to navigate the environment efficiently. \cite{evens_learning_2021} propose an MPC-based scheme to handle general traffic situations for PMVs.

\par
Dynamic window approaches (DWA) involve generating a set of reachable velocities based on the robot's kinematics and dynamics \citep{truong_approach_2018}. The set of reachable velocities is used to select the best velocity command that will take the robot closer to the goal while avoiding obstacles. A similar approach presented in \citep{kabtoul_proactivesmooth_2022} uses a dynamic channel to maneuver around pedestrians while anticipating their cooperation. \cite{dondrup_qualitative_2016} propose some modifications to DWA to generate safe and efficient trajectories to reach the goal while trying to ensure human acceptance. Due to its conceptual simplicity, DWA is one of the widely used approaches and can be easily employed to test new ideas \citep{truong_approach_2018, dugas_ian_2020}. 

\par
Elastic band approaches, on the other hand, require a slightly more complex implementation. They involve generating a path for the robot to follow and simulating an elastic band to smooth the path while stretching (or compressing) it around obstacles and generate velocity commands \citep{khambhaita_viewing_2017, vega_socially_2019, pimentel_evaluation_2021}.
Traditional elastic bands could include only kinematics constraints \citep{vega-magro_flexible_2018, vega_socially_2019} whereas the timed elastic bands proposed by \cite{rosmann_online_2017} can handle kinodynamic constraints.
\cite{khambhaita_viewing_2017} and \cite{singamaneni_human-aware_2021} use these timed elastic bands for proactive planning to solve complex human-robot navigation settings.

\par
Finally, the dynamic obstacle avoidance techniques that are prevalent in the motion planning community have also been modified to accommodate humans and navigate safely in social environments.
The works of \cite{bera_classifying_2018, bera_emotionally_2019} rely on the generalized velocity obstacles approach. Very recent work by \cite{gonon_robots_2022} proposes acceleration obstacles to handle the case of crowd robot navigation specifically. Lastly, \cite{sathyamoorthy_frozone_2020} proposes a hybrid approach to avoid freezing by switching between planned and learned controls.

\par
{\bf Force-based approaches} like potential fields and social forces, involve generating velocity commands based on the potential or force fields in the robot's environment and the interaction forces generated by the way people move. They are widely used in robot navigation to generate velocity commands based on attractive and repulsive forces. The first group uses virtual potential fields to generate attractive forces towards the goal and repulsive forces away from obstacles~\citep{araujo_architecture_2015}.  The robot's motion is then controlled based on the gradient of the potential field. On the other hand, the social force model (SFM) uses the concept of social forces, where the forces result from interactions between individuals or groups and the robot. The forces to avoid collisions with humans are generated using the relative velocities between the robot and the pedestrians, and they are combined with the forces from the other obstacles \citep{ repiso_peoples_2020,alahi_learning_2017}.
Finally, these combined forces are used to generate velocity commands for the robot.

One popular potential field approach is the Artificial Potential Field (APF) method \citep{jiang_interactive_2016}.
The APF method has been widely used in robot navigation and extended to dynamic environments by incorporating time-varying potentials and obstacle-avoidance strategies \citep{ferrara2007sliding}.  
The simplicity of implementation and computational efficiency of APF makes it ideal for real-time applications. Considering the adaptive nature of potential fields, some researchers used it to address both simple \citep{wang_adaptive_2016} and complex socially aware navigation tasks in dynamic environments like offices \citep{araujo_architecture_2015} or sparse crowds \citep{cunningham_mpdm_2019}.

\par
Social force approaches were first proposed by \cite{helbing1995social} to model pedestrian behavior in crowds. This idea was later adapted to control the robot's motion based on the sum of attractive and repulsive forces generated by the social forces from various kinds of interactions \citep{ferrer_robot_2013,ferrer_social-aware_2013,patompak_mobile_2016, repiso_-line_2017}. Since its inception, social force-based robot control has been modified and extended to address different kinds of problems in social robot navigation. \cite{ferrer_robot_2013} used it to navigate the robot in crowded environments and later proposed an extension for proactive kinodynamic planning \citep{ferrer2014proactive}. Building on this extended SFM, a set of works \citep{repiso_robot_2018, repiso_peoples_2019, repiso_peoples_2020, repiso_adaptive_2022} were proposed to approach and accompany individual as well as a group of people. This online adaptive planning in dynamic environments is achieved by incorporating time-varying social forces \citep{repiso_adaptive_2019,repiso_peoples_2020}. In the works presented in \citep{truong_socially_2017} and \citep{truong_toward_2017}, the SFM is extended to include human-object interactions and group interactions to address dynamic crowds. SFM has been adapted to aerial robots as well and the works by \cite{garrell_aerial_2017, garrell_teaching_2019} show these extensions called aerial social force models designed to accompany humans. Some recent approaches combine policy learning \citep{cunningham_mpdm_2019, gil_effects_2019} and machine learning \citep{gil_social_2021} with SFM to achieve better socially aware robot navigation policies.

\par
In \textbf{learning-based approaches} velocity commands are frequently generated directly, without the need to construct a precise trajectory, through the application of machine learning algorithms. These approaches typically involve either reinforcement learning \citep{chen_socially_2017, ga_learning_2020, gil_robot_2022}, deep learning \citep{gil_social_2021, xie_towards_2021}, imitation learning \citep{garrell_teaching_2019, fahad_learning_2020, liu_robot_2020} or inverse reinforcement learning \citep{vasquez_inverse_2014, ramirez_robots_2016}. 

Reinforcement learning (RL) or deep reinforcement learning (DRL) in general is used by several articles in this survey to teach socially aware navigation to a robot. For instance, the works by \cite{chen_socially_2017, chen_crowd-robot_2019, chen_relational_2020} use different kinds of networks architectures to capture the relations and interactions in the crowd and teach a robot to move socially. They use the deep V-learning where the neural network is initialized with regression and then trained using reinforcement learning. In \citep{qiu_learning_2022} researchers also propose a hybrid learning approach combining supervised learning with DRL. The authors learn the interactions among pedestrians using supervised learning, and this interaction policy is used within DRL navigation policy training to learn when to alarm the surrounding pedestrians to clear the path.
This alerting mechanism for path clearing was inspired by \citep{nishimura_l2b_2020}, where the authors learn the balance between human safety and navigation efficiency in a similar manner. 

As mentioned previously, the works by \cite{gil_effects_2019, gil_robot_2022} combine DRL with SFM, in order to study the effects of SFM rewards and human motion prediction strategies on the navigation policy. The work in \citep{ga_learning_2020} presents a DRL-based ROS local planner that is trained to avoid humans in warehouses using 2D LiDAR data as input. DRL is also explored for learning the navigation policies for autonomous vehicles (AVs). The work by \cite{deshpande_behavioral_2020} uses deep recurrent Q-network to handle high-level behavioral decision-making while the AV is navigating among pedestrians. In other works \citep{toghi_altruistic_2021, toghi_social_2022}, different policy learning mechanisms like A2C and multi-agent RL are used to learn social behaviors in traffic with emphasis on coordination and altruism. 

\par
Among the other kinds of learning approaches, deep learning is generally employed for robot perception. However, there are works like \citep{xie_towards_2021} that use deep learning to navigate through crowded environments. Imitation learning is employed to clone the behavior of humans in \citep{fahad_learning_2020} while it is used to assist robot navigation policy learning in \citep{liu_robot_2020}. \cite{garrell_teaching_2019} use imitation learning to make a neural network learn to mimic the expert flying a drone. The work in \citep{ramirez_robots_2016} uses inverse reinforcement learning to train a robot to approach humans appropriately while \cite{perez-higueras_robot_2014} use it to navigate robot in public spaces.

\par
One key advantage of learning-based approaches is that they can capture the nuances of social interaction, such as motion dynamics \citep{chen_relational_2020}, social norms, and respond appropriately to human feedback \citep{ kollmitz_learning_2020}. Hence, learning-based approaches have the potential to significantly improve the field of socially aware robot navigation, enabling robots to navigate and interact with humans in a more natural and intuitive way \citep{toghi_altruistic_2021}. These approaches typically require a model trained on a dataset of sensor inputs and corresponding command velocities, using a suitable loss function and optimization algorithm \citep{triebel_spencer_2016,liu_robot_2020,toghi_social_2022}. Researchers are also focusing on more innovative approaches and applications of socially aware robot navigation for more complex social environments \citep{park_hi_2016,ciou_composite_2018,evens_learning_2021,qiu_learning_2022}.

Last but not least, there exist papers that do not fall under any of the above groups but deal with low-level motion generation. They have been classified as \textbf{others} in our taxonomy. In \citep{yoon_socially_2019} authors introduce a novel framework for path planning that considers the safety perception of humans when a flying robot is present.  
With this, they aim to ensure safe and socially acceptable interactions between humans and flying robots. Researchers in \citep{bera_sociosense_2017} published an approach  to mathematically model social cues in order to predict both human trajectories and personal/social distances, which are key components of socially aware navigation planning. To accomplish this, a Bayesian-based model of personality traits is used, with video data serving as the source for observing and quantifying these traits. By leveraging these models of human behavior, the aim is to enhance the ability of robots to interact with humans in a safe and socially acceptable manner. A novel methodology to unfreeze the robot from unintended collisions with pedestrians is proposed in \citep{paez-granados_unfreezing_2022}. They design a special controller that modulates the velocity upon detection of contact to mitigate the risks. In the work presented by \cite{jiang_design_2022}, a pedestrian-aware controller for an autonomous car was proposed that modulates the speed depending on the estimated pedestrian density.

\subsection{Global Motion Decision-Making}

Global motion decision-making, in the context of socially aware robot navigation is the process of computing a valid robot trajectory at a coarse level, taking into account the requirements of socially aware navigation. It often utilizes a representation of the environment to guide the process, and considers aspects like collision avoidance and the needs of bystanders.
This differs from local decision-making, which relies on sensors and the immediate surroundings to guide motion. The approaches for global motion decision-making mainly consist of \textit{planning-based} approaches (search and sampling) \citep{korkmaz_human-aware_2021, forer_socially-aware_2018, singamaneni_human-aware_2021, kollmitz_time_2015, talebpour_incorporating_2016, vega-magro_socially_2017, chen_socially_2017} and \textit{learning-based} approaches \citep{luber_socially-aware_2012, perez-higueras_learning_2018, karnan_socially_2022, brito_where_2021}.

\textbf{Planning-based} approaches are very frequently used to make global-level decisions and plan the initial path for the robot to follow.  The articles in this survey include \textit{search-based approaches}, like A* methods \citep{luo_porca_2018, banisetty_towards_2018}, D* methods \citep{charalampous_robot_2016}, diffusion maps \citep{chen_socially_2017}, Dijkstra \citep{perez-higueras_robot_2014, truong_approach_2018}, etc.; and \textbf{sampling approaches} like PRM \citep{korkmaz_human-aware_2021}, RRT \citep{becerra_human_2020, shrestha_using_2015}, Risk-RRT \citep{narayanan_formalizing_2018}, PRM-RRT \citep{vega-magro_socially_2017}, Fast Marching methods \citep{talebpour_incorporating_2016}, etc. 

One of the main challenges in this area has been the ability of the robot to adapt to changing environments or unexpected obstacles \citep{repiso_adaptive_2019}. To address this challenge, some researchers have developed methods that incorporate real-time feedback or sensory data into the global planning process \citep{peddi_data-driven_2020,vega-magro_flexible_2018,randhavane_pedestrian_2019}. Other approaches use a combination of global and local maps to generate a decision or plan a better path \citep{dondrup_qualitative_2016, fernandez_carmona_making_2019, singamaneni_watch_2022}. The global map provides a high-level view of the environment, while the local map represents the robot's immediate surroundings in more detail. By combining these two types of maps, the robot can generate a path that is both efficient and able to adapt to local environmental changes \citep{teja_s_hateb-2_2020, singamaneni_watch_2022, kollmitz_time_2015}. Continuous re-planning is also used in some cases \citep{korkmaz_human-aware_2021}.

\par
The papers that use \textbf{learning-based} methods involve training a model on data to make predictions about future states or decisions. These methods can leverage large amounts of data to learn patterns and adapt to new situations but may require significant training time and may not generalize well to novel situations. These methods make use of deep reinforcement learning \citep{brito_where_2021, valiente_robustness_2022}, deep learning architectures like CNNs \citep{perez-higueras_learning_2018}, and inverse reinforcement learning \citep{vasquez_inverse_2014}. Even with the limitations, these approaches can help provide a good initial estimate that assists in better planning \citep{perez-higueras_learning_2018}. Further, they can be used to select sub-goals for guiding a local planner \citep{brito_where_2021}. In PMVs as well, they can be used to make high-level decisions \citep{valiente_robustness_2022}.

\par
There are also papers that do not explicitly focus on either of these approaches, but instead consider additional aspects of global motion decision-making, such as the representation of the environment \citep{arndt_safe_2015}, the incorporation of social cues, the integration of multiple modalities of sensing and communication \citep{che_efficient_2020}, or Wizard-of-Oz studies \citep{lichtenthaler_social_2013}. In general, the choice of a global motion decision-making strategy depends on specific demands and features that characterize the concerned task.


\section{Situation Awareness and Assessment}
\label{sec:situation}

\begin{figure*}[!hb]
    \subfloat{
      \includegraphics[width=0.9\columnwidth]{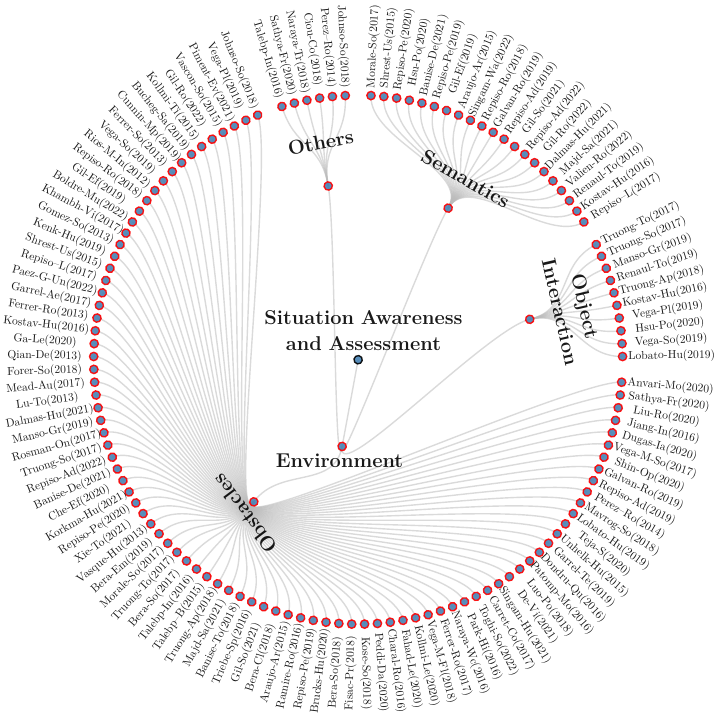}
    }
    ~
    \subfloat{
      \includegraphics[width=1.1\columnwidth]{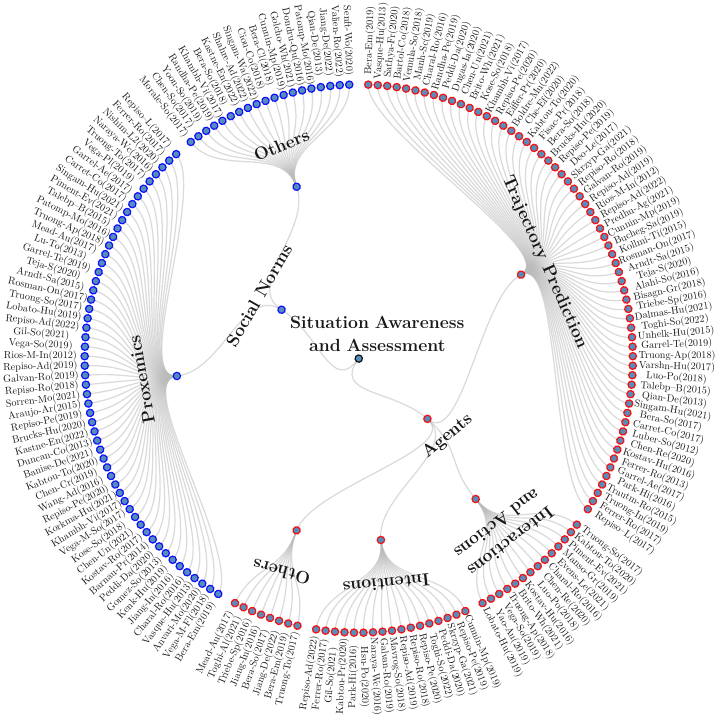}
    }
    \caption{Distribution of papers by situation awareness and assessment. The figures are best viewed zoomed in using a digital version.}
    \label{fig:SA_distribution}
\end{figure*}

Fig.~\ref{fig:SA_distribution} shows the distribution of the papers according to the taxonomic aspects of situation awareness and assessment. 
Most works exploit elements of the three main branches, although none of the reviewed proposals considers all of them.
By a large margin, the most frequent aspects considered in the literature for situation awareness are \textit{obstacles} in the \textit{environment}, \textit{trajectory prediction} of the \textit{agents}, and \textit{proxemics} constraints as the main \textit{social norm}.
The remainder of the section presents a detailed analysis of the different taxa.

\subsection{Environment}
This taxon considers papers representing aspects related to the physical space in which the robot navigates other than the agents in the environment, which are considered in the next section. Collective issues such as the density of humans in the area are also considered.
\par
The \textbf{semantics of the environment} is a topic that has received limited attention in the literature. 
Some proposals assume a specific type of space for navigation, such as the office-like environment in~\citep{araujo_architecture_2015} or the wheelchair navigation system in~\citep{morales_social_2017} that estimates corridor width. 
In~\citep{banisetty_deep_2021}, the navigation system includes a context classification module that distinguishes between four contexts and is used to guide the robot in selecting social objectives.
Other approaches, such as the socially aware variant of a NAMO algorithm in~\citep{renault_towards_2019}, use a semantic map with taboo zones for movable obstacle placement, while~\cite{kostavelis_human_2016} combine a metric map with a structured map containing relevant objects and standing positions for humans that are used to improve future predictions of the occupancy of different areas. 
In a similar way, in~\citep{kostavelis_robots_2017} predefined locations of frequently visited areas of the environment are used for human presence anticipation.
The work presented by \cite{singamaneni_watch_2022} also proposes a geometric approach to anticipate the emergence of humans from occluded locations.
\par
An alternative application of the semantics of the environment can be found in~\citep{hsu_pomdp_2020}. Their proposed system provides estimates for the intentions of pedestrians and nearby pedestrian crossings using semantics information as input.
Other proposals consider environment information, although not of semantic type (e.g. size, structure), for navigation~\citep{vega-magro_flexible_2018, manso_graph_2019}.
\par
In relation to \textbf{object interaction}, most of the proposals considering this element, model the interaction area using predefined functions to prevent the robot from traversing those zones. 
For instance, \cite{lobato_human-robot_2019} and \cite{vega_socially_2019} model the interaction zone as a symmetric trapezoidal area, while others \citep{truong_approach_2018} use Gaussian functions.
Likewise, \cite{truong_toward_2017} and \cite{truong_socially_2017} consider detected object interaction, creating a circular object interaction space to avoid.
Differently, in~\cite{manso_graph_2019} object interactions are included in the representation of a scene, but the interaction areas are not explicitly modeled. 
\par
A different perspective on applying human-object interactions for socially aware navigation can be found in \cite{bruckschen_human-aware_2020} and \cite{vega_planning_2019}. In particular, \cite{bruckschen_human-aware_2020} use observed human-object interactions along with prior knowledge about typical human transitions to predict the most likely navigation goal of the human. 
On the other hand, \cite{vega_planning_2019}  consider only one type of interaction with one type of object, namely doors, and focuses on the relationship where one or more humans are blocking the door, which is used in the proposal to ask for permission to pass.
\par
Other than agents, \textbf{obstacles} constitute the most important type of physical elements of the environment in the vast majority of socially aware navigation works.
Navigation algorithms not considering obstacles work on simple scenarios where humans (or agents in general) are the only entity robots may collide with~\citep{nishimura_l2b_2020}. 
This is particularly common in simulated environments.
Some of the obstacle-aware approaches to socially aware navigation do not use a representation that integrates information over time. 
Instead, they use the instantaneous information perceived through the robot's sensors \citep{de_vicente_deepsocnav_2021,guzzi_human-friendly_2013,sathyamoorthy_frozone_2020,paez-granados_unfreezing_2022}.
\par
\begin{figure}[th]
    \begin{center}
    \vspace{-1cm}
    \hspace{-1cm}
        \resizebox{1.6\columnwidth}{!}{%
        \hypersetup{citecolor=black}
\usetikzlibrary{patterns}
\begin{tikzpicture}[font=\small]
    \begin{axis}[
        /pgf/number format/1000 sep={},
        width=3.8in,
        axis y line*=left,
        axis x line*=bottom,
        scale only axis,
        clip=false,
        y axis line style={draw opacity=0},
        x axis line style={draw opacity=0},
        separate axis lines,
        axis on top,
        xmin=0,
        xmax=5,
        xtick={0.55,1.65,2.75},
        x tick style={draw=none},
        xticklabels={Dense, Sparse, Hybrid},
        ytick=\empty,
        ymin=0,
        ymax=110,
        every axis plot/.append style={
          ybar,
          bar width=0.65,
          bar shift=0pt,
          fill,
        },
        font=\footnotesize
      ]

\addplot[black, draw opacity = 1.0, fill= gray, fill opacity = 0.1, y filter/.code={\pgfmathparse{#1*2.1}\pgfmathresult}]coordinates {(0.55,45)};
\node[overlay, font=\tiny] at (0.55,1.00){\mycite{bruckschen_human-aware_2020}};
\node[overlay, font=\tiny] at (0.55,3.10){\mycite{lu_towards_2013}};
\node[overlay, font=\tiny] at (0.55,5.20){\mycite{unhelkar_human-robot_2015}};
\node[overlay, font=\tiny] at (0.55,7.30){\mycite{bera_classifying_2018}};
\node[overlay, font=\tiny] at (0.55,9.40){\mycite{perez-higueras_robot_2014}};
\node[overlay, font=\tiny] at (0.55,11.50){\mycite{cunningham_mpdm_2019}};
\node[overlay, font=\tiny] at (0.55,13.60){\mycite{kose_socially_2018}};
\node[overlay, font=\tiny] at (0.55,15.70){\mycite{mavrogiannis_social_2018}};
\node[overlay, font=\tiny] at (0.55,17.80){\mycite{teja_s_hateb-2_2020}};
\node[overlay, font=\tiny] at (0.55,19.90){\mycite{singamaneni_human-aware_2021}};
\node[overlay, font=\tiny] at (0.55,22.00){\mycite{khambhaita_viewing_2017}};
\node[overlay, font=\tiny] at (0.55,24.10){\mycite{vega-magro_socially_2017}};
\node[overlay, font=\tiny] at (0.55,26.20){\mycite{dondrup_qualitative_2016}};
\node[overlay, font=\tiny] at (0.55,28.30){\mycite{mead_autonomous_2017}};
\node[overlay, font=\tiny] at (0.55,30.40){\mycite{truong_approach_2018}};
\node[overlay, font=\tiny] at (0.55,32.50){\mycite{fahad_learning_2020}};
\node[overlay, font=\tiny] at (0.55,34.60){\mycite{morales_social_2017}};
\node[overlay, font=\tiny] at (0.55,36.70){\mycite{korkmaz_human-aware_2021}};
\node[overlay, font=\tiny] at (0.55,38.80){\mycite{talebpour_-board_2015}};
\node[overlay, font=\tiny] at (0.55,40.90){\mycite{patompak_mobile_2016}};
\node[overlay, font=\tiny] at (0.55,43.00){\mycite{kenk_human-aware_2019}};
\node[overlay, font=\tiny] at (0.55,45.10){\mycite{rios-martinez_intention_2012}};
\node[overlay, font=\tiny] at (0.55,47.20){\mycite{kostavelis_robots_2017}};
\node[overlay, font=\tiny] at (0.55,49.30){\mycite{sorrentino_modeling_2021}};
\node[overlay, font=\tiny] at (0.55,51.40){\mycite{dugas_ian_2020}};
\node[overlay, font=\tiny] at (0.55,53.50){\mycite{talebpour_incorporating_2016}};
\node[overlay, font=\tiny] at (0.55,55.60){\mycite{gomez_social_2013}};
\node[overlay, font=\tiny] at (0.55,57.70){\mycite{vega_planning_2019}};
\node[overlay, font=\tiny] at (0.55,59.80){\mycite{buchegger_safe_2019}};
\node[overlay, font=\tiny] at (0.55,61.90){\mycite{kollmitz_learning_2020}};
\node[overlay, font=\tiny] at (0.55,64.00){\mycite{kollmitz_time_2015}};
\node[overlay, font=\tiny] at (0.55,66.10){\mycite{vasquez_human_2013}};
\node[overlay, font=\tiny] at (0.55,68.20){\mycite{triebel_spencer_2016}};
\node[overlay, font=\tiny] at (0.55,70.30){\mycite{charalampous_robot_2016}};
\node[overlay, font=\tiny] at (0.55,72.40){\mycite{forer_socially-aware_2018}};
\node[overlay, font=\tiny] at (0.55,74.50){\mycite{banisetty_towards_2018}};
\node[overlay, font=\tiny] at (0.55,76.60){\mycite{mateus_efficient_2016}};
\node[overlay, font=\tiny] at (0.55,78.70){\mycite{che_efficient_2020}};
\node[overlay, font=\tiny] at (0.55,80.80){\mycite{park_hi_2016}};
\node[overlay, font=\tiny] at (0.55,82.90){\mycite{ga_learning_2020}};
\node[overlay, font=\tiny] at (0.55,85.00){\mycite{xie_towards_2021}};
\node[overlay, font=\tiny] at (0.55,87.10){\mycite{qiu_learning_2022}};
\node[overlay, font=\tiny] at (0.55,89.20){\mycite{truc_khaos_2022}};
\node[overlay, font=\tiny] at (0.55,91.30){\mycite{majd_safe_2021}};
\node[overlay, font=\tiny] at (0.55,93.40){\mycite{boldrer_multi-agent_2022}};

\addplot[black, draw opacity = 1.0, fill= gray, fill opacity = 0.1, y filter/.code={\pgfmathparse{#1*2.1}\pgfmathresult}]coordinates {(1.65,16)};
\node[overlay, font=\tiny] at (1.65,1.00){\mycite{manso_graph_2019}};
\node[overlay, font=\tiny] at (1.65,3.10){\mycite{garrell_aerial_2017}};
\node[overlay, font=\tiny] at (1.65,5.20){\mycite{garrell_teaching_2019}};
\node[overlay, font=\tiny] at (1.65,7.30){\mycite{carretero_comfort-oriented_2017}};
\node[overlay, font=\tiny] at (1.65,9.40){\mycite{ferrer_robot_2017}};
\node[overlay, font=\tiny] at (1.65,11.50){\mycite{anvari_modelling_2020}};
\node[overlay, font=\tiny] at (1.65,13.60){\mycite{jiang_interactive_2016}};
\node[overlay, font=\tiny] at (1.65,15.70){\mycite{araujo_architecture_2015}};
\node[overlay, font=\tiny] at (1.65,17.80){\mycite{truong_socially_2017}};
\node[overlay, font=\tiny] at (1.65,19.90){\mycite{fisac_probabilistically_2018}};
\node[overlay, font=\tiny] at (1.65,22.00){\mycite{bera_sociosense_2017}};
\node[overlay, font=\tiny] at (1.65,24.10){\mycite{bera_socially_2018}};
\node[overlay, font=\tiny] at (1.65,26.20){\mycite{randhavane_pedestrian_2019}};
\node[overlay, font=\tiny] at (1.65,28.30){\mycite{ferrer_social-aware_2013}};
\node[overlay, font=\tiny] at (1.65,30.40){\mycite{ferrer_robot_2013}};
\node[overlay, font=\tiny] at (1.65,32.50){\mycite{truong_toward_2017}};

\addplot[black, draw opacity = 1.0, fill= gray, fill opacity = 0.1, y filter/.code={\pgfmathparse{#1*2.1}\pgfmathresult}]coordinates {(2.75,6)};
\node[overlay, font=\tiny] at (2.75,1.00){\mycite{lobato_human-robot_2019}};
\node[overlay, font=\tiny] at (2.75,3.10){\mycite{renault_towards_2019}};
\node[overlay, font=\tiny] at (2.75,5.20){\mycite{kostavelis_human_2016}};
\node[overlay, font=\tiny] at (2.75,7.30){\mycite{johnson_socially-aware_2018}};
\node[overlay, font=\tiny] at (2.75,9.40){\mycite{vega-magro_flexible_2018}};
\node[overlay, font=\tiny] at (2.75,11.50){\mycite{vega_socially_2019}};

    \end{axis}
  \end{tikzpicture}}
        \caption{Classification of papers according to how obstacles are represented.}
        \label{fig:obstacle_representation} 
    \end{center}
    \vspace{-0.5cm}
\end{figure}
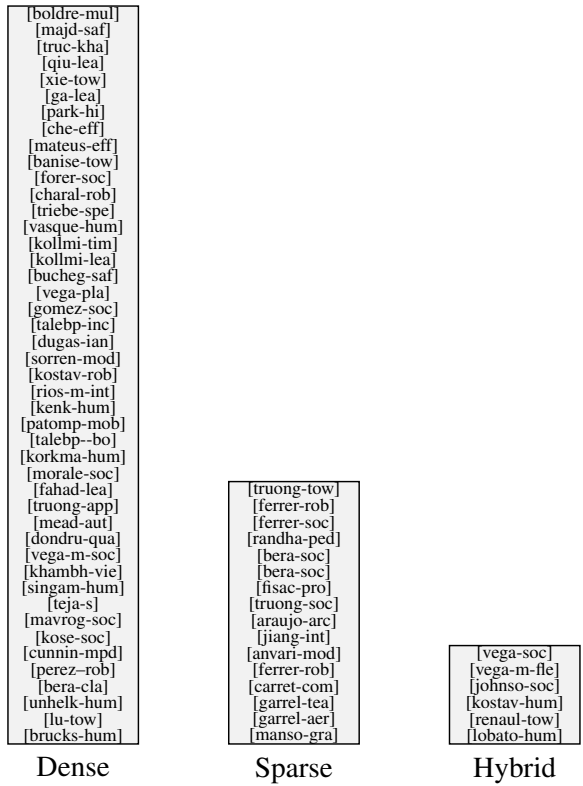

Papers representing obstacles have been classified according to three types of representation: dense, sparse, and hybrid. 
Fig. \ref{fig:obstacle_representation} shows the distribution of a representative subset of the reviewed papers into these three different types of representations.
\par
Dense representations consist of a metric map of the environment where the obstacles are located.
In this case, obstacles cannot be identified as individual entities, but the representation still allows disregarding areas of the environment that the robot cannot cross during navigation.
The commonly used dense representations are occupancy grids and cost maps.
This type of obstacle representation is the most widely used in the literature.
\par
Sparse representations, where each obstacle is an independent element with its own properties, are found on the other side of the spectrum.
Frequently, works using this kind of representation consider position and size as the only properties of the obstacles, assuming circular shapes for them \citep{ferrer_social-aware_2013, bera_socially_2018}.
\par
The third obstacle representation category corresponds to hybrid approaches, which combines dense and sparse models for different purposes.
Examples of hybrid representations can be found in \citep{lobato_human-robot_2019}, \citep{renault_towards_2019}, and \citep{vega-magro_flexible_2018}.
The proposal by \cite{lobato_human-robot_2019} uses a sparse obstacle representation, where detected objects are nodes of a symbolic graph of the environment and a dense one for modeling the occupied space.
In \citep{renault_towards_2019} a 2D metric map is built to compute a first plan.
Then, the plan is refined iterating over movable obstacles.
The work in \citep{vega-magro_flexible_2018} represents the obstacles of the environment by means of a cost map, but also incorporates wall descriptors that are later used by the control system.

Even though most of the environment-related elements considered for socially aware navigation can be included in the three already mentioned taxa (semantics of the environment, object interactions, and obstacles), some works take into account other aspects of the area where the navigation takes place. 
All these aspects are included in a fourth branch labeled as \textit{\textbf{others}}.
After our review, only four papers have been found that can be included in this branch.
Specifically, \cite{perez-higueras_learning_2018, ciou_composite_2018} and \cite{jiang_design_2022} consider people's density as an additional property of the environment.
Besides, \cite{johnson_socially-aware_2018} represent gateways along with other elements of the area where the robot navigates.

\subsection{Agents}
This taxon describes how the agents are represented.
We use the term \textit{agent} instead of \textit{human} to encompass all active entities involving humans that may be present in the environment (\textit{e.g.}, other vehicles in the case of autonomous driving).
\par
All works dealing with socially aware robot navigation consider a common set of attributes that provide, in the simplest case, a minimal representation through which an agent can be treated as a ``special obstacle''.
Strictly speaking, this common set consists of the 2D position of the agents, although the orientation of the agents is frequently considered as well.
Therefore, our taxonomy ignores these pose-related attributes in the proposed classification and focuses on other aspects of the agents for which a more diverse treatment can be found in the existing literature, starting with \textbf{trajectory prediction}.
\par
Predicting the trajectory of an agent involves estimating the agent's future positions based on its past positions and, potentially, information about the environment and other agents.
Pedestrian trajectory prediction is a research field in itself. 
Numerous works have been proposed on this topic, although they are not necessarily framed within a robot navigation proposal.
We will begin by focusing on trajectory prediction proposals that can be directly applied to robot navigation.
Next, we will analyze works that use trajectory prediction within the context of socially aware robot navigation.

\par
Given the recurrent nature of trajectory prediction and the surge in machine learning in recent years, recurrent neural networks (RNN) have a special role in recent trajectory prediction approaches.
One of these learning-based approaches is Social-LSTM \citep{alahi_social_2016}, which proposes an LSTM-based model that can jointly estimate the future trajectory of all the people in a scene, using one LSTM per individual and a pooling layer to share the information between them.
Based on this idea, some variants have been proposed.
For instance, \cite{bisagno_group_2018} first cluster people into coherent groups before using Social-LSTM to predict their trajectories.
\cite{varshneya_human_2017} propose a variant of Social-LSTM that considers several factors such as the dynamic of neighboring subjects and the spatial context in which the subject is.
In \citep{bartoli_context-aware_2018}, human-human and human-space interactions are incorporated into the Social-LSTM model.
Other proposals applying RNNs to trajectory prediction are the one by \cite{vemula_social_2018}, which uses RNNs to model the spatial and temporal dynamics of trajectories in human crowds, and the work by \cite{manh_scene-lstm_2019}, an LSTM-based approach that combines scene information into the human trajectory prediction.
Likewise, \cite{eiffert_probabilistic_2020} combine Recurrent Neural Networks with Mixture Density Networks for pedestrians’ trajectory prediction with the goal of enabling autonomous vehicles to navigate through crowds.
\par
In the context of autonomous driving, other approaches for trajectory prediction can be found.
\cite{kabtoul_towards_2020} present a model that estimates the pedestrian’s cooperation with the vehicle and uses this estimation to predict the trajectory of the pedestrian by a cooperation-based trajectory planning model.
Also, in \citep{predhumeau_agent-based_2021} a pedestrian trajectory prediction model is proposed for autonomous vehicles, combining SFM and a decision model for conflicting pedestrian-vehicle interactions.
Another example is the approach in \citep{deo_learning_2017}, which proposes an extension of the Variational Gaussian Mixture Model-based probabilistic trajectory prediction framework for on-road pedestrians.
The aforementioned proposals constitute a limited subset of the existing trajectory prediction approaches.
Many others can be found. 
For a more comprehensive overview of the topic, readers may refer to specific surveys \citep{ridel_literature_2018}.
\par

Despite advances and new techniques in trajectory prediction, many works in socially aware navigation propose their own approach to the problem. 
Some papers employ uncomplicated solutions to predict the trajectory of agents, using no other information than the last positions/velocities of the agent.
Thus, in the works by \cite{guzzi_human-friendly_2013} and \cite{chen_unified_2021}, all agents are assumed to keep their current heading and speed.
\cite{carretero_comfort-oriented_2017} estimates the new velocity of a human as the average of the last 10 velocities.
\cite{kose_socially_2018} do not predict trajectories as such, but estimate the time to collision according to the current positions and velocities of the humans.
\par
More elaborate solutions have also been proposed using only the past trajectory of the agents.
\cite{garrell_aerial_2017} use a prediction module based on online linear regression.
In \citep{garrell_teaching_2019} a neural network takes the last 10 known positions of a human to predict the new position one second into the future.
\cite{truong_approach_2018} and \cite{talebpour_-board_2015} apply Kalman filters for predicting the future state of pedestrians.
Probabilistic approaches~\citep{bera_sociosense_2017,bera_emotionally_2019,arndt_safe_2015,fisac_probabilistically_2018,trautman_robot_2015,randhavane_pedestrian_2019,dugas_ian_2020,luber_socially-aware_2012,ferrer_robot_2013,sathyamoorthy_frozone_2020,park_hi_2016}, Hidden Markov Models \citep{vasquez_human_2013,peddi_data-driven_2020}, and social force models~\citep{ratsamee_social_2013,boldrer_multi-agent_2022} have also been applied for trajectory forecasting.
\par
Other approaches use additional information for trajectory prediction.
For instance, \cite{unhelkar_human-robot_2015} use turn indicators as features in the prediction of human motion trajectories.
In \citep{park_hi_2016} human intentions are classified and used to predict their motions.
\cite{chen_relational_2020} use a neural model (MLP) for predicting the next states of humans using the relations between agents predicted by a relational graph model.
\par
A subset of proposals predict the goal positions of humans instead of their trajectories \citep{bruckschen_human-aware_2020,ferrer_robot_2017} or along with them \citep{teja_s_hateb-2_2020,singamaneni_human-aware_2021,vemula_social_2018,kostavelis_robots_2017}. \cite{khambhaita_viewing_2017} use the goal positions of humans to predict their paths by means of elastic bands, but the goals are assumed.
Finally, it is worth mentioning that there is a group of proposals that, although they have not been classified in this taxon since they do not make predictions, use the past trajectories of pedestrians for different purposes \citep{bera_classifying_2018}.
\par
\textbf{Interactions} and \textbf{actions} constitute the next aspect of agents taken into account in our classification.
The term interaction can be found frequently in the literature, but, in some cases, it is used with a different meaning than the one we want to reflect here.
For instance, proposals related to SFM use the term interaction to refer to the influence of other agents on the dynamics of an agent.
Similarly, in the field of autonomous driving, the word interaction is commonly used to specify the mutual influence of two or more road users in their actions and reactions~\cite{wang_social_interactions_2022}.
In this review, we adopt a more intuitive interpretation of the term \textbf{interaction}: an intentional combined action between two or more agents, implying a collective behavior.
\par
We make a distinction between interactions that involve the robot (human-robot interactions) and those that are performed among other agents (human-human interactions).
Regarding the last group, in \citep{manso_graph_2019}, although no specific detection technique is used, the proposed model considers interactions between two people standing facing each other.
\cite{vega-magro_socially_2017} cluster individuals into groups according to their social interactions.
In \citep{truong_socially_2017} and \citep{truong_approach_2018} human group interactions are detected using a variant of the Graph Cuts of F-formations.
Other approaches detect and use human-robot interactions.
An example is the work in \cite{park_hi_2016}, which detects when a human is likely to interact with or obstruct the robot.
In other proposals, both human-human and human-robot interactions are considered.
Thus, \cite{lobato_human-robot_2019} consider human-human interactions, but also includes actions for human-robot interactions through a dialogue module.
Also, \cite{chen_relational_2020} propose a relational graph learning approach that uses GCNs to compute interaction features between humans and between humans and the robot.

The context provided by current activities and actions is also exploited in a subset of the works reviewed. Some of the planning-based approaches modify the robot's path based on the human actions like in \citep{mateus_efficient_2016} that consider activities like sitting/standing, and in \citep{charalampous_robot_2016} which selects a different set of actions: talking, walking, and working.

The detection of \textbf{intentions} of the agents is certainly an important feature in a socially aware navigation approach.
The ability to understand the intentions of the agents allows the robot to anticipate and timely adjust its behavior to the agents' preferences and actions.
We consider two different types of intentions: \textbf{\textit{expected}} and \textbf{\textit{unexpected}}.
Expected intentions are those that occur regularly in the context in which the navigation takes place.
A pedestrian's intent to cross the street is an example of expected intention.
On the contrary, unexpected intentions are linked to unusual attitudes/actions of the agents, but which could have a significant impact on navigation, such as for example, the intention to hinder any movement of the robot.
\par
All the papers detecting intentions included in our taxonomy can be classified into the first group (\textit{expected intentions}).
In addition, the intentions considered in some cases are closely related to trajectory prediction~\citep{ferrer_robot_2017,kostavelis_robots_2017}, interaction predisposition~\citep{ratsamee_social_2013,park_hi_2016} and interaction detection~\citep{park_hi_2016}.
Specifically, in the domain of autonomous vehicles, a variety of works targeting different kinds of intentions can be found.
That is the case of works estimating vehicles' predisposition to cooperate~\citep{kabtoul_proactive_2020,evens_learning_2021}, forecasting changes in vehicles' speeds and trajectories~\citep{chandra_forecasting_2020}, or predicting pedestrians intentions to cross~\citep{chandra_forecasting_2020}.

Differently, although linked to trajectory prediction, in \citep{mavrogiannis_social_2018} the proposed system reads signals of intentions or preferences over avoidance strategies.
Also, \cite{skrzypczyk_game_2021} detects and uses signals of intentions to cooperate with the robot.
Another differentiated approach to detecting agents' intentions for socially aware navigation is the one by \cite{cunningham_mpdm_2019}.
In this proposal, the system simulates forward the robot and the other agents under their assigned policies to obtain sequences of predicted states and observations.

Besides these aspects of the agents, some papers consider \textbf{other attributes}. 
For instance, \cite{bera_sociosense_2017} aim at estimating personality traits, \cite{bera_emotionally_2019} and \cite{jiang_interactive_2016} estimate the emotional states of the humans sharing the navigation area with the robot and make the robot act according to the detected emotions.
Another interesting factor proposed in the context of autonomous driving is \textit{altruism}~\citep{toghi_altruistic_2021}, which considers the performance of other vehicles. 
Similarly, \cite{toghi_social_2022} use the concepts of sympathy and cooperation. Specifically, sympathy is defined as the autonomous agent’s altruism toward a human and cooperation is the altruistic behavior among autonomous agents.
\par
Other works working with other geometrical information have been classified within \textit{other attributes}. This is the case of works that consider gestures~\citep{truong_toward_2017}, or the orientation of the humans, which is a proxy to their field of view~\citep{ratsamee_social_2013,truong_toward_2017,truc_khaos_2022}.

\subsection{Social norms}
The last main taxon for situation awareness and assessment deals with \textbf{social norms}, making a distinction between proxemics and other social rules.
\textbf{Proxemics} is one of the main elements considered in the vast majority of socially aware navigation proposals to make a robot behave more suitably when navigating around humans than it would do if using a more general navigation approach.
\par
To integrate the idea of proxemics into the robot's navigation system, some proposals consider a uniform circular area around humans that the robot must avoid traversing \citep{bruckschen_human-aware_2020,anvari_modelling_2020,araujo_architecture_2015,wang_adaptive_2016,peddi_data-driven_2020,korkmaz_human-aware_2021,nishimura_l2b_2020,kenk_human-aware_2019,chen_crowd-robot_2019,qiu_learning_2022}.
Other approaches model the space around humans using Gaussian functions to represent different degrees of discomfort based on proximity~\citep{lobato_human-robot_2019,singamaneni_human-aware_2021,vega-magro_socially_2017,chen_relational_2020,patompak_mobile_2016,rios-martinez_intention_2012,kostavelis_robots_2017,sorrentino_modeling_2021,truong_socially_2017,truong_approach_2018,kostavelis_human_2016, charalampous_robot_2016,vega-magro_flexible_2018,vega_socially_2019,mateus_efficient_2016,ratsamee_social_2013}.
In some cases, additional factors are considered when modelling personal spaces.
For instance, \cite{chen_relational_2020} consider a Gaussian variance proportional to the relative velocity of the person.
Also, in \citep{patompak_mobile_2016}, the space around humans is modeled as a 2D Gaussian function considering the gender, the social distance (familiar/strange), and the physical distance.
Other examples are the approaches in \citep{truong_approach_2018,mateus_efficient_2016} that take into consideration the status of a human (e.g. sitting, standing, moving) as well as their potential interactions with objects to represent their personal space.
Although not using a Gaussian modeling approach, other proposals build personal space around humans considering other factors as well, such as the person's emotion \citep{bera_emotionally_2019,jiang_interactive_2016} or the specific area where the person is located \citep{lu_towards_2013}.
In addition to being a frequently used tool for modeling people's personal space, Gaussian modeling has also been applied to estimate the interaction space of groups of people \citep{lobato_human-robot_2019,vega-magro_socially_2017,truong_socially_2017,truong_approach_2018,rios-martinez_intention_2012}.
\par
The inclusion of proxemics in many proposals on socially aware navigation focuses on defining forbidden or inappropriate areas for navigation that the robot must avoid crossing.
Nevertheless, there are approaches in the current literature in which proxemics is used to perform kinodynamic control.
Specifically, the speed of the robot is limited or modulated depending on the distance to people \citep{garrell_aerial_2017,carretero_comfort-oriented_2017,ferrer_robot_2017,teja_s_hateb-2_2020, singamaneni_human-aware_2021}, alleviating this way the freezing robot problem in complex situations.

Besides proxemics, some proposals consider \textbf{other social norms} during navigation. 
These additional social norms include walking on a specific side of the navigation area \citep{cunningham_mpdm_2019, mateus_efficient_2016} or passing a human from their conventionally preferred side \citep{dondrup_qualitative_2016,ciou_composite_2018, chen_socially_2017}.
In the work by~\cite{morales_social_2017}, along with navigating on a particular side, the proposed system is designed to avoid zigzag motion effects in order to improve the predictability of the autonomous vehicle.
\cite{khambhaita_viewing_2017} propose a planning technique based on a graph optimization approach that considers additional constraints along with proxemics such as directional constraints that penalize motions where humans and the robot are moving straight forward to each other.
The planning method proposed by \cite{bera_classifying_2018,bera_socially_2018} integrates the concept of entitativity to enhance social invisibility in multi-robot systems.
The norms to communicate crossing intention are considered in~\citep{hsu_pomdp_2020}.
In addition, the proposal by \cite{patompak_mobile_2016} considers the gender, the relative distance the robot percepts from the human, and the social distance, distinguishing between familiar humans and strangers, to assign an acceptable physical distance. 
Similarly, in \citep{kastner_enhancing_2022}, social norms are modified to suit three different age groups (child, adult and elder) and the robot is trained to handle such variations.
Finally, in a work by \cite{shahrezaie_advancing_2022}, the authors proposed different kinds of social interaction rules based on the subjective analysis of the data collected through interviews. 
These rules are then used to define different social behaviors for the robot.

\section{Evaluation}
\label{sec:evaluation}

\begin{figure*}[!ht]
    \centering
    \subfloat{
    \includegraphics[width=1.5\columnwidth]{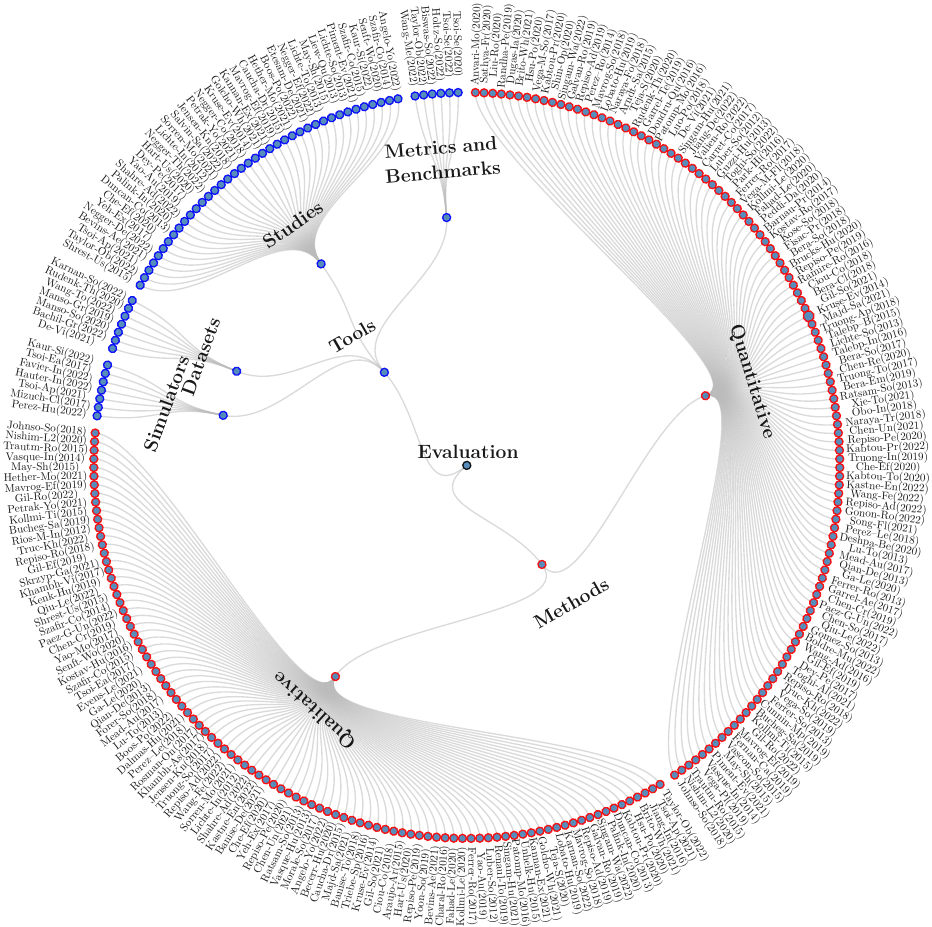}
    }
    \caption{Distribution of papers by evaluation and tools. The figure is best viewed zoomed in using a digital version.}
    \label{fig:eval_list}
\end{figure*}

This section discusses the analysis of the papers according to the taxonomy of evaluation. The distribution of the articles as per this taxonomy is shown in Fig.~\ref{fig:eval_list}. 
For the evaluation, researchers employ either of the \textit{qualitative} or \textit{quantitative} methodology. In some cases both these methodologies are applied to gain deeper insights.
Among the papers in this survey, a major portion of \textit{tools} taxon is dedicated to \textit{studies} rather than \textit{datasets}, \textit{simulators}, and \textit{benchmarks}.

\subsection{Methods}
The subjective and multifaceted nature makes evaluating socially aware navigation challenging. As a result, different kinds of methodologies are frequently applied to understand the full picture. 
Based on the type of methodology employed, they can be broadly divided into two types, (i) \textit{qualitative} and (ii) \textit{quantitative}. The \textit{qualitative} method uses the non-numerical data to explain the behavior and the social-awareness of the robot, while the \textit{quantitative} method tries to provide more objective analysis based on numbers.

The \textbf{qualitative approach} of evaluating robot’s socially aware behavior provides some initial cues about navigational performance. This mode of evaluation is used by many researchers irrespective of whether it is a \textit{mobile robot}~\citep{qian_decision-theoretical_2013, araujo_architecture_2015, jiang_interactive_2016, khambhaita_assessing_2017, kenk_human-aware_2019, ga_learning_2020, repiso_peoples_2020, chen_crowd-robot_2019, banisetty_deep_2021, singamaneni_human-aware_2021, gil_social_2021, qiu_learning_2022}, \textit{wheelchair}~\citep{rios-martinez_intention_2012, vasquez_human_2013, johnson_socially-aware_2018, skrzypczyk_game_2021}, \textit{PMV}~\citep{hsu_pomdp_2020,  kabtoul_proactive_2020, evens_learning_2021, paez-granados_unfreezing_2022} or a \textit{drone}~\citep{yoon_socially_2019, truc_khaos_2022}. Due to the diverse nature of data and techniques it deals with, qualitative methodology is mostly subjective and we haven't found any articles in this survey that does objective analysis.
Typically, the path, trajectory and/or the velocity profiles of the robot and the human(s) are analyzed systematically and logical inferences are drawn~\citep{qian_decision-theoretical_2013, vasquez_inverse_2014, patompak_mobile_2016, khambhaita_assessing_2017, repiso_-line_2017, banisetty_towards_2018, ga_learning_2020, teja_s_hateb-2_2020, nishimura_l2b_2020}. They are sometimes accompanied by step-by-step analysis of a situation using screenshots of simulated~\citep{qian_decision-theoretical_2013, khambhaita_assessing_2017, repiso_peoples_2019, teja_s_hateb-2_2020, ga_learning_2020} and/or real world experiments~\citep{qian_decision-theoretical_2013, ratsamee_social_2013, araujo_architecture_2015, hsu_pomdp_2020, teja_s_hateb-2_2020, singamaneni_human-aware_2021, paez-granados_unfreezing_2022, repiso_adaptive_2022, singamaneni_watch_2022}. This kind of analysis focuses on minute aspects that affect human-robot interaction during navigation.  Sometimes, it can also include comparisons of the given navigation planner with some other planners~\citep{qian_decision-theoretical_2013, khambhaita_assessing_2017, chen_crowd-robot_2019, teja_s_hateb-2_2020, nishimura_l2b_2020, kabtoul_proactive_2020, qiu_learning_2022} followed by the explanations about the improvements or deteriorations. For example, \cite{khambhaita_assessing_2017} qualitatively compare their socially aware planning system with two other systems in various simulated scenarios with sets of screenshots and discusses the advantages of the proposed system. Qualitative methodology is highly useful during the initial stages of development and when standards are not defined, which is the case of socially aware navigation. 

The robot navigating and interacting with humans in the environment needs to have legible motion and acceptable behavior. These criteria are subjective to humans' experiences and cannot be analyzed comparatively as above. Thus, \textit{user studies} are conducted to collect data through questionnaires~\citep{lichtenthaler_increasing_2012, kruse_evaluating_2014, szafir_communicating_2015, morales_social_2017, jensen_knowing_2018, repiso_adaptive_2019, mavrogiannis_effects_2019, petrak_you_2021,  bevins_aerial_2021, dalmasso_human-robot_2021, repiso_adaptive_2022} or interviews~\citep{duncan_comfortable_2013, szafir_communicating_2015, cauchard_drone_2015, mead_autonomous_2017, senft_would_2020, sorrentino_modeling_2021, shahrezaie_advancing_2022} to analyze the humans’ experiences, expectations and perceptions. The data is then used to subjectively evaluate a specific system or the behavior of the robot. Some of the commonly used methods for this employ the Godspeed Questionnaire with Likert Scale~\citep{weiss2015meta, carpinella2017robotic} followed by the analysis of variance (ANOVA). This kind of analysis is also used in \textit{behavior studies}~\citep{lichtenthaler_increasing_2012, kruse_evaluating_2014, szafir_communication_2014, may_show_2015, morales_social_2017, jensen_knowing_2018, hart_using_2020, hetherington_mobile_2021, senft_would_2020} that provide useful insights while designing new socially aware navigation strategies and behaviors for a robot, car or a drone. Despite their remarkable usefulness, user studies require experiments with real humans followed by statistical analysis and are often not easy to replicate and organize.  In our survey, roughly half of the papers employ some kind of qualitative methodology during the evaluation and around 33 of them perform user studies.

\textbf{Quantitative} methods try to measure the effects~\citep{dugas_ian_2020, toghi_altruistic_2021, mavrogiannis_social_2018, peddi_data-driven_2020, hsu_pomdp_2020} or the performance~\citep{ga_learning_2020, repiso_adaptive_2019, singamaneni_human-aware_2021, teja_s_hateb-2_2020, nishimura_l2b_2020, shin_optimization-based_2020, boldrer_multi-agent_2022, chen_unified_2021} of the robot navigation strategies numerically and provide good {objective} means for evaluation. Since socially aware robot navigation will always have to satisfy the \textbf{navigation metrics}, a large number of works~\citep{sathyamoorthy_frozone_2020, chen_relational_2020, chen_crowd-robot_2019, brito_where_2021, fernandez_carmona_making_2019, guzzi_human-friendly_2013, johnson_socially-aware_2018, kollmitz_time_2015, vega-magro_flexible_2018, vega_socially_2019, singamaneni_watch_2022} involve such {objective} analysis while assessing their system and comparing it to other state-of-art systems. Such navigational metrics include success rate, path length, number of collisions, etc. Some of the commonly used navigation metrics are listed in the first row of Table~\ref{table_metrics}.

\begin{table*}[!th]
\caption{Different types of Quantitative metrics. Each metric in the table is for one complete trajectory executed (until an abort or end) to reach a goal.}
\label{table_metrics}
\centering
\begin{tabular}{|m{2cm}|m{13.5cm}|}
\hline
\textbf{Metric Type} & \textbf{Metrics} \\
\hline
\hline
Navigation & success, efficiency, collisions, time to reach the goal (or completion time), distance traveled (or path length), velocity and acceleration~\citep{chen_socially_2017, chen_crowd-robot_2019, bera_socially_2018, bera_emotionally_2019, sathyamoorthy_frozone_2020} \\
\hline
Discomfort & human-robot distance, number of social space intrusions (personal and interaction spaces), time spent in social spaces (personal and interaction spaces), human safety and comfort indices (SII, SGI, RMI), performance metrics~\citep{vega_planning_2019, kollmitz_learning_2020, truong_toward_2017, kostavelis_robots_2017, ferrer_social-aware_2013, talebpour_incorporating_2016, singamaneni_human-aware_2021,manso_graph_2019,bachiller_graph_2022} \\
\hline
Naturalness & average displacement error, final displacement error, non-linear displacement error, cumulative Heading changes \citep{vega_planning_2019, alahi_learning_2017}\\
\hline

\end{tabular}
\end{table*}

\par
The socially aware part, however, requires a different set of metrics that could quantify the \textit{social} quality of navigation and the humans’ discomfort. In this article, we call such a set of metrics \textbf{discomfort metrics} and they try to estimate how acceptable the robot’s motion around humans is. The commonly used metrics in this group and some variations are listed in the second row of Table~\ref{table_metrics}. Even after being an active field for over 20 years, most of these are still based on Hall’s Proxemics Theory. There are some metrics like time-to-collision (\textit{ttc})~\citep{biswas_socnavbench_2022} that are recently gaining more attention. From time to time, some researchers define specialized metrics combining various criteria. For example, the works of Repiso et al.~\citep{repiso_robot_2018, repiso_adaptive_2019, repiso_peoples_2020, repiso_adaptive_2022} define performance metrics that combine various discomfort measures to form a single metric. These performance metrics in each setting are then utilized to evaluate simulated as well as real-world experiments. Similar performance metrics are used in~\citep{ferrer_social-aware_2013, ferrer_robot_2017, garrell_aerial_2017, garrell_teaching_2019} as well. The works in~\citep{truong_approach_2018, truong_toward_2017} define a set of metrics to measure socially aware navigation at various interaction levels. The Social Individual Index (SII) is defined based on the proxemics theory to measure the comfort at the individual level of the human, while the Social Group Index (SGI) is defined to deal with human groups and human-object interactions that occur during the navigation of the robot. They use something similar to \textit{ttc}, called Relative Motion Index (RMI) to measure the relative motion between the human and the robot and state that a lower RMI value results in more acceptable robot navigation. Another comfort index called the Social Direction Index (SDI) is defined to evaluate the direction of approach when the robot approaches humans. The robot's behavior is evaluated based on how close the calculated SDI is to the defined maximum in the situation. It has to be noted that the maximum desired value of SDI changes from one situation to another and also depends on the number of humans. Learning-based discomfort metrics have also appeared recently in \citep{manso_graph_2019, bachiller_graph_2022}. 

Another set of metrics that are commonly used in human trajectory prediction~\citep{alahi_social_2016, bisagno_group_2018, manh_scene-lstm_2019, vemula_social_2018, song_human_2018} and sometimes in socially aware navigation planning are the similarity metrics. The similarity metrics are applied to the socially aware navigation systems when the robot tries to mimic or follow a human's trajectory (or behavior) like in the case of \citep{de_vicente_deepsocnav_2021, fahad_learning_2020, luber_socially-aware_2012}. Some of the works in socially aware robot navigation also measure the path irregularity by employing measures like counting the unnecessary heading or orientation changes~\citep{vega_socially_2019, vega_planning_2019}. The \textbf{similarity metrics} together with the \textbf{irregularity measures} are grouped together as \textbf{naturalness metrics} (like in \citep{gao_evaluation_2022}) in our work and are presented in the last row of Table~\ref{table_metrics}. All these metrics (navigation, discomfort, and naturalness) are usually well-defined with some analytical formulation. They could be calculated automatically in a large number of scenarios to determine the robustness, advantages, and limitations of the defined socially aware navigation scheme. The 
{objective} nature of this methodology also makes the comparisons between different systems simpler. Hence, quantitative methods are often used when a new socially aware navigational system is proposed. As seen from Fig.~\ref{fig:eval_list}, more than 100 articles in this survey were found to include some form of quantitative evaluation. Table~\ref{table_metrics} shows only some of the commonly used metrics. However, researchers define their own set of metrics from time to time like in \citep{cunningham_mpdm_2019, repiso_adaptive_2022, ferrer_robot_2017, paez-granados_unfreezing_2022, teja_s_hateb-2_2020, kabtoul_proactive_2020}.

In Table~\ref{table_metrics}, we present the metrics just for one trajectory to make the presentation homogeneous, but in reality, this may not be the only way they are used. For example, it is a common practice to define the rate of success, efficiency, and performance over a set of trajectories or goals~\citep{ferrer_robot_2013, ratsamee_social_2013, fisac_probabilistically_2018, bera_socially_2018, mavrogiannis_effects_2019, repiso_adaptive_2019, chen_relational_2020, xie_towards_2021, valiente_robustness_2022}. Sometimes, it is applied to collisions as well and the collision rates are compared~\citep{luo_porca_2018, chen_crowd-robot_2019, chen_relational_2020, ga_learning_2020, liu_robot_2020}. Even among the metrics that are calculated per trajectory, the metrics like velocity, acceleration, and human safety and comfort indices present evolution over time for better explanation~\citep{gomez_social_2013, kruse_evaluating_2014, truong_socially_2017, luo_porca_2018, truong_approach_2018, mavrogiannis_effects_2019, anvari_modelling_2020, singamaneni_human-aware_2021, truc_khaos_2022}. 

Although not very common, there are papers that use subjective analysis in quantitative evaluation. The works by \cite{manso_graph_2019, manso_socnav1_2020} use subjective scores provided by human users to create general metrics that allow different types of navigation strategies in robots to be compared.
Finally, it should also be noted that the researchers need not have to employ only one kind of analysis. For instance, the works in \citep{qian_decision-theoretical_2013, patompak_mobile_2016, repiso_adaptive_2019, wang_feedback_2022, kastner_enhancing_2022, mavrogiannis_effects_2019, gil_social_2021} use both quantitative and qualitative evaluation while some works like~\citep{deshpande_behavioral_2020, gonon_robots_2022, kabtoul_proactive_2022} employ only quantitative evaluation. 

\subsection{Tools}
\textbf{Tools} that assist the development and evaluation of a socially aware navigation system fall under this taxon. Although there is no restriction what can be considered a tool, we have identified that \textit{studies, simulators, datasets, benchmarks} and \textit{new metrics} are generally used by researchers in socially aware robot navigation.

\par
\textbf{Studies} can be seen as one of the powerful tools that help the field progress by providing deeper understanding and useful information. For example, the papers studying communication strategies~\citep{szafir_communication_2014, may_show_2015, morales_social_2017, che_efficient_2020, hart_using_2020, bevins_aerial_2021, kannan_external_2021, senft_would_2020, boos_polite_2022, angelopoulos_you_2022} show different ways of expressing intention and how they are perceived by humans. The studies and surveys on pedestrian-vehicle interactions~\citep{ridel_literature_2018, rasouli_autonomous_2020, predhumeau_agent-based_2021} provide necessary information to design or improvise interaction strategies for autonomous vehicles and robots, while the studies on legibility~\citep{lichtenthaler_increasing_2012, kruse_evaluating_2014, hart_using_2020,hetherington_mobile_2021, taylor_observer_2022, neggers_theeffect_2022} show how the designed strategies are assessed by humans. These {studies} on human-robot interaction and navigation~\citep{lichtenthaler_increasing_2012, may_show_2015, morales_social_2017, yeh_exploring_2017, hetherington_mobile_2021, senft_would_2020, salvini_safety_2022, palinko_intention_2020} are great tools to design human complaint behaviors of the robot while it's navigating or interacting. For example, the work in \citep{shahrezaie_advancing_2022} uses the user study to design different social behaviors for the robots. \cite{neggers_comfortable_2018, neggers_determining_2022, neggers_theeffect_2022} through a series of studies provided details on comfortable passing distances and speeds for different type of robots around humans.

\textbf{Simulators} are the other tools that greatly aid the development process and help to test the system under various settings, quickly and efficiently. They can be used to simulate different human-robot navigation scenarios~\citep{kaur_simulators_2022} with varying densities of humans to challenge the socially aware navigation system before its final deployment in the real world. Researchers have recently recognized the importance and usefulness of human simulations and proposed various methodologies and approaches (e.g. Pedsim, ORCA) to include human agents in robotic simulators. Although there are many simulators for simulating robots, only a few simulators support HRI~\citep{kaur_simulators_2022}. Recently, some new simulation tools were proposed by Tsoi et al.~\citep{tsoi_approach_2021, tsoi_sean_2020, tsoi_sean_2022} and \cite{mizuchi_cloud-based_2017} that allow easy data collection and evaluation with simulated human agents or avatars. The works in \citep{tsoi_approach_2021, tsoi_sean_2020, tsoi_sean_2022} focus on simulating semi-crowded or crowded navigational scenarios while that in \citep{mizuchi_cloud-based_2017} focuses on multi-modal interactions. Besides, there are some interesting simulators~\citep{favier_intelligent_2022, hauterville_interactive_2022, perez_hunavsim_2022} that allow the simulation of intelligent human agents with multiple behaviors in small numbers.

Regarding \textbf{datasets}, frequently human-human navigation datasets (e.g. ETH, UCY, etc.) have been used to test how close the robot’s navigation behavior is to one of the humans in the datasets. Nevertheless, humans do not essentially perceive the robot the same way they perceive another human.
Hence, some recent datasets like TH\"{O}R~\citep{rudenko_thor_2020} and SCAND~\citep{karnan_socially_2022} record the data of the robot navigating in the presence of humans. TH\"{O}R's data is obtained from a controlled indoor environment whereas SCAND contains data from both indoors and outdoors. As these datasets contain the natural reactions of the people towards a robot navigating in their environment, it could help researchers to understand human-robot navigation better and design mechanisms that incorporate this information. The datasets SocNav1~\citep{manso_socnav1_2020} and SocNav2~\citep{bachiller_graph_2022} propose a new approach using graph neural networks~\citep{manso_graph_2019} to learn socially aware navigation conventions by using human feedback in simulated environments. Some recent pedestrian datasets like \citep{wang_towards_2022} provide enriched navigation information, including the first-person view, which is more natural compared to the classical top-down view. Further, all these datasets could also be employed to benchmark socially aware navigation systems~\citep{jiang_design_2022}. 

\par
\textbf{Benchmarks and metrics} are required to enable every socially aware robot planning system to have some minimum standards before they could be deployed among humans. The growing interest in the field has led to the development of some benchmarking tools like SocialGym~\citep{holtz_socialgym_2022}, Socnavbench~\citep{biswas_socnavbench_2022} and SEAN~\citep{tsoi_sean_2020, tsoi_sean_2022}. These benchmarks provide some performance metrics that could be used to compare different navigation frameworks numerically. The new version of SEAN~\citep{tsoi_sean_2022} integrates Socnavbench and provides some rich environments to learn or test socially aware robot navigation strategies and collect data. A very recent work by \cite{wang_metrics_2022} proposes protocols for benchmarking crowd navigation algorithms with a set of metrics.

\section{Proposals}
\label{sec:proposals}

The presented analysis provides a wide panoramic view of the current state of socially aware robot navigation. Although other recent surveys cover most of the relevant aspects involved in this field, they miss some key elements, such as the different types of robots and how their specific characteristics may affect navigation strategies. 
The effects of social context and the semantics of the environment are also frequently missing.
The proposed taxonomy intends to fill those gaps, including all the items that could be present in any existing and future approach to socially aware robot navigation.
Additionally, this survey has been designed to learn about \textit{what has been done in the field} and to \textit{identify areas that need more exploration and research}. Regarding this last point and taking our analysis into consideration, in this section, we put forward several proposals to enhance the current state of socially aware robot navigation.

\par 
\textbf{Proposal 1:} \emph{Enhance human models to improve robot behaviors in socially aware navigation, including human intention prediction.}
\\For a robot to select the actions to reach a goal, they require information about its state and its environment. The robot’s state is generally composed of its pose (\textit{i.e.}, position and orientation) and its current velocity with the occasional inclusion of acceleration data~\citep{gul2019comprehensive}. The environmental information is generally limited to the area that can be traversed by the robot; either using range or vision sensors, an obstacle map, or both. This information is typically enough for robots to navigate in human-free environments. socially aware robot navigation, however, requires information about humans, which are treated as special agents in the environment that robots should not collide with. Although most of the works reviewed in this article model humans using their instantaneous position only, some papers consider their speed, some consider human intentions to interact~\citep{park_hi_2016, ratsamee_social_2013} and some others include immediate motion intent predictions~\citep{peddi_data-driven_2020, hsu_pomdp_2020} to improve the social behavior of the robot. Even though extending the variables considered to optimize robot behavior is positive, this information is still insufficient to build good human models. 
For instance, modeling the reactions of humans or the possible mental states involves a lot of uncertainty and variables like positions, velocities or motion intention cannot capture them efficiently. They require more information on action-reaction cycles (through studies) and human psychological models.
This limitation on the human models directly affects the modeling of the robot’s planning and contributes to unexpected robot behavior from time to time.

\par
\textbf{Proposal 2:} \textit{Design user-adaptive robot behaviors. It requires significant improvement in the robot's perception of humans.} 
\\Further refinement in the robot’s social behavior can be realized if the humans can be identified. The perception modules of robots can provide information about certain human characteristics (age, gender, height, etc.) that can be used to filter or shift the robot's behavior according to the type of human it is interacting with~\citep{bera_emotionally_2019, jiang_interactive_2016, patompak_mobile_2016}. With the ever-expanding applications of Artificial Intelligence in the computer vision community, we believe that the online realization of adaptive robot behaviors is not very far.

\par
\textbf{Proposal 3:} \textit{Define context-based benchmarks and set up universal standards in the field. A set of standard contexts and human actions needs to be identified for benchmarking.}
\\The dependency on context cannot be neglected anymore in socially aware robot navigation. Each context requires different types of behaviors and interactions. Even within the same environment, depending on the action a human is performing (for example walking leisurely, rushing to a place, running, approaching a place, etc.) the interactions change and robots should adapt to this. Human action recognition is another active field in computer vision that could soon provide robotic systems with enough information to handle the interactions better. The lack of universal standards and benchmarks makes it very hard to compare different socially aware navigation algorithms. 

\par
\textbf{Proposal 4:} \textit{Focus on robot-specific parameters (e.g., shape, size) that can result in better interactions and develop strategies that can adapt to different robot characteristics.}
\\Studies have shown that the characteristics of robots affect interaction preferences \citep{golchoubian_what_2021, rasouli_autonomous_2020, samarakoon2022review}. Surprisingly, these parameters are disregarded by most socially aware navigation algorithms.

\par
\textbf{Proposal 5:} \textit{Establish good communication protocols to convey the robot's intention.}
\\
It has been shown that conveying the robot's intention has a positive impact on human-robot interaction~\citep{senft_would_2020}.
Implicit communication signals inspired by vehicles and/or humans and acknowledgment of human implicit communication have already been studied in some of the reviewed works~\citep{che_efficient_2020, hsu_pomdp_2020, szafir_communication_2014, singamaneni_human-aware_2021}. Some works like~\citep{bevins_aerial_2021, hart_using_2020, hetherington_mobile_2021, toghi_altruistic_2021} employ explicit communication strategies to improve navigation and interaction. However, these are limited and there is very little research on timing and the means to communicate. This highlights the need for more user studies on interaction and communication strategies.

\par 
\textbf{Proposal 6:} \textit{Explore alternatives to proxemics-based metrics. Well-designed human-robot interaction studies can provide clues about additional factors that affect human comfort around robots.}
\\The evaluation of socially aware navigation has always been challenging. Most discomfort metrics are based on proxemics theory and are not valid in many situations. Because of the lack of a clear rationale behind the relevance of such metrics, researchers tend to propose their own variations of metrics that are more suited to evaluate their systems. Further, the thresholds of the metrics are dependent on the situation, which makes it hard to define unified standards for socially aware navigation. All these issues make it difficult to compare the \textit{social-awareness} of different socially aware robot navigation algorithms, and until recently, there were no datasets or tools to benchmark socially aware navigation. The use of human-human interaction data to compare human-robot interactions is not always advisable and may result in false conclusions. Even though current benchmarks provide rich environments to test different frameworks under similar conditions, they use metrics that are not applicable to all human-robot navigation settings.

\par
\textbf{Proposal 7:} \textit{Prioritize considerations of comfort and trust in addition to safety during the design of socially aware robot navigation approaches.}
\\While significant research efforts have been devoted to ensuring safety in socially aware navigation across the different types of robots, there has been a notable gap in addressing comfort and trust.  
However, the critical aspects of comfort and trust, such as the prediction of the robot's motion, transparent decision-making, and reliable behavior, have received comparatively limited attention~\citep{ferrer_robot_2017, truong_socially_2017, che_efficient_2020}.
Further research efforts are needed to comprehensively address comfort and trust in socially aware navigation, considering the specific requirements across different types of robots.

\par
\textbf{Proposal 8:} \textit{Include the dynamic models of the other agents into the planning scheme.}
\\In relation to the \textit{dynamics of the physical motion}, most of the works done up to date consider that human and robot move slowly. As socially aware navigation progresses this assumption may become more problematic, especially when considering entities, such as  bicycles, motorcycles, or cars. Higher speeds make crucial for robots to account for these dynamics. The presence of fast-moving entities in an urban area introduces additional complexities and potential risks that the robot must consider when generating its plans and actions.

\section{Future Challenges}
\label{sec:future}

All the proposals presented in the previous section can be considered achievable given the current state of the field. However, a number of challenges that will require further long-term research to be addressed and resolved remain. This section discusses some of these \textbf{future challenges}, including aspects that range from technological matters to regulations and considerations about the design of urban spaces.

The \textbf{relationship between humans and robots} poses unresolved issues despite established social conventions. Understanding various aspects of this relationship is essential for robots to effectively assist humans. They need to comprehend typical human behaviors, predict actions, and actively engage in cooperative tasks.
First, \textit{cultural differences can influence social norms}. In the pursuit of enhancing socially aware navigation, conducting research on social norms across diverse cultures is indispensable.
Therefore, it is crucial to study how these norms affect navigational behaviors. Second, \textit{individual preferences} influence human behavior and perception. To ensure successful and comfortable interactions with humans, it is essential to design socially aware robots with this key aspect in mind.
The inclusion of this feature will play a vital role in fostering human acceptance of personal robots in the future. Third, a socially aware navigation system has to guarantee human \textit{acceptability, safety, trust, and privacy}.
Fourth, to effectively collaborate with people in tasks like navigating alongside them, robots need to estimate the \textit{intentions of humans} being accompanied. Understanding and interpreting human intentions play a crucial role in ensuring seamless coordination and cooperation between robots and humans.
While some research, including the Perception-Intention-Action model~\citep{dominguez2023}, has been conducted on Human-Robot Cooperation tasks, the perception and understanding of human intentions remain, and that necessitates further investigation.
Fifth, the occurrence of \textit{abnormal human behaviors towards robots}, such as vandalism or attempts to obstruct robot navigation should also be considered. In such cases, robots will need to adapt their navigation strategies based on the observed human behaviors. This adaptive approach is essential to ensure the safety and functionality of the robot in dynamic and unpredictable environments.

\textbf{Urban design} constitutes another group of important challenges. Cities and urban areas are not prepared for the deployment of robots in open areas and buildings.
Urban typology, like wide streets in new cities or narrow streets in historic districts, may necessitate adjustments in the type of robots allowed to circulate and the corresponding regulations.
The segregation of areas where robots can freely navigate is an important issue, particularly in the context of personal robots or robots involved in goods delivery. Establishing clear boundaries or designated zones for robot movement becomes increasingly relevant to ensure the efficient and safe operation of these robots. Adequate segregation helps prevent unwanted interactions or conflicts between robots and humans, ultimately contributing to the seamless integration of robots into various domains.
Finally, personal mobility devices, bicycles, motorcycles, or cars share the urban space besides humans. These different kinds of \textit{agents} have particularities that will provoke new situations the robot has to take into account. 

Currently, most \textbf{urban regulations} prohibit the circulation of autonomous robots in urban areas. To enable the integration of goods delivery robots, specific regulations similar to those for autonomous cars need to be established. This regulatory framework is vital in addressing the challenges unique to these robots and ensuring their safe and efficient operation in urban environments. By developing appropriate regulations, we can facilitate the widespread deployment of personal and goods delivery robots, prioritizing public safety and societal acceptance. 

\par
The research challenges in terms of \textbf{AI and decision-making} are very diverse.
Significant advancements in \textit{learning} have been achieved, however, socially aware navigation is limited by the lack of sufficient work in the helper domains. For instance, effectively acquiring knowledge about human preferences, intentions, and social norms remains a prominent challenge that requires resolution. 
Two crucial challenges arise in the context of robot navigation tasks. First challenge is to comprehend the \textit{current situation} encountered by a robot, which is of utmost importance to complete the task safely. The detection and understanding of \textit{conflict states} is the second major challenge. Short-term predictions and identification of potential conflict states provide the necessary information to anticipate the evolution of the current situation within a few seconds. This anticipation allows the robot to generate new plans and actions, adopting a proactive approach.
\par
In tasks involving cooperation or negotiation with both a robot and a human, prediction skills play an important role in facilitating coordination and foreseeing possible future outcomes. By anticipating the task's dynamics, predictions can facilitate behavior adaptation, thereby enhancing effective coordination.
This also applies to trajectory planning, which necessitates reasoning about multiple aspects in the present and the future, including the environment, the potential consequences of each agent's actions, and the expected behavior of human individuals. By taking into account these variables, robots can make well-informed decisions. 

Last but not least, \textbf{evaluating the efficacy and efficiency} of socially aware robot navigation systems is one of the most difficult tasks. Traditional metrics for measuring robot navigation, such as path length or collision avoidance, may not represent the social side of interactions properly. Subjective aspects such as user experience, social acceptance, and perceived trustworthiness must be considered when evaluating socially aware robot navigation. It is important to develop strong assessment procedures that include these social characteristics  to guarantee that social robots navigate in a way that is consistent with human expectations and promotes successful encounters.

\section{Conclusions}
\label{sec:conclusions}

The growing use of service and assistive robots as well as autonomous cars in human environments has made socially aware robot navigation an important research subject. The growing number of papers on socially aware robot navigation over the past six years is evidence of its relevance. Robots must be sociable or artificially sociable in order to be accepted by people.
To evaluate the field from different angles, we analyzed 193 articles, classifying them into different taxa spanning across four different faceted taxonomies. This taxonomic analysis allowed us to identify the areas that require more attention and further research.
Although the survey includes socially aware navigation for autonomous cars and drones, the main domain of interest in the papers found remains mobile robots.

\par
Most of the previous surveys and other prior studies in the field are referenced in our study. Each of these studies has added to our understanding of the navigation of social robots that are aware of humans from several angles, including proxemics, planning, perception, mapping, and evaluation techniques. By offering taxonomy-based classifications of the publications in our survey, we believe we have contributed to this body of knowledge.

Future research has fresh chances and challenges as the area expands. The requirement to create more reliable algorithms and techniques that can precisely forecast and adapt to human behavior in dynamic contexts is one such difficulty. In order to properly communicate the robot's goals and behaviors to humans, techniques making use of multiple communication modalities must be developed. There is also a need to address the ethical and legal issues surrounding interactions between humans and robots as the usage of robots in healthcare, education, and other social domains expands.


\section*{Declaration of conflicting interests}
The authors declare that there is no conflict of interest.

\section*{Funding}
This work was supported by the Horizon Europe Framework
Programme [Grant: euROBIN 101070596]; Agence
Nationale de la Recherche [Grant: ANITI ANR-19-PI3A-0004]; European project CANOPIES [Grant: H2020- ICT-2020-2-101016906]; Spanish Government (MCIN/AEI) and European Union (``NextGenerationEU''/PRTR) [Grant: TED2021-131739-C22]; Spanish Government (MCIN) [Grant: PDC2022-133597-C41];  Spanish Government (MCIN) [Grant: PID2022-142039NA-I00].

\bibliographystyle{SageH}
\setcitestyle{square,numbers,comma}
\bibliography{references}

\begin{thebibliography}{247}
\providecommand{\natexlab}[1]{#1}
\providecommand{\url}[1]{\texttt{#1}}
\providecommand{\urlprefix}{URL }
\expandafter\ifx\csname urlstyle\endcsname\relax
  \providecommand{\doi}[1]{DOI:\discretionary{}{}{}#1}\else
  \providecommand{\doi}{DOI:\discretionary{}{}{}\begingroup \urlstyle{rm}\Url}\fi

\bibitem[{Alahi et~al.(2016)Alahi, Goel, Ramanathan, Robicquet, Fei-Fei and Savarese}]{alahi_social_2016}
Alahi A, Goel K, Ramanathan V, Robicquet A, Fei-Fei L and Savarese S (2016) Social {LSTM}: Human trajectory prediction in crowded spaces.
\newblock In: \emph{Proceedings of the IEEE conference on computer vision and pattern recognition}. pp. 961--971.

\bibitem[{Alahi et~al.(2017)Alahi, Ramanathan, Goel, Robicquet, Sadeghian, Fei-Fei and Savarese}]{alahi_learning_2017}
Alahi A, Ramanathan V, Goel K, Robicquet A, Sadeghian AA, Fei-Fei L and Savarese S (2017) Learning to predict human behavior in crowded scenes.
\newblock In: \emph{Group and Crowd Behavior for Computer Vision}. Elsevier, pp. 183--207.

\bibitem[{Angelopoulos et~al.(2022)Angelopoulos, Rossi, Di~Napoli and Rossi}]{angelopoulos_you_2022}
Angelopoulos G, Rossi A, Di~Napoli C and Rossi S (2022) You are in my way: Non-verbal social cues for legible robot navigation behaviors.
\newblock In: \emph{2022 IEEE/RSJ International Conference on Intelligent Robots and Systems (IROS)}. IEEE, pp. 657--662.

\bibitem[{Anvari and Wurdemann(2020)}]{anvari_modelling_2020}
Anvari B and Wurdemann HA (2020) Modelling social interaction between humans and service robots in large public spaces.
\newblock In: \emph{2020 IEEE/RSJ International Conference on Intelligent Robots and Systems (IROS)}. IEEE, pp. 11189--11196.

\bibitem[{Araujo et~al.(2015)Araujo, Caminhas and Pereira}]{araujo_architecture_2015}
Araujo AR, Caminhas DD and Pereira GA (2015) An architecture for navigation of service robots in human-populated office-like environments.
\newblock \emph{IFAC-PapersOnLine} 48: 189--194.

\bibitem[{Arndt and Berns(2015)}]{arndt_safe_2015}
Arndt M and Berns K (2015) Safe predictive mobile robot navigation in aware environments.
\newblock In: \emph{2015 12th International Conference on Informatics in Control, Automation and Robotics (ICINCO)}, volume~2. IEEE, pp. 15--23.

\bibitem[{Bachiller et~al.(2022)Bachiller, Rodriguez-Criado, Jorvekar, Bustos, Faria and Manso}]{bachiller_graph_2022}
Bachiller P, Rodriguez-Criado D, Jorvekar RR, Bustos P, Faria DR and Manso LJ (2022) A graph neural network to model disruption in human-aware robot navigation.
\newblock \emph{Multimedia Tools and Applications} 81(3): 3277--3295.

\bibitem[{Banisetty and Feil-Seifer(2018)}]{banisetty_towards_2018}
Banisetty SB and Feil-Seifer D (2018) Towards a {Unified} {Planner} {For} {Socially}-{Aware} {Navigation}.
\newblock \emph{arXiv:1810.00966 [cs]} .

\bibitem[{Banisetty et~al.(2021)Banisetty, Rajamohan, Vega and Feil-Seifer}]{banisetty_deep_2021}
Banisetty SB, Rajamohan V, Vega F and Feil-Seifer D (2021) A {Deep} {Learning} {Approach} {To} {Multi}-{Context} {Socially}-{Aware} {Navigation}.
\newblock In: \emph{2021 30th {IEEE} {International} {Conference} on {Robot} \& {Human} {Interactive} {Communication} ({RO}-{MAN})}. IEEE, pp. 23--30.

\bibitem[{Barnaud et~al.(2014)Barnaud, Morgado, Palluel-Germain, Diard and Spalanzani}]{barnaud_proxemics_2014}
Barnaud ML, Morgado N, Palluel-Germain R, Diard J and Spalanzani A (2014) Proxemics models for human-aware navigation in robotics: Grounding interaction and personal space models in experimental data from psychology.
\newblock In: \emph{Proceedings of the 3rd IROS’2014 workshop “Assistance and Service Robotics in a Human Environment”}.

\bibitem[{Bartoli et~al.(2018)Bartoli, Lisanti, Ballan and Del~Bimbo}]{bartoli_context-aware_2018}
Bartoli F, Lisanti G, Ballan L and Del~Bimbo A (2018) Context-{Aware} {Trajectory} {Prediction}.
\newblock In: \emph{2018 24th {International} {Conference} on {Pattern} {Recognition} ({ICPR})}. pp. 1941--1946.

\bibitem[{Bayazit et~al.(2004)Bayazit, Lien and Amato}]{bayazit2004better}
Bayazit OB, Lien JM and Amato NM (2004) Better group behaviors using rule-based roadmaps.
\newblock \emph{Algorithmic Foundations of Robotics V} : 95--111.

\bibitem[{Becerra et~al.(2020)Becerra, Suomalainen, Lozano, Mimnaugh, Murrieta-Cid and LaValle}]{becerra_human_2020}
Becerra I, Suomalainen M, Lozano E, Mimnaugh KJ, Murrieta-Cid R and LaValle SM (2020) Human perception-optimized planning for comfortable vr-based telepresence.
\newblock \emph{IEEE Robotics and Automation Letters} 5(4): 6489--6496.

\bibitem[{Bera et~al.(2018{\natexlab{a}})Bera, Randhavane, Kubin, Wang, Gray and Manocha}]{bera_socially_2018}
Bera A, Randhavane T, Kubin E, Wang A, Gray K and Manocha D (2018{\natexlab{a}}) The socially invisible robot navigation in the social world using robot entitativity.
\newblock In: \emph{2018 ieee/rsj international conference on intelligent robots and systems (iros)}. IEEE, pp. 4468--4475.

\bibitem[{Bera et~al.(2019)Bera, Randhavane, Prinja, Kapsaskis, Wang, Gray and Manocha}]{bera_emotionally_2019}
Bera A, Randhavane T, Prinja R, Kapsaskis K, Wang A, Gray K and Manocha D (2019) The {Emotionally} {Intelligent} {Robot}: {Improving} {Social} {Navigation} in {Crowded} {Environments}.
\newblock \emph{arXiv:1903.03217 [cs]} .

\bibitem[{Bera et~al.(2017)Bera, Randhavane, Prinja and Manocha}]{bera_sociosense_2017}
Bera A, Randhavane T, Prinja R and Manocha D (2017) Sociosense: Robot navigation amongst pedestrians with social and psychological constraints.
\newblock In: \emph{2017 IEEE/RSJ International Conference on Intelligent Robots and Systems (IROS)}. IEEE, pp. 7018--7025.

\bibitem[{Bera et~al.(2018{\natexlab{b}})Bera, Randhavane, Wang, Manocha, Kubin and Gray}]{bera_classifying_2018}
Bera A, Randhavane T, Wang A, Manocha D, Kubin E and Gray K (2018{\natexlab{b}}) Classifying group emotions for socially-aware autonomous vehicle navigation.
\newblock In: \emph{Proceedings of the IEEE Conference on Computer Vision and Pattern Recognition Workshops}. pp. 1039--1047.

\bibitem[{Bevins and Duncan(2021)}]{bevins_aerial_2021}
Bevins A and Duncan BA (2021) Aerial {Flight} {Paths} for {Communication}: {How} {Participants} {Perceive} and {Intend} to {Respond} to {Drone} {Movements}.
\newblock In: \emph{Proceedings of the 2021 {ACM}/{IEEE} {International} {Conference} on {Human}-{Robot} {Interaction}}, {HRI} '21. Association for Computing Machinery, pp. 16--23.

\bibitem[{Bisagno et~al.(2018)Bisagno, Zhang and Conci}]{bisagno_group_2018}
Bisagno N, Zhang B and Conci N (2018) Group {LSTM}: Group trajectory prediction in crowded scenarios.
\newblock In: \emph{Proceedings of the European conference on computer vision (ECCV) workshops}. pp. 0--0.

\bibitem[{Biswas et~al.(2022)Biswas, Wang, Silvera, Steinfeld and Admoni}]{biswas_socnavbench_2022}
Biswas A, Wang A, Silvera G, Steinfeld A and Admoni H (2022) Socnavbench: A grounded simulation testing framework for evaluating social navigation.
\newblock \emph{ACM Transactions on Human-Robot Interaction (THRI)} 11(3): 1--24.

\bibitem[{Boldrer et~al.(2022)Boldrer, Antonucci, Bevilacqua, Palopoli and Fontanelli}]{boldrer_multi-agent_2022}
Boldrer M, Antonucci A, Bevilacqua P, Palopoli L and Fontanelli D (2022) Multi-agent navigation in human-shared environments: A safe and socially-aware approach.
\newblock \emph{Robotics and Autonomous Systems} 149: 103979.

\bibitem[{Boos et~al.(2022)Boos, Zimmermann, Zych and Bengler}]{boos_polite_2022}
Boos A, Zimmermann M, Zych M and Bengler K (2022) Polite and unambiguous requests facilitate willingness to help an autonomous delivery robot and favourable social attributions.
\newblock In: \emph{2022 31st IEEE International Conference on Robot and Human Interactive Communication (RO-MAN)}. IEEE, pp. 1620--1626.

\bibitem[{Brito et~al.(2021)Brito, Everett, How and Alonso-Mora}]{brito_where_2021}
Brito B, Everett M, How JP and Alonso-Mora J (2021) Where to go next: Learning a subgoal recommendation policy for navigation in dynamic environments.
\newblock \emph{IEEE Robotics and Automation Letters} 6: 4616--4623.

\bibitem[{Bruckschen et~al.(2020)Bruckschen, Bungert, Dengler and Bennewitz}]{bruckschen_human-aware_2020}
Bruckschen L, Bungert K, Dengler N and Bennewitz M (2020) Human-aware robot navigation by long-term movement prediction.
\newblock In: \emph{2020 IEEE/RSJ International Conference on Intelligent Robots and Systems (IROS)}. IEEE, pp. 11032--11037.

\bibitem[{Buchegger et~al.(2019)Buchegger, Todoran and Bader}]{buchegger_safe_2019}
Buchegger K, Todoran G and Bader M (2019) Safe and efficient autonomous navigation in the presence of humans at control level.
\newblock In: \emph{Advances in Service and Industrial Robotics: Proceedings of the 27th International Conference on Robotics in Alpe-Adria Danube Region (RAAD 2018)}. Springer, pp. 504--511.

\bibitem[{Carpinella et~al.(2017)Carpinella, Wyman, Perez and Stroessner}]{carpinella2017robotic}
Carpinella CM, Wyman AB, Perez MA and Stroessner SJ (2017) The robotic social attributes scale (rosas) development and validation.
\newblock In: \emph{Proceedings of the 2017 ACM/IEEE International Conference on human-robot interaction}. pp. 254--262.

\bibitem[{Carretero(2017)}]{carretero_comfort-oriented_2017}
Carretero V (2017) Comfort-oriented social force model and learned lessons : 8.

\bibitem[{Cauchard et~al.(2015)Cauchard, E, Zhai and Landay}]{cauchard_drone_2015}
Cauchard JR, E JL, Zhai KY and Landay JA (2015) Drone \& me: an exploration into natural human-drone interaction.
\newblock In: \emph{Proceedings of the 2015 {ACM} {International} {Joint} {Conference} on {Pervasive} and {Ubiquitous} {Computing} - {UbiComp} '15}. ACM Press, pp. 361--365.

\bibitem[{Chandra et~al.(2020)Chandra, Guan, Panuganti, Mittal, Bhattacharya, Bera and Manocha}]{chandra_forecasting_2020}
Chandra R, Guan T, Panuganti S, Mittal T, Bhattacharya U, Bera A and Manocha D (2020) Forecasting trajectory and behavior of road-agents using spectral clustering in graph-lstms.
\newblock \emph{IEEE Robotics and Automation Letters} 5: 4882--4890.

\bibitem[{Charalampous et~al.(2016)Charalampous, Kostavelis and Gasteratos}]{charalampous_robot_2016}
Charalampous K, Kostavelis I and Gasteratos A (2016) Robot navigation in large-scale social maps: {An} action recognition approach.
\newblock \emph{Expert Systems with Applications} 66: 261--273.

\bibitem[{Charalampous et~al.(2017)Charalampous, Kostavelis and Gasteratos}]{charalampous_recent_2017}
Charalampous K, Kostavelis I and Gasteratos A (2017) Recent trends in social aware robot navigation: A survey.
\newblock \emph{Robotics and Autonomous Systems} 93: 85--104.

\bibitem[{Che et~al.(2020)Che, Okamura and Sadigh}]{che_efficient_2020}
Che Y, Okamura AM and Sadigh D (2020) Efficient and {Trustworthy} {Social} {Navigation} {Via} {Explicit} and {Implicit} {Robot}-{Human} {Communication}.
\newblock \emph{IEEE Trans. Robot.} 36(3): 692--707.

\bibitem[{Chen et~al.(2020)Chen, Hu, Nikdel, Mori and Savva}]{chen_relational_2020}
Chen C, Hu S, Nikdel P, Mori G and Savva M (2020) Relational graph learning for crowd navigation.
\newblock In: \emph{2020 IEEE/RSJ International Conference on Intelligent Robots and Systems (IROS)}. IEEE, pp. 10007--10013.

\bibitem[{Chen et~al.(2019)Chen, Liu, Kreiss and Alahi}]{chen_crowd-robot_2019}
Chen C, Liu Y, Kreiss S and Alahi A (2019) Crowd-robot interaction: Crowd-aware robot navigation with attention-based deep reinforcement learning.
\newblock In: \emph{2019 international conference on robotics and automation (ICRA)}. IEEE, pp. 6015--6022.

\bibitem[{Chen et~al.(2017{\natexlab{a}})Chen, Liu and Zhang}]{chen2017deep}
Chen X, Liu Y and Zhang D (2017{\natexlab{a}}) A deep learning approach to generating natural language instructions for indoor robot navigation.
\newblock \emph{Robotics and Autonomous Systems} 95: 110--125.

\bibitem[{Chen and Lou(2022)}]{chen_unified_2021}
Chen Y and Lou Y (2022) A unified multiple-motion-mode framework for socially compliant navigation in dense crowds.
\newblock \emph{IEEE Transactions on Automation Science and Engineering} 19(4): 3536--3548.

\bibitem[{Chen et~al.(2017{\natexlab{b}})Chen, Everett, Liu and How}]{chen_socially_2017}
Chen YF, Everett M, Liu M and How JP (2017{\natexlab{b}}) Socially aware motion planning with deep reinforcement learning.
\newblock In: \emph{IEEE/RSJ IROS}. IEEE, pp. 1343--1350.

\bibitem[{Chik et~al.(2016)Chik, Yeong, Su, Lim, Subramaniam and Chin}]{chik_review_2016}
Chik S, Yeong C, Su E, Lim T, Subramaniam Y and Chin P (2016) A review of social-aware navigation frameworks for service robot in dynamic human environments.
\newblock \emph{Journal of Telecommunication, Electronic and Computer Engineering} 8: 41--50.

\bibitem[{Ciou et~al.(2018)Ciou, Hsiao, Wu, Tseng and Fu}]{ciou_composite_2018}
Ciou PH, Hsiao YT, Wu ZZ, Tseng SH and Fu LC (2018) Composite reinforcement learning for social robot navigation.
\newblock In: \emph{2018 IEEE/RSJ International Conference on Intelligent Robots and Systems (IROS)}. IEEE, pp. 2553--2558.

\bibitem[{Clarke et~al.(2015)Clarke, Braun and Hayfield}]{clarke2015thematic}
Clarke V, Braun V and Hayfield N (2015) Thematic analysis.
\newblock \emph{Qualitative psychology: A practical guide to research methods} 3: 222--248.

\bibitem[{Cunningham et~al.(2019)Cunningham, Galceran, Mehta, Ferrer, Eustice and Olson}]{cunningham_mpdm_2019}
Cunningham AG, Galceran E, Mehta D, Ferrer G, Eustice RM and Olson E (2019) Mpdm: multi-policy decision-making from autonomous driving to social robot navigation.
\newblock \emph{Control Strategies for Advanced Driver Assistance Systems and Autonomous Driving Functions: Development, Testing and Verification} : 201--223.

\bibitem[{Dalmasso et~al.(2021)Dalmasso, Garrell, Domínguez, Jiménez and Sanfeliu}]{dalmasso_human-robot_2021}
Dalmasso M, Garrell A, Domínguez JE, Jiménez P and Sanfeliu A (2021) Human-{Robot} {Collaborative} {Multi}-{Agent} {Path} {Planning} using {Monte} {Carlo} {Tree} {Search} and {Social} {Reward} {Sources}.
\newblock In: \emph{2021 {IEEE} {International} {Conference} on {Robotics} and {Automation} ({ICRA})}. pp. 10133--10138.

\bibitem[{de~Vicente and Soto(2021)}]{de_vicente_deepsocnav_2021}
de~Vicente JP and Soto A (2021) {DeepSocNav}: {Social} {Navigation} by {Imitating} {Human} {Behaviors}.
\newblock \emph{arXiv:2107.09170 [cs]} .

\bibitem[{Deo and Trivedi(2017)}]{deo_learning_2017}
Deo N and Trivedi MM (2017) Learning and predicting on-road pedestrian behavior around vehicles.
\newblock In: \emph{2017 IEEE 20th International Conference on Intelligent Transportation Systems (ITSC)}. IEEE, pp. 1--6.

\bibitem[{Deshpande et~al.(2020)Deshpande, Vaufreydaz and Spalanzani}]{deshpande_behavioral_2020}
Deshpande N, Vaufreydaz D and Spalanzani A (2020) Behavioral decision-making for urban autonomous driving in the presence of pedestrians using deep recurrent q-network.
\newblock In: \emph{2020 16th International Conference on Control, Automation, Robotics and Vision (ICARCV)}. IEEE, pp. 428--433.

\bibitem[{Dey and Terken(2017)}]{dey_pedestrian_2017}
Dey D and Terken J (2017) Pedestrian {Interaction} with {Vehicles}: {Roles} of {Explicit} and {Implicit} {Communication}.
\newblock In: \emph{Proceedings of the 9th {International} {Conference} on {Automotive} {User} {Interfaces} and {Interactive} {Vehicular} {Applications}}. ACM, pp. 109--113.

\bibitem[{Di~Carlo et~al.(2018)Di~Carlo, Wensing, Katz, Bledt and Kim}]{di2018dynamic}
Di~Carlo J, Wensing PM, Katz B, Bledt G and Kim S (2018) Dynamic locomotion in the mit cheetah 3 through convex model-predictive control.
\newblock In: \emph{2018 IEEE/RSJ international conference on intelligent robots and systems (IROS)}. IEEE, pp. 1--9.

\bibitem[{Dom{\'\i}nguez-Vidal et~al.(2023)Dom{\'\i}nguez-Vidal, Rodr{\'\i}guez and Sanfeliu}]{dominguez2023}
Dom{\'\i}nguez-Vidal JE, Rodr{\'\i}guez N and Sanfeliu A (2023) Perception-intention-action cycle as a human acceptable way for improving human-robot collaborative tasks.
\newblock In: \emph{Companion of the 2023 ACM/IEEE International Conference on Human-Robot Interaction}. pp. 567--571.

\bibitem[{Dondrup and Hanheide(2016)}]{dondrup_qualitative_2016}
Dondrup C and Hanheide M (2016) Qualitative constraints for human-aware robot navigation using {Velocity} {Costmaps}.
\newblock In: \emph{2016 25th {IEEE} {International} {Symposium} on {Robot} and {Human} {Interactive} {Communication} ({RO}-{MAN})}. IEEE, pp. 586--592.

\bibitem[{Dugas et~al.(2020)Dugas, Nieto, Siegwart and Chung}]{dugas_ian_2020}
Dugas D, Nieto J, Siegwart R and Chung JJ (2020) Ian: Multi-behavior navigation planning for robots in real, crowded environments.
\newblock In: \emph{2020 IEEE/RSJ International Conference on Intelligent Robots and Systems (IROS)}. IEEE, pp. 11368--11375.

\bibitem[{Duncan and Murphy(2013)}]{duncan_comfortable_2013}
Duncan BA and Murphy RR (2013) Comfortable approach distance with small {Unmanned} {Aerial} {Vehicles}.
\newblock In: \emph{2013 {IEEE} {RO}-{MAN}}. IEEE, pp. 786--792.

\bibitem[{Eiffert et~al.(2020)Eiffert, Li, Shan, Worrall, Sukkarieh and Nebot}]{eiffert_probabilistic_2020}
Eiffert S, Li K, Shan M, Worrall S, Sukkarieh S and Nebot E (2020) Probabilistic {Crowd} {GAN}: {Multimodal} {Pedestrian} {Trajectory} {Prediction} {Using} a {Graph} {Vehicle}-{Pedestrian} {Attention} {Network}.
\newblock \emph{IEEE Robot. Autom. Lett.} 5(4): 5026--5033.

\bibitem[{Ensley(1995)}]{ensley1995toward}
Ensley M (1995) Toward a theory of situation awareness in dynamic systems.
\newblock \emph{Human factors} 37: 32--64.

\bibitem[{Etesami et~al.(2021)Etesami, Nemati, Meghdari, Ge and Taheri}]{etesami_design_2021}
Etesami E, Nemati A, Meghdari AF, Ge SS and Taheri A (2021) Design and fabrication of a floating social robot: Ceb the social blimp.
\newblock In: \emph{Social Robotics: 13th International Conference, ICSR 2021, Singapore, Singapore, November 10--13, 2021, Proceedings 13}. Springer, pp. 660--670.

\bibitem[{Evens et~al.(2022)Evens, Schuurmans and Patrinos}]{evens_learning_2021}
Evens B, Schuurmans M and Patrinos P (2022) Learning {MPC} for interaction-aware autonomous driving: A game-theoretic approach.
\newblock In: \emph{2022 European Control Conference (ECC)}. IEEE, pp. 34--39.

\bibitem[{Fahad et~al.(2020)Fahad, Yang and Guo}]{fahad_learning_2020}
Fahad M, Yang G and Guo Y (2020) Learning human navigation behavior using measured human trajectories in crowded spaces.
\newblock In: \emph{2020 IEEE/RSJ International Conference on Intelligent Robots and Systems (IROS)}. IEEE, pp. 11154--11160.

\bibitem[{Favier et~al.(2022)Favier, Singamaneni and Alami}]{favier_intelligent_2022}
Favier A, Singamaneni PT and Alami R (2022) An intelligent human avatar to debug and challenge human-aware robot navigation systems.
\newblock In: \emph{2022 17th ACM/IEEE International Conference on Human-Robot Interaction (HRI)}. IEEE, pp. 760--764.

\bibitem[{Fernandez~Carmona et~al.(2019)Fernandez~Carmona, Parekh and Hanheide}]{fernandez_carmona_making_2019}
Fernandez~Carmona M, Parekh T and Hanheide M (2019) Making the {Case} for {Human}-{Aware} {Navigation} in {Warehouses}.
\newblock In: \emph{Towards {Autonomous} {Robotic} {Systems}}, volume 11650. Springer International Publishing, pp. 449--453.

\bibitem[{Ferrara and Rubagotti(2007)}]{ferrara2007sliding}
Ferrara A and Rubagotti M (2007) Sliding mode control of a mobile robot for dynamic obstacle avoidance based on a time-varying harmonic potential field.
\newblock In: \emph{ICRA 2007 Workshop: Planning, Perception and Navigation for Intelligent Vehicles}, volume 160. Citeseer.

\bibitem[{Ferrer et~al.(2013{\natexlab{a}})Ferrer, Garrell and Sanfeliu}]{ferrer_robot_2013}
Ferrer G, Garrell A and Sanfeliu A (2013{\natexlab{a}}) Robot companion: {A} social-force based approach with human awareness-navigation in crowded environments.
\newblock In: \emph{2013 {IEEE}/{RSJ} {International} {Conference} on {Intelligent} {Robots} and {Systems}}. pp. 1688--1694.

\bibitem[{Ferrer et~al.(2013{\natexlab{b}})Ferrer, Garrell and Sanfeliu}]{ferrer_social-aware_2013}
Ferrer G, Garrell A and Sanfeliu A (2013{\natexlab{b}}) Social-aware robot navigation in urban environments.
\newblock In: \emph{2013 {European} {Conference} on {Mobile} {Robots}}. pp. 331--336.

\bibitem[{Ferrer and Sanfeliu(2014)}]{ferrer2014proactive}
Ferrer G and Sanfeliu A (2014) Proactive kinodynamic planning using the extended social force model and human motion prediction in urban environments.
\newblock In: \emph{2014 IEEE/RSJ International Conference on Intelligent Robots and Systems}. IEEE, pp. 1730--1735.

\bibitem[{Ferrer et~al.(2017)Ferrer, Zulueta, Cotarelo and Sanfeliu}]{ferrer_robot_2017}
Ferrer G, Zulueta AG, Cotarelo FH and Sanfeliu A (2017) Robot social-aware navigation framework to accompany people walking side-by-side.
\newblock \emph{Autonomous robots} 41: 775--793.

\bibitem[{Fisac et~al.(2018)Fisac, Bajcsy, Herbert, Fridovich-Keil, Wang, Tomlin and Dragan}]{fisac_probabilistically_2018}
Fisac JF, Bajcsy A, Herbert SL, Fridovich-Keil D, Wang S, Tomlin CJ and Dragan AD (2018) Probabilistically {Safe} {Robot} {Planning} with {Confidence}-{Based} {Human} {Predictions}.
\newblock \emph{arXiv:1806.00109 [cs]} .

\bibitem[{Forer et~al.(2018)Forer, Banisetty, Yliniemi, Nicolescu and Feil-Seifer}]{forer_socially-aware_2018}
Forer S, Banisetty SB, Yliniemi L, Nicolescu M and Feil-Seifer D (2018) Socially-{Aware} {Navigation} {Using} {Non}-{Linear} {Multi}-{Objective} {Optimization}.
\newblock In: \emph{2018 {IEEE}/{RSJ} {International} {Conference} on {Intelligent} {Robots} and {Systems} ({IROS})}. IEEE, pp. 1--9.

\bibitem[{Galvan et~al.(2019)Galvan, Repiso and Sanfeliu}]{galvan_robot_2019}
Galvan M, Repiso E and Sanfeliu A (2019) Robot navigation to approach people using g2-spline path planning and extended social force models.
\newblock In: \emph{Robot 2019: {Fourth} {Iberian} {Robotics} {Conference}}. Springer International Publishing, pp. 15--27.

\bibitem[{Gao and Huang(2022)}]{gao_evaluation_2022}
Gao Y and Huang CM (2022) Evaluation of socially-aware robot navigation.
\newblock \emph{Frontiers in Robotics and AI} 8: 420.

\bibitem[{Garrell et~al.(2019)Garrell, Coll, Alquezar and Sanfeliu}]{garrell_teaching_2019}
Garrell A, Coll C, Alquezar R and Sanfeliu A (2019) Teaching a {Drone} to {Accompany} a {Person} from {Demonstrations} using {Non}-{Linear} {ASFM}.
\newblock In: \emph{2019 {IEEE}/{RSJ} {International} {Conference} on {Intelligent} {Robots} and {Systems} ({IROS})}. IEEE, pp. 1985--1991.

\bibitem[{Garrell et~al.(2017)Garrell, Garza-Elizondo, Villamizar, Herrero and Sanfeliu}]{garrell_aerial_2017}
Garrell A, Garza-Elizondo L, Villamizar M, Herrero F and Sanfeliu A (2017) Aerial social force model: {A} new framework to accompany people using autonomous flying robots.
\newblock In: \emph{2017 {IEEE}/{RSJ} {International} {Conference} on {Intelligent} {Robots} and {Systems} ({IROS})}. IEEE, pp. 7011--7017.

\bibitem[{Gil et~al.(2021)Gil, Garrell and Sanfeliu}]{gil_social_2021}
Gil O, Garrell A and Sanfeliu A (2021) Social {Robot} {Navigation} {Tasks}: {Combining} {Machine} {Learning} {Techniques} and {Social} {Force} {Model}.
\newblock \emph{Sensors} 21(21): 7087.

\bibitem[{Gil and Sanfeliu(2019)}]{gil_effects_2019}
Gil O and Sanfeliu A (2019) Effects of a {Social} {Force} {Model} {Reward} in {Robot} {Navigation} {Based} on {Deep} {Reinforcement} {Learning}.
\newblock In: \emph{Robot 2019: {Fourth} {Iberian} {Robotics} {Conference}}, Advances in {Intelligent} {Systems} and {Computing}. Springer International Publishing, pp. 213--224.

\bibitem[{Gil and Sanfeliu(2022)}]{gil_robot_2022}
Gil {\'O} and Sanfeliu A (2022) Robot navigation anticipative strategies in deep reinforcement motion planning.
\newblock In: \emph{ROBOT2022: Fifth Iberian Robotics Conference: Advances in Robotics, Volume 2}. Springer, pp. 67--78.

\bibitem[{Golchoubian et~al.(2021)Golchoubian, Ghafurian, Azad and Dautenhahn}]{golchoubian_what_2021}
Golchoubian M, Ghafurian M, Azad NL and Dautenhahn K (2021) What are {Social} {Norms} for {Low}-speed {Autonomous} {Vehicle} {Navigation} in {Crowded} {Environments}? {An} {Online} {Survey}.
\newblock In: \emph{Proceedings of the 9th {International} {Conference} on {Human}-{Agent} {Interaction}}. ACM, pp. 148--156.

\bibitem[{G{\'o}mez et~al.(2013)G{\'o}mez, Mavridis and Garrido}]{gomez_social_2013}
G{\'o}mez JV, Mavridis N and Garrido S (2013) Social path planning: Generic human-robot interaction framework for robotic navigation tasks.
\newblock In: \emph{2nd Intl. workshop on cognitive robotics systems: replicating human actions and activities}.

\bibitem[{Gonon et~al.(2022)Gonon, Paez-Granados and Billard}]{gonon_robots_2022}
Gonon DJ, Paez-Granados D and Billard A (2022) Robots' motion planning in human crowds by acceleration obstacles.
\newblock \emph{IEEE Robotics and Automation Letters} 7(4): 11236--11243.

\bibitem[{Gul et~al.(2019)Gul, Rahiman and Nazli~Alhady}]{gul2019comprehensive}
Gul F, Rahiman W and Nazli~Alhady SS (2019) A comprehensive study for robot navigation techniques.
\newblock \emph{Cogent Engineering} 6: 1632046.

\bibitem[{Guldenring et~al.(2020)Guldenring, G{\"o}rner, Hendrich, Jacobsen and Zhang}]{ga_learning_2020}
Guldenring R, G{\"o}rner M, Hendrich N, Jacobsen NJ and Zhang J (2020) Learning local planners for human-aware navigation in indoor environments.
\newblock In: \emph{2020 IEEE/RSJ International Conference on Intelligent Robots and Systems (IROS)}. IEEE, pp. 6053--6060.

\bibitem[{Guzzi et~al.(2013)Guzzi, Giusti, Gambardella, Theraulaz and Di~Caro}]{guzzi_human-friendly_2013}
Guzzi J, Giusti A, Gambardella LM, Theraulaz G and Di~Caro GA (2013) Human-friendly robot navigation in dynamic environments.
\newblock In: \emph{2013 {IEEE} {International} {Conference} on {Robotics} and {Automation}}. IEEE, pp. 423--430.

\bibitem[{Hart et~al.(2020)Hart, Mirsky, Xiao, Tejeda, Mahajan, Goo, Baldauf, Owen and Stone}]{hart_using_2020}
Hart J, Mirsky R, Xiao X, Tejeda S, Mahajan B, Goo J, Baldauf K, Owen S and Stone P (2020) Using human-inspired signals to disambiguate navigational intentions.
\newblock In: \emph{Social Robotics: 12th International Conference, ICSR 2020}. Springer, pp. 320--331.

\bibitem[{Hassanalian and Abdelkefi(2017)}]{hassanalian2017classifications}
Hassanalian M and Abdelkefi A (2017) Classifications, applications, and design challenges of drones: A review.
\newblock \emph{Progress in Aerospace Sciences} 91: 99--131.

\bibitem[{Hauterville et~al.(2022)Hauterville, Fern{\'a}ndez, Singamaneni, Favier, Matell{\'a}n and Alami}]{hauterville_interactive_2022}
Hauterville O, Fern{\'a}ndez C, Singamaneni PT, Favier A, Matell{\'a}n V and Alami R (2022) Interactive social agents simulation tool for designing choreographies for human-robot-interaction research.
\newblock In: \emph{ROBOT2022: Fifth Iberian Robotics Conference: Advances in Robotics, Volume 2}. Springer International Publishing Cham, pp. 514--527.

\bibitem[{Helbing and Molnar(1995)}]{helbing1995social}
Helbing D and Molnar P (1995) Social force model for pedestrian dynamics.
\newblock \emph{Physical review E} 51: 4282.

\bibitem[{Hetherington et~al.(2021)Hetherington, Lee, Haase, Croft and Machiel Van~der Loos}]{hetherington_mobile_2021}
Hetherington NJ, Lee R, Haase M, Croft EA and Machiel Van~der Loos HF (2021) Mobile {Robot} {Yielding} {Cues} for {Human}-{Robot} {Spatial} {Interaction}.
\newblock In: \emph{2021 {IEEE}/{RSJ} {International} {Conference} on {Intelligent} {Robots} and {Systems} ({IROS})}. IEEE, pp. 3028--3033.

\bibitem[{Holtz and Biswas(2022)}]{holtz_socialgym_2022}
Holtz J and Biswas J (2022) Socialgym: A framework for benchmarking social robot navigation.
\newblock In: \emph{2022 IEEE/RSJ International Conference on Intelligent Robots and Systems (IROS)}. IEEE, pp. 11246--11252.

\bibitem[{Honig et~al.(2018)Honig, Oron-Gilad, Zaichyk, Sarne-Fleischmann, Olatunji and Edan}]{honig_toward_2018}
Honig SS, Oron-Gilad T, Zaichyk H, Sarne-Fleischmann V, Olatunji S and Edan Y (2018) Toward socially aware person-following robots.
\newblock \emph{IEEE Transactions on Cognitive and Developmental Systems} 10: 936--954.

\bibitem[{Hsu et~al.(2020)Hsu, Gopalswamy, Saripalli and Shell}]{hsu_pomdp_2020}
Hsu YC, Gopalswamy S, Saripalli S and Shell DA (2020) A pomdp treatment of vehicle-pedestrian interaction: Implicit coordination via uncertainty-aware planning.
\newblock In: \emph{2020 IEEE/RSJ International Conference on Intelligent Robots and Systems (IROS)}. IEEE, pp. 1984--1991.

\bibitem[{Hua et~al.(2021)Hua, Zeng, Li and Ju}]{hua2021learning}
Hua J, Zeng L, Li G and Ju Z (2021) Learning for a robot: Deep reinforcement learning, imitation learning, transfer learning.
\newblock \emph{Sensors} 21(4): 1278.

\bibitem[{Huang et~al.(2016)Huang, Mutlu and Hong}]{huang2016machines}
Huang CM, Mutlu B and Hong YY (2016) When machines break the social norm: exploring the effects of norm violations by a social robot on its perceived sociability.
\newblock In: \emph{Proceedings of the 2016 CHI conference on human factors in computing systems}. ACM, pp. 1418--1429.

\bibitem[{Jensen et~al.(2018)Jensen, Hansen and Knoche}]{jensen_knowing_2018}
Jensen W, Hansen S and Knoche H (2018) Knowing {You}, {Seeing} {Me}: {Investigating} {User} {Preferences} in {Drone}-{Human} {Acknowledgement}.
\newblock In: \emph{Proceedings of the 2018 {CHI} {Conference} on {Human} {Factors} in {Computing} {Systems}}. Association for Computing Machinery, pp. 1--12.

\bibitem[{Jiang et~al.(2022)Jiang, Worrall and Shan}]{jiang_design_2022}
Jiang D, Worrall S and Shan M (2022) The design of a pedestrian aware contextual speed controller for autonomous driving.
\newblock In: \emph{2022 IEEE 25th International Conference on Intelligent Transportation Systems (ITSC)}. IEEE, pp. 3899--3906.

\bibitem[{Jiang et~al.(2016)Jiang, Ge, Tangirala and Lee}]{jiang_interactive_2016}
Jiang R, Ge SS, Tangirala NT and Lee TH (2016) Interactive navigation of mobile robots based on human’s emotion.
\newblock In: \emph{Social Robotics: 8th International Conference, ICSR 2016}. Springer, pp. 243--252.

\bibitem[{Johnson and Kuipers(2018)}]{johnson_socially-aware_2018}
Johnson C and Kuipers B (2018) Socially-{Aware} {Navigation} {Using} {Topological} {Maps} and {Social} {Norm} {Learning}.
\newblock In: \emph{Proceedings of the 2018 {AAAI}/{ACM} {Conference} on {AI}, {Ethics}, and {Society} - {AIES} '18}. ACM Press, pp. 151--157.

\bibitem[{Kabtoul(2021)}]{kabtoul_proactive_2022}
Kabtoul M (2021) \emph{Proactive and social navigation of autonomous vehicles in shared spaces}.
\newblock PhD Thesis, Universit{\'e} Grenoble Alpes.

\bibitem[{Kabtoul et~al.(2020{\natexlab{a}})Kabtoul, Martinet and Spalanzani}]{kabtoul_proactive_2020}
Kabtoul M, Martinet P and Spalanzani A (2020{\natexlab{a}}) Proactive longitudinal velocity control in pedestrians-vehicle interaction scenarios.
\newblock In: \emph{2020 IEEE 23rd International Conference on Intelligent Transportation Systems (ITSC)}. IEEE, pp. 1--6.

\bibitem[{Kabtoul et~al.(2020{\natexlab{b}})Kabtoul, Spalanzani and Martinet}]{kabtoul_towards_2020}
Kabtoul M, Spalanzani A and Martinet P (2020{\natexlab{b}}) Towards {Proactive} {Navigation}: {A} {Pedestrian}-{Vehicle} {Cooperation} {Based} {Behavioral} {Model}.
\newblock In: \emph{{ICRA} 2020 - {IEEE} {International} {Conference} on {Robotics} and {Automation}}, {IEEE} {International} {Conference} on {Robotics} and {Automation} {Proceedings}. pp. 6958--6964.

\bibitem[{Kabtoul et~al.(2022)Kabtoul, Spalanzani and Martinet}]{kabtoul_proactivesmooth_2022}
Kabtoul M, Spalanzani A and Martinet P (2022) Proactive and smooth maneuvering for navigation around pedestrians.
\newblock In: \emph{2022 International Conference on Robotics and Automation (ICRA)}. IEEE, pp. 4723--4729.

\bibitem[{Kannan et~al.(2021)Kannan, Lee and Min}]{kannan_external_2021}
Kannan SS, Lee A and Min BC (2021) External {Human}-{Machine} {Interface} on {Delivery} {Robots}: {Expression} of {Navigation} {Intent} of the {Robot}.
\newblock In: \emph{2021 30th {IEEE} {International} {Conference} on {Robot} \& {Human} {Interactive} {Communication} ({RO}-{MAN})}. IEEE, pp. 1305--1312.

\bibitem[{Karnan et~al.(2022)Karnan, Nair, Xiao, Warnell, Pirk, Toshev, Hart, Biswas and Stone}]{karnan_socially_2022}
Karnan H, Nair A, Xiao X, Warnell G, Pirk S, Toshev A, Hart J, Biswas J and Stone P (2022) Socially compliant navigation dataset ({SCAND}): A large-scale dataset of demonstrations for social navigation.
\newblock \emph{IEEE Robotics and Automation Letters} 7: 11807--11814.

\bibitem[{K{\"a}stner et~al.(2022)K{\"a}stner, Lil, Shen and Lambrecht}]{kastner_enhancing_2022}
K{\"a}stner L, Lil J, Shen Z and Lambrecht J (2022) Enhancing navigational safety in crowded environments using semantic-deep-reinforcement-learning-based navigation.
\newblock In: \emph{2022 IEEE International Symposium on Safety, Security, and Rescue Robotics (SSRR)}. IEEE, pp. 87--93.

\bibitem[{Kaur et~al.(2022)Kaur, Liu and Shi}]{kaur_simulators_2022}
Kaur P, Liu Z and Shi W (2022) Simulators for mobile social robots: State-of-the-art and challenges.
\newblock In: \emph{2022 Fifth International Conference on Connected and Autonomous Driving (MetroCAD)}. IEEE, pp. 47--56.

\bibitem[{Kenk et~al.(2019)Kenk, Hassaballah and Brethé}]{kenk_human-aware_2019}
Kenk M, Hassaballah M and Brethé JF (2019) Human-aware {Robot} {Navigation} in {Logistics} {Warehouses}:.
\newblock In: \emph{Proceedings of the 16th {International} {Conference} on {Informatics} in {Control}, {Automation} and {Robotics}}. SCITEPRESS - Science and Technology Publications, pp. 371--378.

\bibitem[{Khambhaita and Alami(2017{\natexlab{a}})}]{khambhaita_assessing_2017}
Khambhaita H and Alami R (2017{\natexlab{a}}) Assessing the social criteria for human-robot collaborative navigation: {A} comparison of human-aware navigation planners.
\newblock In: \emph{2017 26th {IEEE} {International} {Symposium} on {Robot} and {Human} {Interactive} {Communication} ({RO}-{MAN})}. IEEE, pp. 1140--1145.

\bibitem[{Khambhaita and Alami(2017{\natexlab{b}})}]{khambhaita_viewing_2017}
Khambhaita H and Alami R (2017{\natexlab{b}}) Viewing robot navigation in human environment as a cooperative activity.
\newblock In: \emph{Robotics Research: The 18th International Symposium ISRR}. Springer, pp. 285--300.

\bibitem[{Khambhaita et~al.(2016)Khambhaita, Rios-Martinez and Alami}]{khambhaita2016head}
Khambhaita H, Rios-Martinez J and Alami R (2016) Head-body motion coordination for human aware robot navigation.
\newblock In: \emph{9th International workshop on Human-Friendlly Robotics (HFR 2016)}. p.~8p.

\bibitem[{Kivrak et~al.(2018)Kivrak, Cakmak, Kose and Yavuz}]{kose_socially_2018}
Kivrak H, Cakmak F, Kose H and Yavuz S (2018) Socially aware robot navigation using the collision prediction based pedestrian model.
\newblock In: \emph{IEEE/RSJ IROS: Workshop on Robotic Co-workers}, volume~4.

\bibitem[{Kollmitz et~al.(2015)Kollmitz, Hsiao, Gaa and Burgard}]{kollmitz_time_2015}
Kollmitz M, Hsiao K, Gaa J and Burgard W (2015) Time dependent planning on a layered social cost map for human-aware robot navigation.
\newblock In: \emph{2015 {European} {Conference} on {Mobile} {Robots} ({ECMR})}. pp. 1--6.

\bibitem[{Kollmitz et~al.(2020)Kollmitz, Koller, Boedecker and Burgard}]{kollmitz_learning_2020}
Kollmitz M, Koller T, Boedecker J and Burgard W (2020) Learning human-aware robot navigation from physical interaction via inverse reinforcement learning.
\newblock In: \emph{IEEE/RSJ IROS}. IEEE, pp. 11025--11031.

\bibitem[{Korkmaz(2021)}]{korkmaz_human-aware_2021}
Korkmaz M (2021) Human-{Aware} {Dynamic} {Path} {Planning}.
\newblock In: \emph{2021 {International} {Conference} on {INnovations} in {Intelligent} {SysTems} and {Applications} ({INISTA})}. IEEE, pp. 1--5.

\bibitem[{Kostavelis et~al.(2016)Kostavelis, Giakoumis, Malassiotis and Tzovaras}]{kostavelis_human_2016}
Kostavelis I, Giakoumis D, Malassiotis S and Tzovaras D (2016) Human {Aware} {Robot} {Navigation} in {Semantically} {Annotated} {Domestic} {Environments}.
\newblock In: Antona M and Stephanidis C (eds.) \emph{Universal {Access} in {Human}-{Computer} {Interaction}. {Interaction} {Techniques} and {Environments}}, Lecture {Notes} in {Computer} {Science}. Springer International Publishing, pp. 414--423.

\bibitem[{Kostavelis et~al.(2017)Kostavelis, Kargakos, Giakoumis and Tzovaras}]{kostavelis_robots_2017}
Kostavelis I, Kargakos A, Giakoumis D and Tzovaras D (2017) Robot’s {Workspace} {Enhancement} with {Dynamic} {Human} {Presence} for {Socially}-{Aware} {Navigation}.
\newblock In: \emph{Computer {Vision} {Systems}}, Lecture {Notes} in {Computer} {Science}. Springer International Publishing, pp. 279--288.

\bibitem[{Kruse et~al.(2014)Kruse, Kirsch, Khambhaita and Alami}]{kruse_evaluating_2014}
Kruse T, Kirsch A, Khambhaita H and Alami R (2014) Evaluating directional cost models in navigation.
\newblock In: \emph{Proceedings of the 2014 {ACM}/{IEEE} international conference on {Human}-robot interaction - {HRI} '14}. ACM Press, pp. 350--357.

\bibitem[{Kruse et~al.(2013)Kruse, Pandey, Alami and Kirsch}]{kruse_human-aware_2013}
Kruse T, Pandey AK, Alami R and Kirsch A (2013) Human-aware robot navigation: A survey.
\newblock \emph{Robotics and Autonomous Systems} 61: 1726--1743.

\bibitem[{Li et~al.(2019)Li, Zhang and Zhao}]{li2019deep}
Li H, Zhang Q and Zhao D (2019) Deep reinforcement learning-based automatic exploration for navigation in unknown environment.
\newblock \emph{IEEE transactions on neural networks and learning systems} 31(6): 2064--2076.

\bibitem[{Liang et~al.(2018)Liang, Wang, Yang and Xing}]{liang2018cirl}
Liang X, Wang T, Yang L and Xing E (2018) Cirl: Controllable imitative reinforcement learning for vision-based self-driving.
\newblock In: \emph{Proceedings of the European conference on computer vision (ECCV)}. pp. 584--599.

\bibitem[{Lichtenth{\"a}ler and Kirsch(2013)}]{lichtenthaler_towards_2013}
Lichtenth{\"a}ler C and Kirsch A (2013) Towards legible robot navigation-how to increase the intend expressiveness of robot navigation behavior.
\newblock In: \emph{International Conference on Social Robotics-Workshop Embodied Communication of Goals and Intentions}.

\bibitem[{Lichtenth{\"a}ler et~al.(2013)Lichtenth{\"a}ler, Peters, Griffiths and Kirsch}]{lichtenthaler_social_2013}
Lichtenth{\"a}ler C, Peters A, Griffiths S and Kirsch A (2013) Social navigation-identifying robot navigation patterns in a path crossing scenario.
\newblock In: \emph{Social Robotics: 5th International Conference, ICSR 2013}. Springer, pp. 84--93.

\bibitem[{Lichtenthäler et~al.(2012)Lichtenthäler, Lorenz, Karg and Kirsch}]{lichtenthaler_increasing_2012}
Lichtenthäler C, Lorenz T, Karg M and Kirsch A (2012) Increasing perceived value between human and robots — {Measuring} legibility in human aware navigation.
\newblock In: \emph{2012 {IEEE} {Workshop} on {Advanced} {Robotics} and its {Social} {Impacts} ({ARSO})}. pp. 89--94.

\bibitem[{Liew and Yairi(2013)}]{liew_quadrotor_2013}
Liew CF and Yairi T (2013) Quadrotor or blimp? noise and appearance considerations in designing social aerial robot.
\newblock In: \emph{2013 8th ACM/IEEE International Conference on Human-Robot Interaction (HRI)}. IEEE, pp. 183--184.

\bibitem[{Liu et~al.(2020)Liu, Dugas, Cesari, Siegwart and Dub{\'e}}]{liu_robot_2020}
Liu L, Dugas D, Cesari G, Siegwart R and Dub{\'e} R (2020) Robot navigation in crowded environments using deep reinforcement learning.
\newblock In: \emph{2020 IEEE/RSJ International Conference on Intelligent Robots and Systems (IROS)}. IEEE, pp. 5671--5677.

\bibitem[{Lobato et~al.(2019)Lobato, Vega-Magro, Núñez and Manso}]{lobato_human-robot_2019}
Lobato C, Vega-Magro A, Núñez P and Manso L (2019) Human-robot dialogue and {Collaboration} for social navigation in crowded environments.
\newblock In: \emph{2019 {IEEE} {International} {Conference} on {Autonomous} {Robot} {Systems} and {Competitions} ({ICARSC})}. pp. 1--6.

\bibitem[{Lu and Smart(2013)}]{lu_towards_2013}
Lu DV and Smart WD (2013) Towards more efficient navigation for robots and humans.
\newblock In: \emph{2013 {IEEE}/{RSJ} {International} {Conference} on {Intelligent} {Robots} and {Systems}}. IEEE, pp. 1707--1713.

\bibitem[{Luber et~al.(2012)Luber, Spinello, Silva and Arras}]{luber_socially-aware_2012}
Luber M, Spinello L, Silva J and Arras KO (2012) Socially-aware robot navigation: {A} learning approach.
\newblock In: \emph{2012 {IEEE}/{RSJ} {International} {Conference} on {Intelligent} {Robots} and {Systems}}. pp. 902--907.

\bibitem[{Luo et~al.(2018)Luo, Cai, Bera, Hsu, Lee and Manocha}]{luo_porca_2018}
Luo Y, Cai P, Bera A, Hsu D, Lee WS and Manocha D (2018) Porca: Modeling and planning for autonomous driving among many pedestrians.
\newblock \emph{IEEE Robotics and Automation Letters} 3: 3418--3425.

\bibitem[{Macenski et~al.(2020)Macenski, Mart{\'\i}n, White and Clavero}]{macenski2020marathon2}
Macenski S, Mart{\'\i}n F, White R and Clavero JG (2020) The marathon 2: A navigation system.
\newblock In: \emph{2020 IEEE/RSJ International Conference on Intelligent Robots and Systems (IROS)}. IEEE, pp. 2718--2725.

\bibitem[{Majd et~al.(2021)Majd, Yaghoubi, Yamaguchi, Hoxha, Prokhorov and Fainekos}]{majd_safe_2021}
Majd K, Yaghoubi S, Yamaguchi T, Hoxha B, Prokhorov D and Fainekos G (2021) Safe {Navigation} in {Human} {Occupied} {Environments} {Using} {Sampling} and {Control} {Barrier} {Functions}.
\newblock In: \emph{2021 {IEEE}/{RSJ} {International} {Conference} on {Intelligent} {Robots} and {Systems} ({IROS})}. IEEE, pp. 5794--5800.

\bibitem[{Manh and Alaghband(2018)}]{manh_scene-lstm_2019}
Manh H and Alaghband G (2018) Scene-{LSTM}: A model for human trajectory prediction.
\newblock \emph{arXiv:1808.04018} .

\bibitem[{Manso et~al.(2019)Manso, Jorvekar, Faria, Bustos and Bachiller}]{manso_graph_2019}
Manso LJ, Jorvekar RR, Faria DR, Bustos P and Bachiller P (2019) Graph {Neural} {Networks} for {Human}-aware {Social} {Navigation}.
\newblock \emph{arXiv:1909.09003 [cs]} .

\bibitem[{Manso et~al.(2020)Manso, Nu{\~n}ez, Calderita, Faria and Bachiller}]{manso_socnav1_2020}
Manso LJ, Nu{\~n}ez P, Calderita LV, Faria DR and Bachiller P (2020) Socnav1: A dataset to benchmark and learn social navigation conventions.
\newblock \emph{Data} 5(1): 7.

\bibitem[{Marge et~al.(2017)Marge, Bonial, Foots, Hayes, Henry, Pollard, Artstein, Voss and Traum}]{marge2017exploring}
Marge M, Bonial C, Foots A, Hayes C, Henry C, Pollard K, Artstein R, Voss C and Traum D (2017) Exploring variation of natural human commands to a robot in a collaborative navigation task.
\newblock In: \emph{Proceedings of the first workshop on language grounding for robotics}. pp. 58--66.

\bibitem[{Mateus et~al.(2019)Mateus, Ribeiro, Miraldo and Nascimento}]{mateus_efficient_2016}
Mateus A, Ribeiro D, Miraldo P and Nascimento JC (2019) Efficient and robust pedestrian detection using deep learning for human-aware navigation.
\newblock \emph{Robotics and Autonomous Systems} 113: 23--37.

\bibitem[{Mavrogiannis et~al.(2023)Mavrogiannis, Baldini, Wang, Zhao, Trautman, Steinfeld and Oh}]{mavrogiannis_core_2023}
Mavrogiannis C, Baldini F, Wang A, Zhao D, Trautman P, Steinfeld A and Oh J (2023) Core challenges of social robot navigation: A survey.
\newblock \emph{ACM Transactions on Human-Robot Interaction} 12(3): 1--39.

\bibitem[{Mavrogiannis et~al.(2019)Mavrogiannis, Hutchinson, Macdonald, Alves-Oliveira and Knepper}]{mavrogiannis_effects_2019}
Mavrogiannis C, Hutchinson AM, Macdonald J, Alves-Oliveira P and Knepper RA (2019) Effects of {Distinct} {Robot} {Navigation} {Strategies} on {Human} {Behavior} in a {Crowded} {Environment}.
\newblock In: \emph{2019 14th {ACM}/{IEEE} {International} {Conference} on {Human}-{Robot} {Interaction} ({HRI})}. IEEE, pp. 421--430.

\bibitem[{Mavrogiannis et~al.(2018)Mavrogiannis, Thomason and Knepper}]{mavrogiannis_social_2018}
Mavrogiannis CI, Thomason WB and Knepper RA (2018) Social {Momentum}: {A} {Framework} for {Legible} {Navigation} in {Dynamic} {Multi}-{Agent} {Environments}.
\newblock In: \emph{Proceedings of the 2018 {ACM}/{IEEE} {International} {Conference} on {Human}-{Robot} {Interaction} - {HRI} '18}. ACM Press, pp. 361--369.

\bibitem[{May et~al.(2015)May, Dondrup and Hanheide}]{may_show_2015}
May AD, Dondrup C and Hanheide M (2015) Show me your moves! {Conveying} navigation intention of a mobile robot to humans.
\newblock In: \emph{2015 {European} {Conference} on {Mobile} {Robots} ({ECMR})}. pp. 1--6.

\bibitem[{Mead and Matari{\'c}(2017)}]{mead_autonomous_2017}
Mead R and Matari{\'c} MJ (2017) Autonomous human–robot proxemics: socially aware navigation based on interaction potential.
\newblock \emph{Autonomous Robots} 41(5): 1189--1201.

\bibitem[{Mirsky et~al.(2021)Mirsky, Xiao, Hart and Stone}]{mirsky_prevention_2021}
Mirsky R, Xiao X, Hart J and Stone P (2021) Prevention and resolution of conflicts in social navigation--a survey.
\newblock \emph{arXiv preprint arXiv:2106.12113} .

\bibitem[{Mizuchi and Inamura(2017)}]{mizuchi_cloud-based_2017}
Mizuchi Y and Inamura T (2017) Cloud-based multimodal human-robot interaction simulator utilizing {ROS} and unity frameworks.
\newblock In: \emph{2017 {IEEE}/{SICE} {International} {Symposium} on {System} {Integration} ({SII})}. IEEE, pp. 948--955.

\bibitem[{M{\"o}ller et~al.(2021)M{\"o}ller, Furnari, Battiato, H{\"a}rm{\"a} and Farinella}]{moller_survey_2021}
M{\"o}ller R, Furnari A, Battiato S, H{\"a}rm{\"a} A and Farinella GM (2021) A survey on human-aware robot navigation.
\newblock \emph{Robotics and Autonomous Systems} 145: 103837.

\bibitem[{Morales et~al.(2017)Morales, Miyashita and Hagita}]{morales_social_2017}
Morales Y, Miyashita T and Hagita N (2017) Social robotic wheelchair centered on passenger and pedestrian comfort.
\newblock \emph{Robotics and Autonomous Systems} 87: 355--362.

\bibitem[{Narayanan et~al.(2018{\natexlab{a}})Narayanan, Miyashita and Hagita}]{narayanan_formalizing_2018}
Narayanan VK, Miyashita T and Hagita N (2018{\natexlab{a}}) Formalizing a {Transient}-{Goal} {Driven} {Approach} for {Pedestrian}-{Aware} {Robot} {Navigation}.
\newblock In: \emph{2018 27th {IEEE} {International} {Symposium} on {Robot} and {Human} {Interactive} {Communication} ({RO}-{MAN})}. pp. 862--867.

\bibitem[{Narayanan et~al.(2018{\natexlab{b}})Narayanan, Miyashita, Horikawa and Hagita}]{narayanan_transient-goal_2018}
Narayanan VK, Miyashita T, Horikawa Y and Hagita N (2018{\natexlab{b}}) A {Transient}-{Goal} {Driven} {Communication}-{Aware} {Navigation} {Strategy} for {Large} {Human}-{Populated} {Environments}.
\newblock In: \emph{2018 {IEEE}/{RSJ} {International} {Conference} on {Intelligent} {Robots} and {Systems} ({IROS})}. pp. 1--9.

\bibitem[{Narayanan et~al.(2016)Narayanan, Spalanzani and Babel}]{narayanan_WC_2016}
Narayanan VK, Spalanzani A and Babel M (2016) A semi-autonomous framework for human-aware and user intention driven wheelchair mobility assistance.
\newblock In: \emph{2016 IEEE/RSJ International Conference on Intelligent Robots and Systems (IROS)}. pp. 4700--4707.

\bibitem[{Neggers et~al.(2018)Neggers, Cuijpers and Ruijten}]{neggers_comfortable_2018}
Neggers MM, Cuijpers RH and Ruijten PA (2018) Comfortable passing distances for robots.
\newblock In: \emph{Social Robotics: 10th International Conference, ICSR 2018, Qingdao, China, November 28-30, 2018, Proceedings 10}. Springer, pp. 431--440.

\bibitem[{Neggers et~al.(2022{\natexlab{a}})Neggers, Cuijpers, Ruijten and IJsselsteijn}]{neggers_determining_2022}
Neggers MM, Cuijpers RH, Ruijten PA and IJsselsteijn WA (2022{\natexlab{a}}) Determining shape and size of personal space of a human when passed by a robot.
\newblock \emph{International Journal of Social Robotics} : 1--12.

\bibitem[{Neggers et~al.(2022{\natexlab{b}})Neggers, Cuijpers, Ruijten and IJsselsteijn}]{neggers_theeffect_2022}
Neggers MM, Cuijpers RH, Ruijten PA and IJsselsteijn WA (2022{\natexlab{b}}) The effect of robot speed on comfortable passing distances.
\newblock \emph{Frontiers in Robotics and AI} 9: 915972.

\bibitem[{Neggers et~al.(2022{\natexlab{c}})Neggers, Ruijten, Cuijpers and IJsselsteijn}]{neggers_effect_2022}
Neggers MM, Ruijten PA, Cuijpers RH and IJsselsteijn WA (2022{\natexlab{c}}) Effect of robot gazing behavior on human comfort and robot predictability in navigation.
\newblock In: \emph{2022 IEEE International Conference on Advanced Robotics and Its Social Impacts (ARSO)}. IEEE, pp. 1--6.

\bibitem[{Ngo(2021)}]{ngo_recent_2021}
Ngo HQT (2021) Recent researches on human-aware navigation for autonomous system in the dynamic environment: An international survey.
\newblock In: \emph{Context-Aware Systems and Applications: 10th EAI International Conference, ICCASA 2021, Virtual Event, October 28--29, 2021, Proceedings 10}. Springer, pp. 267--282.

\bibitem[{Nishimura and Yonetani(2020)}]{nishimura_l2b_2020}
Nishimura M and Yonetani R (2020) {L2B}: Learning to balance the safety-efficiency trade-off in interactive crowd-aware robot navigation.
\newblock In: \emph{IEEE/RSJ IROS}. IEEE, pp. 11004--11010.

\bibitem[{Obo(2018)}]{obo_intelligent_2018}
Obo T (2018) Intelligent {Fuzzy} {Controller} for {Human}-{Aware} {Robot} {Navigation}.
\newblock In: \emph{2018 12th {France}-{Japan} and 10th {Europe}-{Asia} {Congress} on {Mechatronics}}. pp. 392--397.

\bibitem[{Paez-Granados et~al.(2022)Paez-Granados, Gupta and Billard}]{paez-granados_unfreezing_2022}
Paez-Granados D, Gupta V and Billard A (2022) Unfreezing social navigation: Dynamical systems based compliance for contact control in robot navigation.
\newblock In: \emph{2022 International Conference on Robotics and Automation (ICRA)}. IEEE, pp. 8368--8374.

\bibitem[{Palinko et~al.(2020)Palinko, Ram{\'\i}rez, Juel, Kr{\"u}ger and Bodenhagen}]{palinko_intention_2020}
Palinko O, Ram{\'\i}rez ER, Juel WK, Kr{\"u}ger N and Bodenhagen L (2020) Intention indication for human aware robot navigation.
\newblock In: \emph{Proceedings of the 15th {International} {Joint} {Conference} on {Computer} {Vision}, {Imaging} and {Computer} {Graphics} {Theory} and {Applications}}. SCITEPRESS - Science and Technology Publications, pp. 64--74.

\bibitem[{Parhi and Singh(2010)}]{parhi2010heuristic}
Parhi D and Singh M (2010) Heuristic-rule-based hybrid neural network for navigation of a mobile robot.
\newblock \emph{Proceedings of the Institution of Mechanical Engineers, Part B: Journal of Engineering Manufacture} 224(7): 1103--1118.

\bibitem[{Park et~al.(2016)Park, Ondřej, Gilbert, Freeman and O'Sullivan}]{park_hi_2016}
Park C, Ondřej J, Gilbert M, Freeman K and O'Sullivan C (2016) {HI} {Robot}: {Human} intention-aware robot planning for safe and efficient navigation in crowds.
\newblock In: \emph{2016 {IEEE}/{RSJ} {International} {Conference} on {Intelligent} {Robots} and {Systems} ({IROS})}. pp. 3320--3326.

\bibitem[{Patompak et~al.(2016)Patompak, Jeong, Chong and Nilkhamhang}]{patompak_mobile_2016}
Patompak P, Jeong S, Chong NY and Nilkhamhang I (2016) Mobile robot navigation for human-robot social interaction.
\newblock In: \emph{2016 16th {International} {Conference} on {Control}, {Automation} and {Systems} ({ICCAS})}. pp. 1298--1303.

\bibitem[{Peddi et~al.(2020)Peddi, Di~Franco, Gao and Bezzo}]{peddi_data-driven_2020}
Peddi R, Di~Franco C, Gao S and Bezzo N (2020) A data-driven framework for proactive intention-aware motion planning of a robot in a human environment.
\newblock In: \emph{2020 IEEE/RSJ International Conference on Intelligent Robots and Systems (IROS)}. IEEE, pp. 5738--5744.

\bibitem[{P{\'e}rez-Higueras et~al.(2018)P{\'e}rez-Higueras, Caballero and Merino}]{perez-higueras_learning_2018}
P{\'e}rez-Higueras N, Caballero F and Merino L (2018) Learning human-aware path planning with fully convolutional networks.
\newblock In: \emph{2018 IEEE international conference on robotics and automation (ICRA)}. IEEE, pp. 5897--5902.

\bibitem[{P{\'e}rez-Higueras et~al.(2022)P{\'e}rez-Higueras, Otero, Caballero and Merino}]{perez_hunavsim_2022}
P{\'e}rez-Higueras N, Otero R, Caballero F and Merino L (2022) Hunavsim: A ros2 human navigation simulator for benchmarking human-aware robot navigation.
\newblock \emph{IEEE/RSJ IROS Workshop: Benchmarking for Motion Planning Applications} : 49.

\bibitem[{P{\'e}rez-Higueras et~al.(2014)P{\'e}rez-Higueras, Ramón-Vigo, Caballero and Merino}]{perez-higueras_robot_2014}
P{\'e}rez-Higueras N, Ramón-Vigo R, Caballero F and Merino L (2014) Robot local navigation with learned social cost functions.
\newblock In: \emph{2014 11th {International} {Conference} on {Informatics} in {Control}, {Automation} and {Robotics} ({ICINCO})}, volume~02. pp. 618--625.

\bibitem[{Petrak et~al.(2021)Petrak, Sopper, Weitz and Andre}]{petrak_you_2021}
Petrak B, Sopper G, Weitz K and Andre E (2021) Do {You} {Mind} if {I} {Pass} {Through}? {Studying} the {Appropriate} {Robot} {Behavior} when {Traversing} two {Conversing} {People} in a {Hallway} {Setting}.
\newblock In: \emph{2021 30th {IEEE} {International} {Conference} on {Robot} \& {Human} {Interactive} {Communication} ({RO}-{MAN})}. IEEE, pp. 369--375.

\bibitem[{Pimentel and Aquino-Jr(2021)}]{pimentel_evaluation_2021}
Pimentel FdAM and Aquino-Jr PT (2021) Evaluation of {ROS} navigation stack for social navigation in simulated environments.
\newblock \emph{Journal of Intelligent \& Robotic Systems} 102: 1--18.

\bibitem[{Pol and Murugan(2015)}]{pol_review_2015}
Pol RS and Murugan M (2015) A review on indoor human aware autonomous mobile robot navigation through a dynamic environment survey of different path planning algorithm and methods.
\newblock In: \emph{2015 International conference on industrial instrumentation and control (ICIC)}. IEEE, pp. 1339--1344.

\bibitem[{Pr{\'e}dhumeau et~al.(2021)Pr{\'e}dhumeau, Spalanzani and Dugdale}]{m_predhumeau_a_spalanzani_j_dugdale_pedestrian_2021}
Pr{\'e}dhumeau M, Spalanzani A and Dugdale J (2021) Pedestrian behavior in shared spaces with autonomous vehicles: An integrated framework and review.
\newblock \emph{IEEE Transactions on Intelligent Vehicles} .

\bibitem[{Prédhumeau et~al.(2021)Prédhumeau, Mancheva, Dugdale and Spalanzani}]{predhumeau_agent-based_2021}
Prédhumeau M, Mancheva L, Dugdale J and Spalanzani A (2021) An {Agent}-{Based} {Model} to {Predict} {Pedestrians} {Trajectories} with an {Autonomous} {Vehicle} in {Shared} {Spaces}.
\newblock In: \emph{{AAMAS} 2021 - 20th {International} {Conference} on {Autonomous} {Agents} and {Multiagent} {Systems}}. International Foundation for Autonomous Agents and Multiagent Systems (IFAAMAS), pp. 1--9.

\bibitem[{Puig-Pey et~al.(2023)Puig-Pey, Zamora, Amante, Garrell, Grau, Bolea, Santamaria and Sanfeliu}]{puig-pey2023}
Puig-Pey A, Zamora J, Amante J B~Moreno, Garrell A, Grau A, Bolea Y, Santamaria A and Sanfeliu A (2023) Human acceptance in the human-robot interaction scenario for last-mile goods delivery.
\newblock In: \emph{2023 IEEE International Conference on Advanced Robotics and Its Social Impacts(ARSO)}. IEEE, pp. 1--6.

\bibitem[{Qian et~al.(2013)Qian, Ma, Dai, Fang and Zhou}]{qian_decision-theoretical_2013}
Qian K, Ma X, Dai X, Fang F and Zhou B (2013) Decision-{Theoretical} {Navigation} of {Service} {Robots} {Using} {POMDPs} with {Human}-{Robot} {Co}-{Occurrence} {Prediction}.
\newblock \emph{International Journal of Advanced Robotic Systems} 10(2): 143.

\bibitem[{Qiu et~al.(2022)Qiu, Yao, Wang, Ma, Chen and Ji}]{qiu_learning_2022}
Qiu Q, Yao S, Wang J, Ma J, Chen G and Ji J (2022) Learning to socially navigate in pedestrian-rich environments with interaction capacity.
\newblock In: \emph{2022 International Conference on Robotics and Automation (ICRA)}. IEEE, pp. 279--285.

\bibitem[{Ramirez et~al.(2016)Ramirez, Khambhaita, Chatila, Chetouani and Alami}]{ramirez_robots_2016}
Ramirez OAI, Khambhaita H, Chatila R, Chetouani M and Alami R (2016) Robots learning how and where to approach people.
\newblock In: \emph{2016 25th {IEEE} {International} {Symposium} on {Robot} and {Human} {Interactive} {Communication} ({RO}-{MAN})}. IEEE, pp. 347--353.

\bibitem[{Randhavane et~al.(2019)Randhavane, Bera, Kubin, Wang, Gray and Manocha}]{randhavane_pedestrian_2019}
Randhavane T, Bera A, Kubin E, Wang A, Gray K and Manocha D (2019) Pedestrian dominance modeling for socially-aware robot navigation.
\newblock In: \emph{2019 International Conference on Robotics and Automation (ICRA)}. IEEE, pp. 5621--5628.

\bibitem[{Rasouli and Tsotsos(2019)}]{rasouli_autonomous_2020}
Rasouli A and Tsotsos JK (2019) Autonomous vehicles that interact with pedestrians: A survey of theory and practice.
\newblock \emph{IEEE Transactions on Intelligent Transportation systems} 21: 900--918.

\bibitem[{Ratsamee et~al.(2013)Ratsamee, Mae, Ohara, Kojima and Arai}]{ratsamee_social_2013}
Ratsamee P, Mae Y, Ohara K, Kojima M and Arai T (2013) Social navigation model based on human intention analysis using face orientation.
\newblock In: \emph{2013 IEEE/RSJ International Conference on Intelligent Robots and Systems}. IEEE, pp. 1682--1687.

\bibitem[{Renault et~al.(2019)Renault, Saraydaryan and Simonin}]{renault_towards_2019}
Renault B, Saraydaryan J and Simonin O (2019) Towards s-namo: Socially-aware navigation among movable obstacles.
\newblock In: \emph{RoboCup 2019: Robot World Cup XXIII 23}. pp. 241--254.

\bibitem[{Repiso et~al.(2017)Repiso, Ferrer and Sanfeliu}]{repiso_-line_2017}
Repiso E, Ferrer G and Sanfeliu A (2017) On-line adaptive side-by-side human robot companion in dynamic urban environments.
\newblock In: \emph{2017 {IEEE}/{RSJ} {International} {Conference} on {Intelligent} {Robots} and {Systems} ({IROS})}. pp. 872--877.

\bibitem[{Repiso et~al.(2018)Repiso, Garrell and Sanfeliu}]{repiso_robot_2018}
Repiso E, Garrell A and Sanfeliu A (2018) Robot {Approaching} and {Engaging} {People} in a {Human}-{Robot} {Companion} {Framework}.
\newblock In: \emph{2018 {IEEE}/{RSJ} {International} {Conference} on {Intelligent} {Robots} and {Systems} ({IROS})}. pp. 8200--8205.

\bibitem[{Repiso et~al.(2020{\natexlab{a}})Repiso, Garrell and Sanfeliu}]{repiso_adaptive_2019}
Repiso E, Garrell A and Sanfeliu A (2020{\natexlab{a}}) Adaptive side-by-side social robot navigation to approach and interact with people.
\newblock \emph{International Journal of Social Robotics} 12: 909--930.

\bibitem[{Repiso et~al.(2020{\natexlab{b}})Repiso, Garrell and Sanfeliu}]{repiso_peoples_2020}
Repiso E, Garrell A and Sanfeliu A (2020{\natexlab{b}}) People's {Adaptive} {Side}-by-{Side} {Model} {Evolved} to {Accompany} {Groups} of {People} by {Social} {Robots}.
\newblock \emph{IEEE Robotics and Automation Letters} 5(2): 2387--2394.

\bibitem[{Repiso et~al.(2022)Repiso, Garrell and Sanfeliu}]{repiso_adaptive_2022}
Repiso E, Garrell A and Sanfeliu A (2022) Adaptive social planner to accompany people in real-life dynamic environments.
\newblock \emph{International Journal of Social Robotics} : 1--33.

\bibitem[{Repiso et~al.(2019)Repiso, Zanlungo, Kanda, Garrell and Sanfeliu}]{repiso_peoples_2019}
Repiso E, Zanlungo F, Kanda T, Garrell A and Sanfeliu A (2019) People’s {V}-{Formation} and {Side}-by-{Side} {Model} {Adapted} to {Accompany} {Groups} of {People} by {Social} {Robots}.
\newblock In: \emph{2019 {IEEE}/{RSJ} {International} {Conference} on {Intelligent} {Robots} and {Systems} ({IROS})}. pp. 2082--2088.

\bibitem[{Ridel et~al.(2018)Ridel, Rehder, Lauer, Stiller and Wolf}]{ridel_literature_2018}
Ridel D, Rehder E, Lauer M, Stiller C and Wolf D (2018) A literature review on the prediction of pedestrian behavior in urban scenarios.
\newblock In: \emph{2018 21st International Conference on Intelligent Transportation Systems (ITSC)}. IEEE, pp. 3105--3112.

\bibitem[{Rios-Martinez et~al.(2012)Rios-Martinez, Escobedo, Spalanzani and Laugier}]{rios-martinez_intention_2012}
Rios-Martinez J, Escobedo A, Spalanzani A and Laugier C (2012) Intention driven human aware navigation for assisted mobility.
\newblock In: \emph{Workshop on Assistance and Service robotics in a human environment at IROS}.

\bibitem[{Rios-Martinez et~al.(2015)Rios-Martinez, Spalanzani and Laugier}]{rios-martinez_proxemics_2015}
Rios-Martinez J, Spalanzani A and Laugier C (2015) From proxemics theory to socially-aware navigation: A survey.
\newblock \emph{International Journal of Social Robotics} 7: 137--153.

\bibitem[{R{\"o}smann et~al.(2017)R{\"o}smann, Oeljeklaus, Hoffmann and Bertram}]{rosmann_online_2017}
R{\"o}smann C, Oeljeklaus M, Hoffmann F and Bertram T (2017) Online trajectory prediction and planning for social robot navigation.
\newblock In: \emph{2017 {IEEE} {International} {Conference} on {Advanced} {Intelligent} {Mechatronics} ({AIM})}. pp. 1255--1260.

\bibitem[{Rudenko et~al.(2020{\natexlab{a}})Rudenko, Kucner, Swaminathan, Chadalavada, Arras and Lilienthal}]{rudenko_thor_2020}
Rudenko A, Kucner TP, Swaminathan CS, Chadalavada RT, Arras KO and Lilienthal AJ (2020{\natexlab{a}}) {TH{\"O}R}: {Human}-{Robot} {Navigation} {Data} {Collection} and {Accurate} {Motion} {Trajectories} {Dataset}.
\newblock \emph{IEEE Robotics and Automation Letters} 5(2): 676--682.

\bibitem[{Rudenko et~al.(2020{\natexlab{b}})Rudenko, Palmieri, Herman, Kitani, Gavrila and Arras}]{rudenko_human_2020}
Rudenko A, Palmieri L, Herman M, Kitani KM, Gavrila DM and Arras KO (2020{\natexlab{b}}) Human motion trajectory prediction: A survey.
\newblock \emph{The International Journal of Robotics Research} 39: 895--935.

\bibitem[{Salvini et~al.(2022)Salvini, Paez-Granados and Billard}]{salvini_safety_2022}
Salvini P, Paez-Granados D and Billard A (2022) Safety concerns emerging from robots navigating in crowded pedestrian areas.
\newblock \emph{International Journal of Social Robotics} 14: 441--462.

\bibitem[{Samarakoon et~al.(2022)Samarakoon, Muthugala and Jayasekara}]{samarakoon2022review}
Samarakoon SBP, Muthugala MVJ and Jayasekara ABP (2022) A review on human--robot proxemics.
\newblock \emph{Electronics} 11: 2490.

\bibitem[{Sathyamoorthy et~al.(2020)Sathyamoorthy, Patel, Guan and Manocha}]{sathyamoorthy_frozone_2020}
Sathyamoorthy AJ, Patel U, Guan T and Manocha D (2020) Frozone: Freezing-free, pedestrian-friendly navigation in human crowds.
\newblock \emph{IEEE Robotics and Automation Letters} 5: 4352--4359.

\bibitem[{Senft et~al.(2020)Senft, Satake and Kanda}]{senft_would_2020}
Senft E, Satake S and Kanda T (2020) Would {You} {Mind} {Me} if {I} {Pass} by {You}?: {Socially}-{Appropriate} {Behaviour} for an {Omni}-based {Social} {Robot} in {Narrow} {Environment}.
\newblock In: \emph{Proceedings of the 2020 {ACM}/{IEEE} {International} {Conference} on {Human}-{Robot} {Interaction}}. ACM, pp. 539--547.

\bibitem[{Shahrezaie et~al.(2022)Shahrezaie, Manalo, Brantley, Lynch and Feil-Seifer}]{shahrezaie_advancing_2022}
Shahrezaie RS, Manalo BN, Brantley AG, Lynch CR and Feil-Seifer D (2022) Advancing socially-aware navigation for public spaces.
\newblock In: \emph{2022 31st IEEE International Conference on Robot and Human Interactive Communication (RO-MAN)}. IEEE, pp. 1015--1022.

\bibitem[{Shin and Yoon(2020)}]{shin_optimization-based_2020}
Shin H and Yoon SE (2020) Optimization-based path planning for person following using following field.
\newblock In: \emph{2020 IEEE/RSJ International Conference on Intelligent Robots and Systems (IROS)}. IEEE, pp. 11352--11359.

\bibitem[{Shrestha et~al.(2015)Shrestha, Nohisa, Schmitz, Hayakawa, Uno, Yokoyama, Yanagawa, Or and Sugano}]{shrestha_using_2015}
Shrestha MC, Nohisa Y, Schmitz A, Hayakawa S, Uno E, Yokoyama Y, Yanagawa H, Or K and Sugano S (2015) Using contact-based inducement for efficient navigation in a congested environment.
\newblock In: \emph{2015 24th {IEEE} {International} {Symposium} on {Robot} and {Human} {Interactive} {Communication} ({RO}-{MAN})}. pp. 456--461.

\bibitem[{Singamaneni(2022)}]{singamaneni2022combining}
Singamaneni PT (2022) \emph{Combining proactive planning and situation analysis for human-aware robot navigation}.
\newblock PhD Thesis, Toulouse 3.

\bibitem[{Singamaneni et~al.(2021)Singamaneni, Favier and Alami}]{singamaneni_human-aware_2021}
Singamaneni PT, Favier A and Alami R (2021) Human-{Aware} {Navigation} {Planner} for {Diverse} {Human}-{Robot} {Contexts}.
\newblock In: \emph{2021 {IEEE}/{RSJ} {International} {Conference} on {Intelligent} {Robots} and {Systems} ({IROS})}. IEEE, pp. 5817--5824.

\bibitem[{Singamaneni et~al.(2022)Singamaneni, Favier and Alami}]{singamaneni_watch_2022}
Singamaneni PT, Favier A and Alami R (2022) Watch out! there may be a human. addressing invisible humans in social navigation.
\newblock In: \emph{2022 IEEE/RSJ International Conference on Intelligent Robots and Systems (IROS)}. IEEE, pp. 11344--11351.

\bibitem[{Sivakanthan et~al.(2022)Sivakanthan, Candiotti, Sundaram, Duvall, Sergeant, Cooper, Satpute, Turner and Cooper}]{sivakanthan2022mini}
Sivakanthan S, Candiotti JL, Sundaram AS, Duvall JA, Sergeant JJG, Cooper R, Satpute S, Turner RL and Cooper RA (2022) Mini-review: Robotic wheelchair taxonomy and readiness.
\newblock \emph{Neuroscience letters} : 136482.

\bibitem[{Skrzypczyk(2021)}]{skrzypczyk_game_2021}
Skrzypczyk K (2021) Game {Against} {Nature} {Based} {Control} of an {Intelligent} {Wheelchair} with {Adaptation} to {Pedestrians}' {Behaviour}.
\newblock In: \emph{2021 25th {International} {Conference} on {Methods} and {Models} in {Automation} and {Robotics} ({MMAR})}. IEEE, pp. 285--290.

\bibitem[{Song et~al.(2018)Song, Chen, Qi, Zhao, Hu, Liu and Zhang}]{song_human_2018}
Song C, Chen Z, Qi X, Zhao B, Hu Y, Liu S and Zhang J (2018) Human trajectory prediction for automatic guided vehicle with recurrent neural network.
\newblock \emph{The Journal of Engineering} 2018: 1574--1578.

\bibitem[{Song et~al.(2021)Song, Naji, Kaufmann, Loquercio and Scaramuzza}]{song_flightmare_2021}
Song Y, Naji S, Kaufmann E, Loquercio A and Scaramuzza D (2021) Flightmare: A flexible quadrotor simulator.
\newblock In: \emph{Conference on Robot Learning}. PMLR, pp. 1147--1157.

\bibitem[{Sorrentino et~al.(2021)Sorrentino, Khalid, Coviello, Cavallo and Fiorini}]{sorrentino_modeling_2021}
Sorrentino A, Khalid O, Coviello L, Cavallo F and Fiorini L (2021) Modeling human-like robot personalities as a key to foster socially aware navigation.
\newblock In: \emph{2021 30th {IEEE} {International} {Conference} on {Robot} \& {Human} {Interactive} {Communication} ({RO}-{MAN})}. IEEE, pp. 95--101.

\bibitem[{Szafir et~al.(2014)Szafir, Mutlu and Fong}]{szafir_communication_2014}
Szafir D, Mutlu B and Fong T (2014) Communication of intent in assistive free flyers.
\newblock In: \emph{Proceedings of the 2014 {ACM}/{IEEE} international conference on {Human}-robot interaction}. ACM, pp. 358--365.

\bibitem[{Szafir et~al.(2015)Szafir, Mutlu and Fong}]{szafir_communicating_2015}
Szafir D, Mutlu B and Fong T (2015) Communicating {Directionality} in {Flying} {Robots}.
\newblock In: \emph{Proceedings of the {Tenth} {Annual} {ACM}/{IEEE} {International} {Conference} on {Human}-{Robot} {Interaction}}. ACM, pp. 19--26.

\bibitem[{Tadokoro et~al.(1995)Tadokoro, Hayashi, Manabe, Nakami and Takamori}]{tadokoro1995motion}
Tadokoro S, Hayashi M, Manabe Y, Nakami Y and Takamori T (1995) On motion planning of mobile robots which coexist and cooperate with human.
\newblock In: \emph{Proceedings 1995 IEEE/RSJ International Conference on Intelligent Robots and Systems. Human Robot Interaction and Cooperative Robots}, volume~2. IEEE, pp. 518--523.

\bibitem[{Talebpour et~al.(2015)Talebpour, Navarro and Martinoli}]{talebpour_-board_2015}
Talebpour Z, Navarro I and Martinoli A (2015) On-board human-aware navigation for indoor resource-constrained robots: {A} case-study with the ranger.
\newblock In: \emph{2015 {IEEE}/{SICE} {International} {Symposium} on {System} {Integration} ({SII})}. pp. 63--68.

\bibitem[{Talebpour et~al.(2016)Talebpour, Viswanathan, Ventura, Englebienne and Martinoli}]{talebpour_incorporating_2016}
Talebpour Z, Viswanathan D, Ventura R, Englebienne G and Martinoli A (2016) Incorporating perception uncertainty in human-aware navigation: {A} comparative study.
\newblock In: \emph{2016 25th {IEEE} {International} {Symposium} on {Robot} and {Human} {Interactive} {Communication} ({RO}-{MAN})}. pp. 570--577.

\bibitem[{Taylor et~al.(2022)Taylor, Mamantov and Admoni}]{taylor_observer_2022}
Taylor AV, Mamantov E and Admoni H (2022) Observer-aware legibility for social navigation.
\newblock In: \emph{2022 31st IEEE International Conference on Robot and Human Interactive Communication (RO-MAN)}. IEEE, pp. 1115--1122.

\bibitem[{Teja~S. and Alami(2020)}]{teja_s_hateb-2_2020}
Teja~S P and Alami R (2020) {HATEB}-2: {Reactive} {Planning} and {Decision} making in {Human}-{Robot} {Co}-navigation.
\newblock In: \emph{2020 29th {IEEE} {International} {Conference} on {Robot} and {Human} {Interactive} {Communication} ({RO}-{MAN})}. IEEE, pp. 179--186.

\bibitem[{Toghi et~al.(2021)Toghi, Valiente, Sadigh, Pedarsani and Fallah}]{toghi_altruistic_2021}
Toghi B, Valiente R, Sadigh D, Pedarsani R and Fallah YP (2021) Altruistic {Maneuver} {Planning} for {Cooperative} {Autonomous} {Vehicles} {Using} {Multi}-agent {Advantage} {Actor}-{Critic}.
\newblock \emph{arXiv:2107.05664 [cs]} .

\bibitem[{Toghi et~al.(2022)Toghi, Valiente, Sadigh, Pedarsani and Fallah}]{toghi_social_2022}
Toghi B, Valiente R, Sadigh D, Pedarsani R and Fallah YP (2022) Social coordination and altruism in autonomous driving.
\newblock \emph{IEEE Transactions on Intelligent Transportation Systems} 23: 24791--24804.

\bibitem[{Trautman et~al.(2015)Trautman, Ma, Murray and Krause}]{trautman_robot_2015}
Trautman P, Ma J, Murray RM and Krause A (2015) Robot navigation in dense human crowds: {Statistical} models and experimental studies of human–robot cooperation.
\newblock \emph{The International Journal of Robotics Research} 34(3): 335--356.

\bibitem[{Triebel et~al.(2016)Triebel, Arras, Alami, Beyer, Breuers, Chatila, Chetouani, Cremers, Evers, Fiore, Hung, Ramírez, Joosse, Khambhaita, Kucner, Leibe, Lilienthal, Linder, Lohse, Magnusson, Okal, Palmieri, Rafi, van Rooij and Zhang}]{triebel_spencer_2016}
Triebel R, Arras K, Alami R, Beyer L, Breuers S, Chatila R, Chetouani M, Cremers D, Evers V, Fiore M, Hung H, Ramírez OAI, Joosse M, Khambhaita H, Kucner T, Leibe B, Lilienthal AJ, Linder T, Lohse M, Magnusson M, Okal B, Palmieri L, Rafi U, van Rooij M and Zhang L (2016) {SPENCER}: {A} {Socially} {Aware} {Service} {Robot} for {Passenger} {Guidance} and {Help} in {Busy} {Airports}.
\newblock In: \emph{Field and {Service} {Robotics}}, volume 113. Springer International Publishing, pp. 607--622.

\bibitem[{Truc et~al.(2022)Truc, Singamaneni, Sidobre, Ivaldi and Alami}]{truc_khaos_2022}
Truc J, Singamaneni PT, Sidobre D, Ivaldi S and Alami R (2022) {KHAOS}: a {Kinematic} {Human} {Aware} {Optimization}-based {System} for {Reactive} {Planning} of {Flying}-{Coworker}.
\newblock In: \emph{{ICRA} 2022- {IEEE} {International} {Conference} on {Robotics} and {Automation} 2022}. pp. 4764--4770.

\bibitem[{Truong and Ngo(2017)}]{truong_toward_2017}
Truong XT and Ngo TD (2017) Toward {Socially} {Aware} {Robot} {Navigation} in {Dynamic} and {Crowded} {Environments}: {A} {Proactive} {Social} {Motion} {Model}.
\newblock \emph{IEEE Transactions on Automation Science and Engineering} 14(4): 1743--1760.

\bibitem[{Truong and Ngo(2018)}]{truong_approach_2018}
Truong XT and Ngo TD (2018) “{To} {Approach} {Humans}?”: {A} {Unified} {Framework} for {Approaching} {Pose} {Prediction} and {Socially} {Aware} {Robot} {Navigation}.
\newblock \emph{IEEE Transactions on Cognitive and Developmental Systems} 10(3): 557--572.

\bibitem[{Truong and Ngo(2019)}]{truong_integrative_2019}
Truong XT and Ngo TD (2019) An integrative approach of social dynamic long short-term memory and deep reinforcement learning for socially aware robot navigation.
\newblock In: \emph{Long-term Human Motion Prediction Workshop ICRA}.

\bibitem[{Truong et~al.(2017)Truong, Yoong and Ngo}]{truong_socially_2017}
Truong XT, Yoong VN and Ngo TD (2017) Socially aware robot navigation system in human interactive environments.
\newblock \emph{Intelligent Service Robotics} 10(4): 287--295.

\bibitem[{Tsoi et~al.(2020)Tsoi, Hussein, Espinoza, Ruiz and V{\'a}zquez}]{tsoi_sean_2020}
Tsoi N, Hussein M, Espinoza J, Ruiz X and V{\'a}zquez M (2020) Sean: Social environment for autonomous navigation.
\newblock In: \emph{Proceedings of the 8th International Conference on Human-Agent Interaction}. pp. 281--283.

\bibitem[{Tsoi et~al.(2021)Tsoi, Hussein, Fugikawa, Zhao and Vazquez}]{tsoi_approach_2021}
Tsoi N, Hussein M, Fugikawa O, Zhao JD and Vazquez M (2021) An {Approach} to {Deploy} {Interactive} {Robotic} {Simulators} on the {Web} for {HRI} {Experiments}: {Results} in {Social} {Robot} {Navigation}.
\newblock In: \emph{2021 {IEEE}/{RSJ} {International} {Conference} on {Intelligent} {Robots} and {Systems} ({IROS})}. IEEE, pp. 7528--7535.

\bibitem[{Tsoi and Vazquez(2017)}]{tsoi_early_2017}
Tsoi N and Vazquez M (2017) Early {Prototyping} and {Human} {Evaluation} of {Social} {Robot} {Navigation} via {Online} {Interactive} {Simulations} : 6.

\bibitem[{Tsoi et~al.(2022)Tsoi, Xiang, Yu, Sohn, Schwartz, Ramesh, Hussein, Gupta, Kapadia and V{\'a}zquez}]{tsoi_sean_2022}
Tsoi N, Xiang A, Yu P, Sohn SS, Schwartz G, Ramesh S, Hussein M, Gupta AW, Kapadia M and V{\'a}zquez M (2022) Sean 2.0: Formalizing and generating social situations for robot navigation.
\newblock \emph{IEEE Robotics and Automation Letters} 7(4): 11047--11054.

\bibitem[{Unhelkar et~al.(2015)Unhelkar, Perez-D'Arpino, Stirling and Shah}]{unhelkar_human-robot_2015}
Unhelkar VV, Perez-D'Arpino C, Stirling L and Shah JA (2015) Human-robot co-navigation using anticipatory indicators of human walking motion.
\newblock In: \emph{2015 {IEEE} {International} {Conference} on {Robotics} and {Automation} ({ICRA})}. IEEE, pp. 6183--6190.

\bibitem[{Valiente et~al.(2022)Valiente, Toghi, Pedarsani and Fallah}]{valiente_robustness_2022}
Valiente R, Toghi B, Pedarsani R and Fallah YP (2022) Robustness and adaptability of reinforcement learning-based cooperative autonomous driving in mixed-autonomy traffic.
\newblock \emph{IEEE Open Journal of Intelligent Transportation Systems} 3: 397--410.

\bibitem[{Varshneya and Srinivasaraghavan(2017)}]{varshneya_human_2017}
Varshneya D and Srinivasaraghavan G (2017) Human trajectory prediction using spatially aware deep attention models.
\newblock \emph{arXiv preprint arXiv:1705.09436} .

\bibitem[{Vasconcelos et~al.(2015)Vasconcelos, Pereira, Macharet and Nascimento}]{vasconcelos_socially_2015}
Vasconcelos PA, Pereira HN, Macharet DG and Nascimento ER (2015) Socially {Acceptable} {Robot} {Navigation} in the {Presence} of {Humans}.
\newblock In: \emph{2015 12th {Latin} {American} {Robotics} {Symposium} and 2015 3rd {Brazilian} {Symposium} on {Robotics} ({LARS}-{SBR})}. IEEE, pp. 222--227.

\bibitem[{Vasquez et~al.(2014)Vasquez, Okal and Arras}]{vasquez_inverse_2014}
Vasquez D, Okal B and Arras KO (2014) Inverse {Reinforcement} {Learning} algorithms and features for robot navigation in crowds: {An} experimental comparison.
\newblock In: \emph{2014 {IEEE}/{RSJ} {International} {Conference} on {Intelligent} {Robots} and {Systems}}. pp. 1341--1346.

\bibitem[{Vasquez et~al.(2013)Vasquez, Stein, Rios-Martinez, Escobedo, Spalanzani and Laugier}]{vasquez_human_2013}
Vasquez D, Stein P, Rios-Martinez J, Escobedo A, Spalanzani A and Laugier C (2013) Human {Aware} {Navigation} for {Assistive} {Robotics}.
\newblock In: \emph{Experimental {Robotics}}, volume~88. Springer International Publishing, pp. 449--462.

\bibitem[{Vega et~al.(2019{\natexlab{a}})Vega, Manso, Cintas and N{\'u}{\~n}ez}]{vega_planning_2019}
Vega A, Manso LJ, Cintas R and N{\'u}{\~n}ez P (2019{\natexlab{a}}) Planning human-robot interaction for social navigation in crowded environments.
\newblock In: \emph{Advances in Physical Agents: Proceedings of the 19th International Workshop of Physical Agents (WAF 2018), November 22-23, 2018, Madrid, Spain}. Springer, pp. 195--208.

\bibitem[{Vega et~al.(2019{\natexlab{b}})Vega, Manso, Macharet, Bustos and Núñez}]{vega_socially_2019}
Vega A, Manso LJ, Macharet DG, Bustos P and Núñez P (2019{\natexlab{b}}) Socially aware robot navigation system in human-populated and interactive environments based on an adaptive spatial density function and space affordances.
\newblock \emph{Pattern Recognition Letters} 118: 72--84.

\bibitem[{Vega-Magro et~al.(2017)Vega-Magro, Manso, Bustos, Nunez and Macharet}]{vega-magro_socially_2017}
Vega-Magro A, Manso L, Bustos P, Nunez P and Macharet DG (2017) Socially acceptable robot navigation over groups of people.
\newblock In: \emph{2017 26th {IEEE} {International} {Symposium} on {Robot} and {Human} {Interactive} {Communication} ({RO}-{MAN})}. IEEE, pp. 1182--1187.

\bibitem[{Vega-Magro et~al.(2018)Vega-Magro, Manso, Bustos and Nunez}]{vega-magro_flexible_2018}
Vega-Magro A, Manso LJ, Bustos P and Nunez P (2018) A {Flexible} and {Adaptive} {Spatial} {Density} {Model} for {Context}-{Aware} {Social} {Mapping}: {Towards} a {More} {Realistic} {Social} {Navigation}.
\newblock In: \emph{2018 15th {International} {Conference} on {Control}, {Automation}, {Robotics} and {Vision} ({ICARCV})}. IEEE, pp. 1727--1732.

\bibitem[{Vemula et~al.(2018)Vemula, Muelling and Oh}]{vemula_social_2018}
Vemula A, Muelling K and Oh J (2018) Social attention: Modeling attention in human crowds.
\newblock In: \emph{2018 IEEE international Conference on Robotics and Automation (ICRA)}. IEEE, pp. 4601--4607.

\bibitem[{Wang et~al.(2022{\natexlab{a}})Wang, Biswas, Admoni and Steinfeld}]{wang_towards_2022}
Wang A, Biswas A, Admoni H and Steinfeld A (2022{\natexlab{a}}) Towards rich, portable, and large-scale pedestrian data collection.
\newblock \emph{arXiv preprint arXiv:2203.01974} .

\bibitem[{Wang et~al.(2016)Wang, Li, Ge and Lee}]{wang_adaptive_2016}
Wang C, Li Y, Ge SS and Lee TH (2016) Adaptive control for robot navigation in human environments based on social force model.
\newblock In: \emph{2016 {IEEE} {International} {Conference} on {Robotics} and {Automation} ({ICRA})}. IEEE, pp. 5690--5695.

\bibitem[{Wang et~al.(2022{\natexlab{b}})Wang, Chan, Carreno-Medrano, Cosgun and Croft}]{wang_metrics_2022}
Wang J, Chan WP, Carreno-Medrano P, Cosgun A and Croft E (2022{\natexlab{b}}) Metrics for evaluating social conformity of crowd navigation algorithms.
\newblock In: \emph{2022 IEEE International Conference on Advanced Robotics and Its Social Impacts (ARSO)}. IEEE, pp. 1--6.

\bibitem[{Wang et~al.(2022{\natexlab{c}})Wang, Wang and Min}]{wang_feedback_2022}
Wang R, Wang W and Min BC (2022{\natexlab{c}}) Feedback-efficient active preference learning for socially aware robot navigation.
\newblock In: \emph{2022 IEEE/RSJ International Conference on Intelligent Robots and Systems (IROS)}. IEEE, pp. 11336--11343.

\bibitem[{Wang et~al.(2018)Wang, Li, Chen, Diekel and Jia}]{wang2018facilitating}
Wang W, Li R, Chen Y, Diekel ZM and Jia Y (2018) Facilitating human--robot collaborative tasks by teaching-learning-collaboration from human demonstrations.
\newblock \emph{IEEE Transactions on Automation Science and Engineering} 16(2): 640--653.

\bibitem[{Wang et~al.(2022{\natexlab{d}})Wang, Wang, Zhang, Liu and Sun}]{wang_social_interactions_2022}
Wang W, Wang L, Zhang C, Liu C and Sun L (2022{\natexlab{d}}) \emph{Social Interactions for Autonomous Driving: A Review and Perspectives}.
\newblock ISBN 978-1-63828-128-3.
\newblock \doi{10.1561/9781638281290}.

\bibitem[{Weiss and Bartneck(2015)}]{weiss2015meta}
Weiss A and Bartneck C (2015) Meta analysis of the usage of the godspeed questionnaire series.
\newblock In: \emph{2015 24th IEEE International Symposium on Robot and Human Interactive Communication (RO-MAN)}. IEEE, pp. 381--388.

\bibitem[{Wilkes et~al.(1998)Wilkes, Alford, Pack, Rogers, Peters and Kawamura}]{wilkes1998toward}
Wilkes DM, Alford A, Pack RT, Rogers T, Peters R and Kawamura K (1998) Toward socially intelligent service robots.
\newblock \emph{Applied Artificial Intelligence} 12: 729--766.

\bibitem[{Winkle and Dautenhahn(2016)}]{winkle2016examining}
Winkle K and Dautenhahn K (2016) Examining the effects of robot behaviour on people's impression and empathy.
\newblock In: \emph{2016 25th IEEE International Symposium on Robot and Human Interactive Communication (RO-MAN)}. IEEE, pp. 352--357.

\bibitem[{Xie et~al.(2021)Xie, Xin and Dames}]{xie_towards_2021}
Xie Z, Xin P and Dames P (2021) Towards {Safe} {Navigation} {Through} {Crowded} {Dynamic} {Environments}.
\newblock In: \emph{2021 {IEEE}/{RSJ} {International} {Conference} on {Intelligent} {Robots} and {Systems} ({IROS})}. IEEE, pp. 4934--4940.

\bibitem[{Yao et~al.(2017)Yao, Anaya, Tao, Cho, Zheng and Zhang}]{yao_monocular_2017}
Yao N, Anaya E, Tao Q, Cho S, Zheng H and Zhang F (2017) Monocular vision-based human following on miniature robotic blimp.
\newblock In: \emph{2017 IEEE International Conference on Robotics and Automation (ICRA)}. IEEE, pp. 3244--3249.

\bibitem[{Yao et~al.(2019)Yao, Tao, Liu, Liu, Tian, Wang, Li and Zhang}]{yao_autonomous_2019}
Yao Ns, Tao Qy, Liu Wy, Liu Z, Tian Y, Wang Py, Li T and Zhang F (2019) Autonomous flying blimp interaction with human in an indoor space.
\newblock \emph{Frontiers of Information Technology \& Electronic Engineering} 20(1): 45--59.

\bibitem[{Ye et~al.(2020)Ye, Kim and Lee}]{ye2020context}
Ye J, Kim Y and Lee KM (2020) Context-aware natural language instructions for social robot navigation in indoor environments.
\newblock \emph{IEEE Transactions on Cognitive and Developmental Systems} 12(2): 143--152.

\bibitem[{Yeh et~al.(2017)Yeh, Ratsamee, Kiyokawa, Uranishi, Mashita, Takemura, Fjeld and Obaid}]{yeh_exploring_2017}
Yeh A, Ratsamee P, Kiyokawa K, Uranishi Y, Mashita T, Takemura H, Fjeld M and Obaid M (2017) Exploring {Proxemics} for {Human}-{Drone} {Interaction}.
\newblock In: \emph{Proceedings of the 5th {International} {Conference} on {Human} {Agent} {Interaction}}. ACM, pp. 81--88.

\bibitem[{Yen and Hickey(2004)}]{yen2004reinforcement}
Yen GG and Hickey TW (2004) Reinforcement learning algorithms for robotic navigation in dynamic environments.
\newblock \emph{ISA transactions} 43(2): 217--230.

\bibitem[{Yoon et~al.(2019)Yoon, Widdowson, Marinho, Wang and Hovakimyan}]{yoon_socially_2019}
Yoon HJ, Widdowson C, Marinho T, Wang RF and Hovakimyan N (2019) Socially {Aware} {Path} {Planning} for a {Flying} {Robot} in {Close} {Proximity} of {Humans}.
\newblock \emph{ACM Transactions on Cyber-Physical Systems} 3(4): 1--24.

\bibitem[{Zhu and Zhang(2021)}]{zhu2021deep}
Zhu K and Zhang T (2021) Deep reinforcement learning based mobile robot navigation: A review.
\newblock \emph{Tsinghua Science and Technology} 26(5): 674--691.

\bibitem[{Zimmermann et~al.(2021)Zimmermann, Poranne and Coros}]{zimmermann2021go}
Zimmermann S, Poranne R and Coros S (2021) Go fetch!-dynamic grasps using boston dynamics spot with external robotic arm.
\newblock In: \emph{2021 IEEE International Conference on Robotics and Automation (ICRA)}. IEEE, pp. 4488--4494.

\end{thebibliography}

\end{document}